\documentclass[11pt]{article}

\usepackage[final]{acl}

\usepackage{times}
\usepackage{latexsym}
\usepackage{comment}
\PassOptionsToPackage{hyphens}{url}\usepackage{hyperref}
\usepackage[dvipsnames]{xcolor}
\usepackage{amssymb}
\usepackage{subcaption}
\usepackage[table]{xcolor}
\usepackage{pifont}
\usepackage{tikz}
\usepackage{pgfplotstable}
\usepackage{tcolorbox} 
\usepackage{siunitx}
\usepackage{multirow}

\usepackage{pgfplots}
\pgfplotsset{compat=1.17}
\usepackage{tikz}
\usepackage{tikzscale}
\usetikzlibrary{calligraphy,shapes.callouts} 

\newcommand{\blackcircle}[1]{%
\tikz[baseline=-0.5ex]{\fill (0,0) circle (#1);}}

\usepackage[T1]{fontenc}

\usepackage[utf8]{inputenc}

\usepackage{microtype}
\usepackage{enumitem}

\usepackage{inconsolata}
\usepackage{booktabs} 
\usepackage{graphicx}
\usepackage{tabularx}
\usepackage{xspace}
 \usepackage{amsmath}
\usepackage{cleveref}
\usepackage[normalem]{ulem}
\definecolor{ao}{rgb}{0.0, 0.5, 0.0}
\definecolor{cyelllow}{HTML}{FFC300}
\definecolor{ablue}{HTML}{006795}
\definecolor{jgreen}{HTML}{98CC70}
\definecolor{xgbblue}{HTML}{006795}
\definecolor{dbertagreen}{HTML}{3a6a00}
\definecolor{cblue}{HTML}{4d4dff}
\newcommand{\JL}[2][]{\textcolor{black}{#2}}

\newcommand{\TODO}[1]{\textcolor{red}{\textbf{TODO:} #1}}
\newcommand{\clm}[0]{CLM\xspace}
\newcommand{\clms}[0]{CLMs\xspace}
\newcommand{\oorange}[0]{\textbf{$\textcolor{Orange}{\blackcircle{3pt}}$}}
\newcommand{\xorange}[0]{\textcolor{Orange}{$\boldsymbol{\mathsf{x}}$}}
\newcommand{\oblue}[0]{\textbf{$\textcolor{NavyBlue}{\blackcircle{3pt}}$}}
\newcommand{\xblue}[0]{\textcolor{NavyBlue}{$\boldsymbol{\mathsf{x}}$}}
\newcommand{\ogreen}[0]{\textbf{$\textcolor{ForestGreen}{\blackcircle{3pt}}$}}
\newcommand{\xgreen}[0]{\textcolor{ForestGreen}{$\boldsymbol{\mathsf{x}}$}}

\title{Probing Chemical Language Models: Effects of Pre-training and Fine-tuning}

\author{
  \textbf{Anna Karnysheva\textsuperscript{1,2}},
  \textbf{Dietrich Klakow\textsuperscript{1,2,3}}$^*$,
  \textbf{Ji-Ung Lee\textsuperscript{1}}\thanks{Co senior authors.}
\\
  \textsuperscript{1}RTG Neuroexplicit Models
  \textsuperscript{2}Spoken Language Systems,
  \textsuperscript{3}PharmaScienceHub (PSH)
\\
Saarland University
\\
    \href{mailto:akarnysheva@lsv.uni-saarland.de}{akarnysheva@lsv.uni-saarland.de}, \\
    \href{mailto:dietrich.klakow@lsv.uni-saarland.de}{dietrich.klakow@lsv.uni-saarland.de}, \\
    \href{mailto:ji-ung.lee@uni-saarland.de}{ji-ung.lee@uni-saarland.de}}

\begin{document}
\maketitle
\begin{abstract}

Chemical language models (\clms) are trained with linearized representations such as SMILES, yet it remains unclear which \textit{chemically meaningful} substructures they encode.
To foster a better understanding of \clms, we conduct a systematic study and probe for 78 molecular substructures across eight pre-trained and six randomly initialized models.
We furthermore study how fine-tuning on chemical downstream tasks affects the learned representations of molecular substructures.
Our results show that pre-training generally improves molecular structure awareness of \clms, particularly in the upper layers.
Moreover, randomly initialized models already encode ring structures well in the first layer.  
Our analysis on two chemical downstream tasks further reveals that, interestingly, fine-tuning affects task-relevant molecular substructures more than others, indicating that the changes in the representations follow chemical theory.\footnote{Code and data are available at: \url{https://github.com/basagashka/substructure-probing-clms}}  

\end{abstract}

\section{Introduction}\label{sec:introduction}

Drug discovery is an inherently expensive process that includes labor- and time-intensive steps such as designing molecules that are both effective against a target disease and can be safely administered to humans~\citep{JIA2020248}.
One important task in this process is molecular property prediction (MPP, \citealt{Shen-2019-MPP}), i.e., to reliably predict properties such as the lipophilicity, solubility, permeability, bioactivity, or toxicity of a molecule.\footnote{We provide an introduction into chemistry in \Cref{sec:appendix-background}.}

Many deep neural network (DNN) archi\-tectures---including general-purpose large language models (LLMs, \citealt{zhao2023what}), GNNs~\citep{Scarselli-2009-GNNs}, graph transformers~\citep{Yun-2019-GraphTransformer}, and sequence-based chemical language models (\clms, \citealt{Wang-2019-SmilesBert})---have been explored for MPP tasks, however, they still frequently fall behind feature-based models~\citep{Dias2023Limitations, dl_limitations_mpp, sadeghi2024moleval}.
Moreover, they often exhibit poor out-of-distribution generalization~\citep{ood_mpp} and are not evaluated on regression tasks which make up a substantial portion of MPP tasks~\citep{liu2025roft}. 
While a few works have tried to establish a better understanding of the shortcomings of DNNs by probing their representation for individual molecular substructures, they are often limited to a small set of molecular substructures and graph-based models which are often trained for individual MPP tasks~\citep{pgr, ESSL,volkov2022frustration}. 

In this work, we focus on \clms trained on linearized molecular representations (i.e., SMILES, \citealt{smiles}) which have been frequently used~\citep{chemberta3}, but not well studied, especially regarding whether they learn to capture molecular substructures during pre-training (\textbf{RQ1}) and how fine-tuning on chemical downstream tasks affects these representations (\textbf{RQ2}). 
Our systematic empirical investigation based on a new probing dataset comprising 78 molecular substructures evaluated across eight pre-trained (PT) and six randomly initialized (RI) models reveals that:
\begin{itemize}
    \item Pre-training improves molecular structure awareness towards the upper layers. Also, all molecular substructures exhibit a change larger than $\pm1\%$ in at least one model.
    \item RI models already encode ring structures well, but not other substructures.
    \item Some molecular substructures are unlearned in all models during pre-training.
\end{itemize}

Studying how fine-tuning on lipophilicity and solubility prediction affects the representations of molecular substructures reveals that:

\begin{itemize}
    \item Pre-training increases the robustness of molecular substructures during fine-tuning.
    \item Fine-tuning affects the representations less compared to pre-training with changes occuring more frequently in the upper layers.
    \item Molecular substructures that are theoretically more relevant for lipophilicity and solubility prediction are more affected by fine-tuning.
\end{itemize}

Finally, we explore how probing can be used to identify molecular substructures on which models may not have been sufficiently trained, and whether further pre-training on molecules containing these substructures can improve their representations.

\section{Related Work}\label{sec:related-work}

\begin{figure*}[htb]
    \includegraphics[width=1.0\linewidth,trim=0cm 9cm 0cm 3cm,
    clip]
{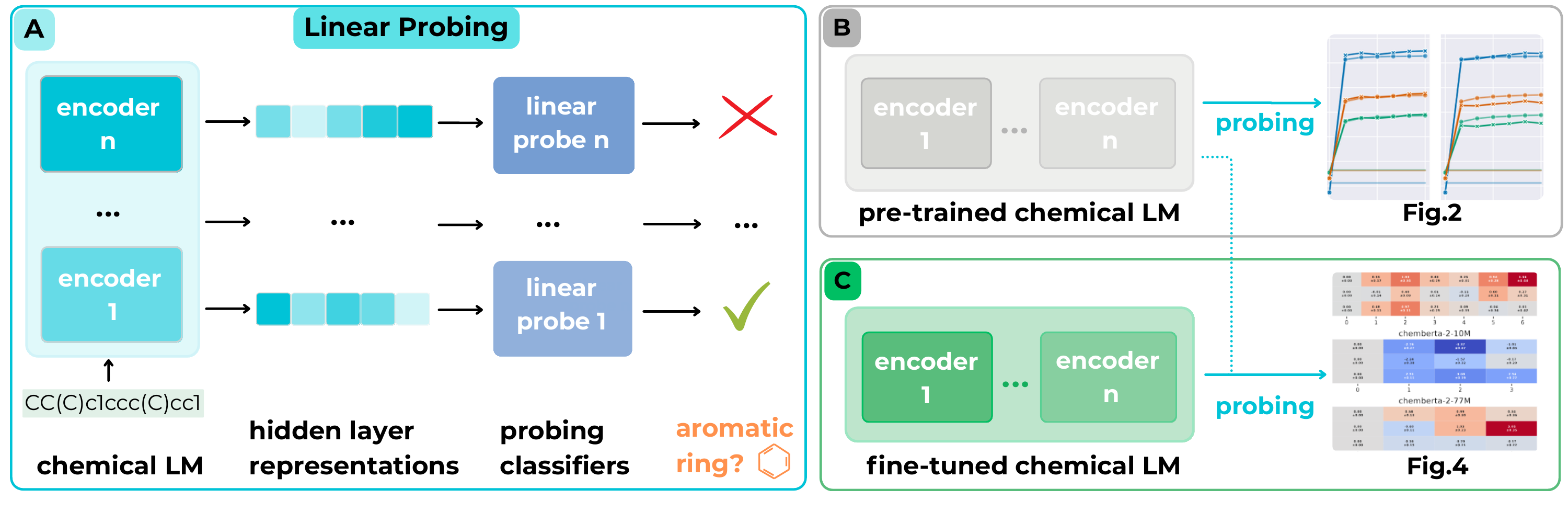}
\caption{Overview of the experimental setup. \textbf{\textcolor{SkyBlue}{A}} illustrates the general linear probing setup described in \S\ref{sec:experimental-setup} for a single probing task, here the presence of an aromatic ring. \textbf{\textcolor{Gray}{B}} and \textbf{\textcolor{Green}{C}} correspond to the two research questions investigated using this setup. RQ1 (\S\ref{sec:pretraining}) investigates the effect of pre-training on molecular substructure encoding, as shown in \textbf{\textcolor{Gray}{B}}. In contrast, RQ2 (\S\ref{sec:finetuning}) examines changes in molecular substructure encoding following fine-tuning by comparing the probing performance of pre-trained and fine-tuned models, as shown in \textbf{\textcolor{Green}{C}}.}
\label{fig:pipeline_overview}

\end{figure*}
\subsection{Molecular Representation Learning}

Molecules can be represented in various ways in order to be processed by models or hand-crafted algorithms.
The choice of representation directly affects what architectures are suited; e.g., representing molecules as graphs enables the use of different kinds of GNNs. 
To train language models, we use linearized molecule representations such as SMILES~\citep{smiles}. 
\JL[]{For chemistry, works have trained encoder-decoder, encoder-only, and decoder-only models.} 

\paragraph{Encoder-decoder models}
Works have utilized encoder-decoder models in tasks that mirror their sequence-to-sequence nature~\citep{moltransformer, Irwin_2022, luzhangt5}.
Example tasks are chemical reaction prediction (CRP), i.e., predicting the outputs for a given set of inputs~\citep{fooshee2018deep} and molecular optimization (MO), where an input molecule is altered to achieve desired properties~\citep{He2021MolecularOptimization}.  

\paragraph{Encoder-only models}
Encoder-only models (here, referred to as \clms) have primarily been trained for MPP.
To improve task performance, works have utilized different linear molecular representations~\citep{selfies, Yüksel_2023, Leon2024}, domain-specific auxiliary training objectives~\citep{MolBERT, chemberta2, kbert, li2021mol, park-etal-2024-moleco, park-etal-2024-moltres}, different tokenization schemes~\citep{chemberta, chemberta2, Leon2024}, positional encodings~\citep{ross2022large, molropebert} and attention mechanisms~\citep{ross2022large}. 

\paragraph{Decoder-only models}
While early works utilize decoder-only models to generate new molecules~\citep{xmol, molgpt, Wang2023_cMolGPT}, later studies consider them for other tasks such as MPP, CRP, and MO~\citep{He2022TransformerMolecularOptimization,taiga2023}.
Others have even devised novel text-centric tasks such as molecule captioning~\citep{edwards-etal-2021-text2mol} or augmented tasks with textual instructions~\citep{molt5, Wang2023_cMolGPT, MolXPT, pmlr-v202-christofidellis23a, fang2024molinstructions, lin2026attrilensmolattributeguidedreinforcement}. 


\paragraph{Limitations}
Despite all efforts to utilize LLMs in chemistry~\citep{Wang2023_cMolGPT, drugassist, xian-etal-2025-molrag}, recent works found that both chemical and general-purpose LLMs struggle to understand molecular structure~\citep{jang-etal-2025-structural, ganeeva-etal-2024-lost} or are outperformed by simple baselines~\citep{guo2023largelanguagemodelschemistry}.\footnote{Our experiments in \Cref{sec:appendix-baselines} support these findings.}
For instance, \citet{xian-etal-2025-molrag} show that GNNs outperform general-purpose LLMs on classification as well as regression tasks such as lipophilicity, which constitute a large portion of MPP tasks~\citep{liu2025roft}. 

\subsection{Probing}\label{sec:relwork-probing}
Probing is widely used in NLP to investigate the extent to which LMs capture linguistic knowledge, such as syntactic \citep{jawahar-etal-2019-bert, tenney2018what, liu-etal-2019-linguistic, hou-sachan-2021-birds} or semantic information~\citep{tenney2018what}. 
Typically, probing involves training a classifier for a specific probing task (e.g., part-of-speech tagging) using hidden representations extracted from a pre-trained LM~\citep{belinkov-2022-probing}. 

\paragraph{Probing for linguistic knowledge} 

Various works localize linguistic knowledge in pre-trained LMs (particularly BERT), attributing syntax and semantics to different layers~\citep{peters-etal-2018-dissecting, Tenney2019BERTRT, jawahar-etal-2019-bert, hewitt-manning-2019-structural}.
Others investigate the effect of fine-tuning, finding that changes are more centered around upper layers~\citep{mosbach-etal-2020-interplay, zhou-srikumar-2022-closer} and are task-dependent~\citep{merchant-etal-2020-happens}.
In general, works have found that many of these changes are less pronounced than during pre-training and vary from task to task.

\paragraph{Probing for chemical knowledge} 
Few works explore the encoding of molecular substructures in models. 
Prior works focus on graph-based models, finding that GTs generally encode ten molecular substructures better than message-passing GNNs; and that the molecular substructures are already encoded well by random initializations~\citep{pgr}. 
Others study differences between pre-training and fine-tuning of GNNs on MPP tasks and report a positive correlation between probing and MPP performance~\citep{ESSL}.
Only \citet{payne2020bertlearnsandteaches} and \citet{Fender2025} investigate \clms either for visualization or only for two molecular substructures.
Finally, some works have shown that even image-based models capture chemical and biological knowledge in their representations~\citep{alampara2025probing, naghdloo2025representation}.

In summary, it is not well understood whether \clms learn molecular substructures well and if this follows any chemical theory.
With this work, we make a first attempt to address this gap by conducting systematic probing experiments with \clms across 78  molecular substructures.




\section{Probing Dataset Creation}\label{sec:probingdataset}

Our goal is to curate a probing dataset that captures a wide range of molecular substructures and at the same time allows us to conduct meaningful analysis. 
Each probing task is formulated as a binary classification problem predicting the presence or absence of a molecular substructure (e.g., functional group, ring, etc.) in a molecule. 
The probing dataset is derived from PCQM4Mv2, a publicly available dataset designed for predicting the HOMO-LUMO energy gap~\citep{hu2021ogblsc}. 

\paragraph{Preprocessing}
We first discard all molecules with invalid SMILES strings or those that lead to processing errors (cf. \Cref{sec:probing_set_creation}). 
The remaining molecules---represented as SMILES strings---are then canonicalized and annotated with binary labels. 
Preprocessing results in an initial set of 101 unique molecular substructures.
All processing except for binarization is performed using RDKit~\citep{greg_landrum_2020_3732262}.

\paragraph{Data sampling and cleaning}
The PCQM4Mv2 dataset comprises $\approx\!3.7$ million molecules---too many to conduct extensive probing experiments. 
Hence, we create three subsets by uniformly randomly sampling 100k and 20k instances from the preprocessed train and validation splits of PCQM4Mv2, respectively. 
Note that sampling from the original splits prevents any leakage between train and test sets.
Finally, we discard probing tasks for molecular substructures that appear fewer than 200 times in either the train or test set.

Our final probing dataset comprises a diverse set of 78 molecular substructures---from functional groups such as amides or phenols to different types of ring structures (cf. \Cref{sec:probing_set_tasks}). 

\section{Experimental Setup}\label{sec:experimental-setup}

We investigate our research questions across eight pre-trained \clms, which we first probe for the presence of molecular substructures, comparing them against their randomly initialized counterparts.
We then study the effects of fine-tuning on two well-studied MPP tasks (lipophilicity and solubility prediction). 
This allows us to compare the changes of a model's molecular substructure representation against existing chemical knowledge. An overview of the experimental setup is provided in \Cref{fig:pipeline_overview}.

\paragraph{Pre-trained \clms}

We focus on models pre-trained on small molecules, particularly encoder-only \clms trained with the masked language modeling (MLM) objective. These models can consider both left and right context, in contrast to decoder-only models trained on causal language modeling.

\noindent
\textbf{Chemberta} \citet{chemberta} release multiple six-layer models based on RoBERTa~\citep{liu2020roberta}. 
We use both publicly available models, \texttt{chemberta-base} and \texttt{chemberta}.
\\
\textbf{Chemberta-2} In subsequent work, \citet{chemberta2} release models pre-trained on different numbers of molecules (\texttt{chemberta-2-5M}, \texttt{chemberta-2-10M}, and \texttt{chemberta-2-77M}). 
\\
\textbf{Chemberta-3} Most recently, \citet{chemberta3} released their training framework along with a 12-layer model trained on 100M molecules (\texttt{chemberta-3}).
\\
\textbf{Molformer} \citet{ross2022large} train a model with 12 layers using linear attention and rotary positional embeddings.
We use the publicly available model trained on 100M molecules (\texttt{molformer}). 
\\
\textbf{Roberta-zinc-480m} \citet{heyer2023roberta} release a 14-layer RoBERTa-based model trained on 480M molecules (\texttt{roberta-zinc-480m}).

All models are trained on molecules sampled from either ZINC15~\citep{ZINC15} or ZINC20~\citep{zinc20}. \texttt{molformer} is additionally trained on PubChem \citep{pubchem}. 
A more detailed description of all models is provided in \Cref{sec:app_model_selection}.

\paragraph{Probing setup}
For each probing task, we train a linear classifier on the CLS token representation of each encoder layer (cf. \Cref{sec:app_probing_setup}). 
We evaluate probing performance using the macro-averaged F1 score to account for class imbalances in the test splits. 
We compare each pre-trained model (PT) against its randomly initialized (RI) counterpart (note, \texttt{chemberta-2} has one shared RI model) and a majority class prediction baseline (maj). 
We further downsample instances in the training set to account for class imbalances. 
Preliminary experiments show that this substantially improves probing performance.
For the remainder of the paper, all reported results refer to the performance on the downsampled probing tasks. 
All dataset statistics are provided in \Cref{sec:probing_set_stats}.

\section{The Effect of Pre-Training (RQ1)}\label{sec:pretraining} 

\begin{figure*}[htb]
        \includegraphics[width=1.0\linewidth]
        {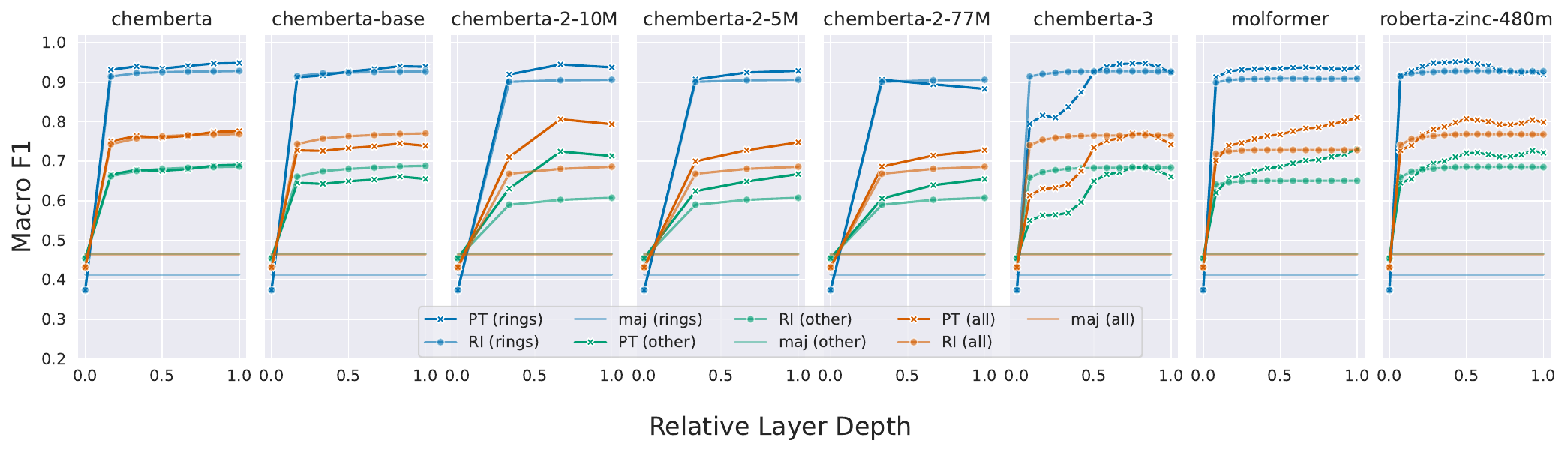}
    \caption{Average probing performance (macro-averaged F1 score $\uparrow$) of pre-trained (PT), randomly initialized (RI), and the majority class prediction (maj) models on molecular substructures.
    We report average performance on 12 ring types (avg rings), all other 66 substructures (avg other) and all 78 substructures (avg all).}
 \label{fig:bp}
\end{figure*}

We study the effect of pre-training (RQ1) using our probing dataset (\S\ref{sec:probingdataset}) and analyzing the results with increasing levels of granularity.
We first compare the probing performance of pre-trained (PT) and randomly initialized (RI) models (\S\ref{sec:pre-training-probing}), then with respect to different ring structures (\S\ref{sec:pre-training-rings}), and, finally, for individual molecular substructures (\S\ref{sec:pre-training-molecules}).

\subsection{Probing Results}\label{sec:pre-training-probing}
\Cref{fig:bp} shows the macro-averaged F1 score of all eight PT and six RI models averaged across all 78 probing tasks and three datasetes each. 
In addition, we show the average performance of the majority class prediction baseline (maj).
On average, models \textbf{learn to better encode molecular substructures during pre-training}: most pre-trained models (\xorange)~exhibit substantial improvements in performance relative to their randomly initialized counterparts (\oorange) and the majority classifier (\textcolor{orange}{--}). 
We also observe that for most PT models (excluding \texttt{chemberta} and \texttt{chemberta-base}), the probing performance is \textbf{higher in the upper layers} (0.5--1.0). 
In particular, we see that \texttt{chemberta-2-10M}, \texttt{molformer} and \texttt{chemberta-2-5M} benefit the most from pre-training. 
In contrast, we observe negligible improvement for \texttt{chemberta} or even small drops for \texttt{chemberta-base}.
Most notably, we find that in the most recent model (\texttt{chemberta-3}), probing performance deteriorates substantially in the lower layers.
We further investigate this phenomenon in \Cref{sec:appendix-chemberta-3} by further pre-training the model on different datasets.



\subsection{Ring Structures}\label{sec:pre-training-rings}
We further analyze the representations of rings and other molecular substructures, finding that the representations of both RI (\oblue) and PT (\xblue) models perform exceptionally well at identifying ring structures compared to all other groups (\textbf{\textcolor{ForestGreen}{$\blackcircle{3pt}$}} and \xgreen). 
Moreover, RI models \textbf{encode ring structures well already at the first encoder layer}, suggesting that these surface-level patterns are easy for the models to extract directly from the input, even without pre-training. 
We also find that the benefit of pre-training diminishes for ring structures compared to that of other molecular substructures (i.e., the gap between \xblue~and \oblue~is much smaller compared to the gap between \xgreen~and \ogreen).

\paragraph{Different types of rings}
A closer analysis of different types of ring structures (i.e., aliphatic, aromatic, and saturated rings) reveals that particularly aromatic and aliphatic are already well encoded in random initializations, with pre-training slightly reducing probing performance in the upper layers (cf. \Cref{fig:performance-rings}) for all models except for \texttt{chemberta-3} (see \Cref{sec:appendix-chemberta-3}).
The high performance on aromatic rings might stem from a distinct surface-level pattern.
When SMILES strings are canonicalized, atoms in aromatic rings (see \Cref{fig:example_molecule} for an example) are represented with lowercase letters.
This contrast to other molecular substructures (which consist of upper-cased atoms)  results in a strong signal for the model.

\subsection{Individual Molecular Substructures}\label{sec:pre-training-molecules}
\begin{figure*}[t]
  \centering
  \includegraphics[scale=0.3]{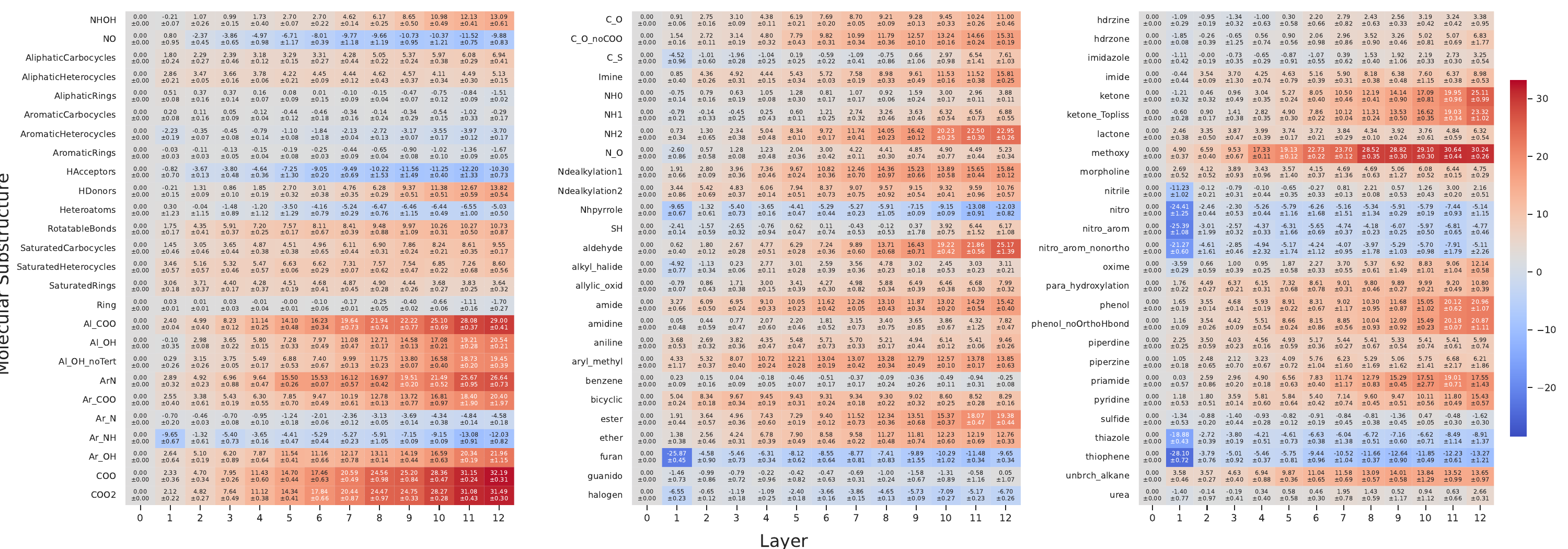}
  \caption {Relative difference in probing performance (\% macro-averaged F1) between RI and PT \texttt{molformer} on 78 molecular substructures (cf. \Cref{tab:probing_train_sizes}). Each cell denotes the relative difference for a specific probing task (y-axis) and layer (x-axis). \textcolor{red}{Red} indicates an \textcolor{red}{increase} in performance after pretraining, while \textcolor{blue}{blue} denotes a \textcolor{blue}{decrease}.} 
  \label{fig:delta_bp}
\end{figure*}

Finally, we investigate if there are molecular substructures that undergo changes consistently across all models. 
For visualization, we focus on the model with the most pronounced changes (\texttt{molformer}) and provide the rest in \Cref{sec:pt_effect_apx}.
We observe three patterns shown in \Cref{fig:delta_bp}.

\paragraph{\textcolor{red}{PT > RI}} 
First, we find that \textbf{pre-training generally leads to a better encoding of most molecular substructures in upper layers}, as reflected by the higher density of the red shade in the middle and upper layers.~In particular, all models exhibit substantial improvement on 
\emph{carboxylic acids} (\texttt{COO}, \texttt{COO2}, \texttt{Al\_COO}, 
\texttt{Ar\_COO}), 
\emph{aromatic hydroxy groups} (\texttt{Ar\_OH}), 
\emph{phenol groups} (\texttt{phenol}, \texttt{phenol\_nonorthobound}),  \emph{amides}, and 
\emph{ketones} (\texttt{ketone}, \texttt{ketone\_Topliss}).~Furthermore, all models improve on \emph{carbonyls} (\texttt{C\_O\_noCOO}), 
\emph{aldehyde} and \emph{imide}, although to a lesser degree.

\paragraph{\textcolor{blue}{PT < RI}} 
Second, \textbf{some molecular substructures consistently exhibit lower performance after pre-training}. 
In particular, probing performance decreases for aromatic nitrogens (\texttt{Ar\_N}) and certain \emph{heterocycles} such as \texttt{thiazole}, \texttt{thiophene} and \texttt{furan} across all models. 
Likewise, some ring substructures such as \texttt{AromaticHeterocycles}, \texttt{AromaticCarbocycles} and \texttt{benzene} show slight performance degradation in upper layers.
The decrease in performance on \texttt{AromaticHeterocycles} may potentially reflect the substantial drops in performance of \texttt{thiazole}, \texttt{thiophene} and \texttt{furan}---all aromatic heterocycles.
Finally, most models (except for \texttt{chemberta-2-5M/10M}) show evidence of unlearning \texttt{halogens} in upper layers \JL[, resulting in better encoding in low encoder layers.]{while preserving a better encoding in the lower layers.}

\paragraph{\textcolor{Gray}{PT $\approx$ RI}} 
Third, we observe that only a handful molecular substructures undergo small amounts of change ($\pm1\%$), resulting in an almost uniform distribution of information across layers. However, this behavior is \textbf{not consistent across models}.





\subsection{Discussion}
Overall, our results suggest that \textbf{pre-training improves the molecular structure awareness} of \clms, considerably changing the encoding of many molecular substructures. 
We further observe that \textbf{RI models already encode aromatic and aliphatic rings very well} performing on-par with or better than PT models. 
Interestingly, unlearning of molecular substructures largely varies between models, however, a \textbf{few molecular substructures are consistently unlearned during pre-training}.
We conjecture this might stem from a disparity in the pre-training data and conduct further pre-training experiments for five molecular substructures and across different models (\S\ref{sec:takeaways}). 
Finally, we find that \textbf{all molecular substructures exhibit a change larger than $\pm$1\% in at least one model}.

\section{The Effect of Fine-Tuning (RQ2)}\label{sec:finetuning} 

Our probing experiments have shown how pre-training reconfigures the information encoded in representations and that RI models already encode ring structures well.
Next, we investigate changes of RI and PT models during fine-tuning on two well-studied tasks in chemistry, which allows us to contextualize our findings within chemical theory.  


\subsection{Chemistry Background}
We focus on predicting the lipophilicity and aqueous solubility (in short, solubility) of molecules. 
Here, we provide brief task descriptions and introduce important molecular substructures that affect the lipophilicity and solubility of a molecule; and refer to \Cref{sec:appendix-background} for more details. 

\paragraph{Lipophilicity} 
Lipophilicity refers to the ability of a chemical compound to dissolve in fat-like solvents (lipids, fats, oils; \citealt{ijms24086970}). 
It is an important physicochemical property of molecules which correlates with the (oral) absorption, (tissue) distribution, metabolism, excretion, and toxcicity (ADMET) properties of drugs~\citep{MANNHOLD2009861}, essential in determining how a candidate drug will interact with the human body~\citep{WARING20092844}. 
The goal of lipophilicity prediction is to estimate the octanol/water distribution coefficient (logD) of a specific molecule.

\paragraph{Aqueous solubility}
Aqueous solubility refers to the ability of a molecule to dissolve in water. 
For drug development, predicting the solubility of a molecule is equally important as predicting the lipophilicity as it also affects their biovailability and ADMET profiles~\citep{llompart2024will, aq_sol_importance}. 
The goal of solubility prediction is to estimate the log solubility (logS) of a specific molecule in water. 
While solubility is closely related to lipophilicity, it is also dependent on other factors such as the melting point of a molecule~\citep{HILL2010648}. 

\paragraph{Important molecular substructures}
Chemical literature distinguishes between two groups of molecular substructures that are known to affect lipophilicity and solubility~\citep{alma99901018739001842}.
First, \emph{hydrophilic} substructures such as carboxylic acids substantially decrease a molecule's lipophilicity while increasing its solubility.
Second, \emph{lipophilic} substructures such as aromatic rings increase a molecule's lipophilicity while decreasing its solubility. 
We follow this classification of molecular substructures in our analysis and put all other molecular substructures that do not substantially affect lipophilicity into a third group (\emph{other}). 
We provide a list of all molecular substructures along with their group in \Cref{sec:app_ft_setup}. 

\subsection{Experimental Setup}
\label{subsec:ft_setup}
For fine-tuning, we replace the classification head of the \clm with either a linear regression layer or a two-layer MLP and minimize the mean squared error loss.
Following \citet{wu2018moleculenetbenchmarkmolecularmachine}, we use the root mean squared error (RMSE) as our evaluation metric for both tasks.
Since all models were pre-trained on canonicalized SMILES strings, we canonicalize the input SMILES accordingly.


\paragraph{Dataset} 
Both datasets are sampled from  the MoleculeNet benchmark~\citep{wu2018moleculenetbenchmarkmolecularmachine} and consist of 4,200 (lipophilicity) and 1,127 (solubility) molecules.  
We use the train–validation–test splits (80/10/10) provided by \citet{ross2022large}. 
Detailed dataset statistics and analysis for both tasks are provided in \Cref{sec:app_ft_setup}. 

\paragraph{Hyperparameters}
We perform hyperparameter tuning separately for both tasks, considering different batch sizes and learning rates.
All pre-trained and randomly initialized models are trained for up to 10--20 epochs.
We deploy early stopping with a patience of 2 and use AdamW as our optimizer.
We report all hyperparameters in \Cref{sec:app_ft_setup}.

\paragraph{Baselines}
As baselines, we evaluate multiple traditionally used models, namely, linear regression models (LR), support vector machines (SVM), and gradient boosted trees (XGB).
For each model, we evaluate four algorithms to extract molecule representation vectors, also known as fingerprints, provided by RDKit~\citep{greg_landrum_2020_3732262}.
Finally, we evaluate two large language models (LLMs): \texttt{Llama-3.2-3B-Instruct}~\citep{dubey2024llama} and \texttt{gpt-oss-20B}~\citep{openai2025gptoss120bgptoss20bmodel} with additional chemical knowledge that is important for the respective downstream task. 
We provide detailed hyperparameters and experimental results for all baselines in appendices~\ref{sec:appendix-fingerprints} and \ref{sec:appendix-LLM}.

   

\paragraph{Probing dataset adjustment}
In order to conduct meaningful analyses, we accommodate changes to the probing dataset introduced in \S\ref{sec:probingdataset} that consider dataset-specific properties of the respective downstream task.
More specifically, we discard any molecular substructure which appears fewer than ten times in either the training or test split of the task-specific dataset; effectively removing outliers from our analysis.
This results in probing 60 and 39 molecular substructures for lipophilicity and solubility prediction, respectively.


\subsection{Downstream Task Results} 
\Cref{tab:ft_results} shows the results of all randomly initialized (RI) and pre-trained models (PT) as well as the best performing model using fingerprints (SVM) and LLM (\texttt{gpt-oss-20b}) for both downstream tasks.
We further include the results of the graph-based models ($^\text{mol}$) that were reported by \citet{ross2022large} who use the same data splits.
Overall, we observe that PT models consistently outperform RI ones on both tasks with \texttt{molformer} consistently performing best, highlighting the benefit of pre-training \clms.  
We further find that the \texttt{chemberta-2} models, differing only in the pre-training datasets, exhibit differences of 0.073 RMSE on lipophilicity (0.046 on solubility), suggesting that pre-training data plays a major role for downstream task performance.
Moreover, the \texttt{chemberta-2-77M} model is often outperformed by its smaller counterparts.
This indicates that data quality may play a more important role than data quantity. 
Finally, consistent with prior findings, we observe that fingerprint-based models perform rather well~\citep{Dias2023Limitations, dl_limitations_mpp}; and that LLMs perform even worse than the mean predictor~\citep{zhao2023what}.

\begin{table}[htb]
\small
  \centering
  \begin{tabular}{lcccc}
    \toprule
    \textbf{Model} &  \multicolumn{2}{c}{\textbf{Lipo}}  & \multicolumn{2}{c}{\textbf{ESOL}}\\
     & \textbf{RI}  &\textbf{PT} & \textbf{RI}  &\textbf{PT}\\
    \midrule
    GC$^\text{mol}$  & -& 0.655 & -& 0.970 \\
    A-FP$^\text{mol}$ &-& 0.578 & -& 0.503\\
    MPNN$^\text{mol}$ & - & 0.719 & - & 0.580 \\
    \midrule
    \texttt{mean predictor}         & -            &    1.013  &   -            &    2.057   \\
    SVM$_\text{(c=16)}$ + ATFP & - & 0.640  & - & 0.830 \\
    \texttt{gpt-oss-20b} & - & 2.532 & - & 8.964 \\ 
    \midrule
\texttt{molformer}       & 0.832              & \textbf{0.565}  &      0.808       &   \textbf{0.587} \\ 
\texttt{roberta-zinc-480m}& 0.788               & 0.580  &        0.878       &    0.746          \\
\texttt{chemberta-base}   & 0.785              & 0.663   &       0.832        &        0.739     \\
\texttt{chemberta}        & 0.779              & 0.675 &    0.822         &             0.693 \\
\texttt{chemberta-2-5M}       & 0.850               & 0.664   &     0.872           &  0.682\\
\texttt{chemberta-2-10M}        & "               & 0.591    & "               &    0.724        \\
\texttt{chemberta-2-77M}   &  "               & 0.632    &  "               &         0.728   \\
\texttt{chemberta-3}      & 1.026& 0.637  &       0.960         &     0.757    \\
\bottomrule
  \end{tabular}
  \caption{Test performance (RMSE, $\downarrow$) for lipophilicity (Lipo) and solubility (ESOL) prediction \citep{wu2018moleculenetbenchmarkmolecularmachine}. Besides all \clms, we also include results of the best performing fingerprint-based model (SVM) and LLM (\texttt{gpt-oss-20b}). $^\text{mol}$ denotes results of graph-based models reported by \citet{ross2022large}. } 
  \label{tab:ft_results}
\end{table}

\begin{figure*}[htb]
  \begin{subfigure}[t]{0.5\textwidth}
        \centering
         \includegraphics[width=1.0\linewidth]
         {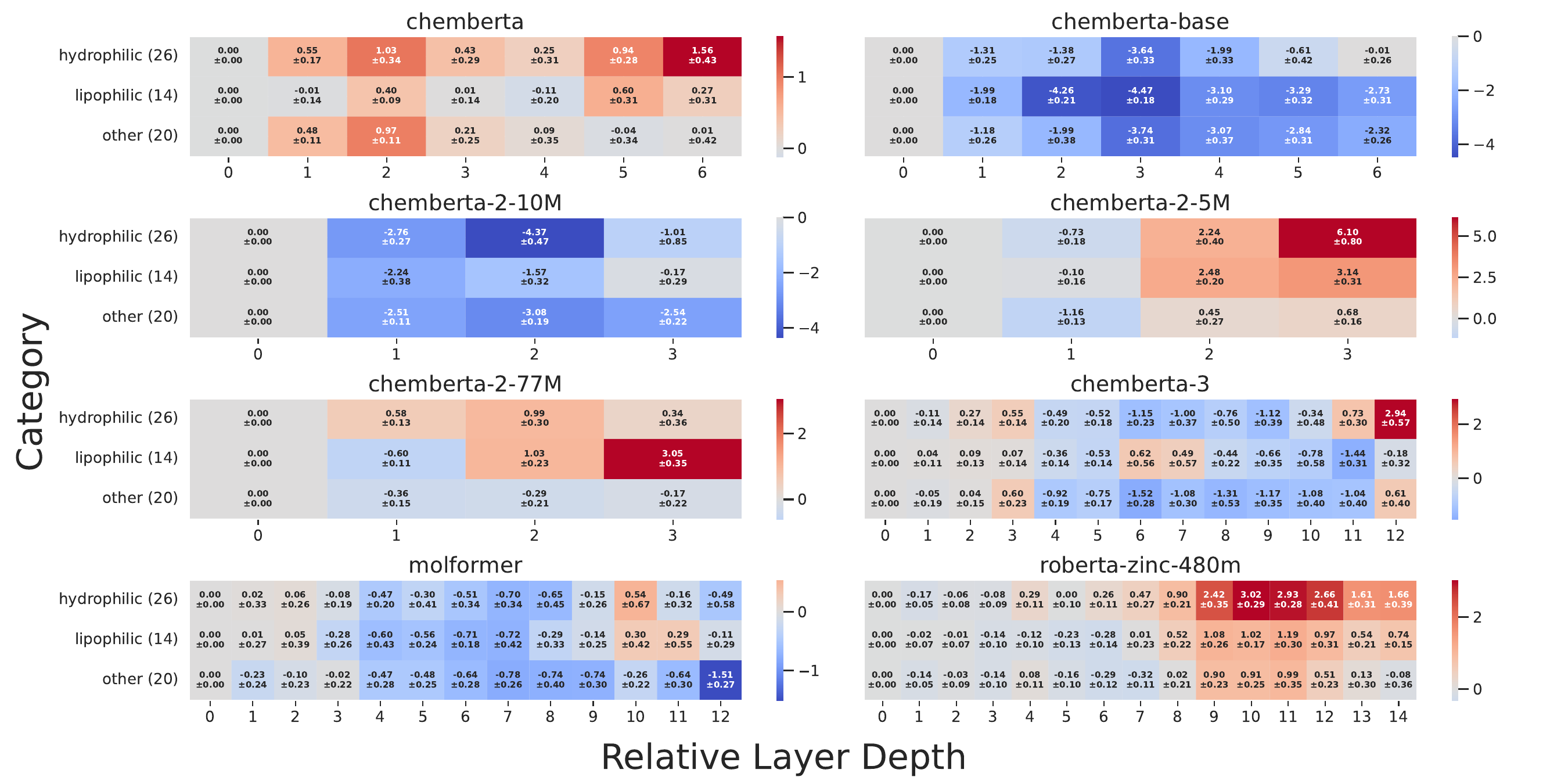}
    \end{subfigure}
  \hfill
  \begin{subfigure}[t]{0.5\textwidth}
        \centering
         \includegraphics[width=1.0\linewidth]
         {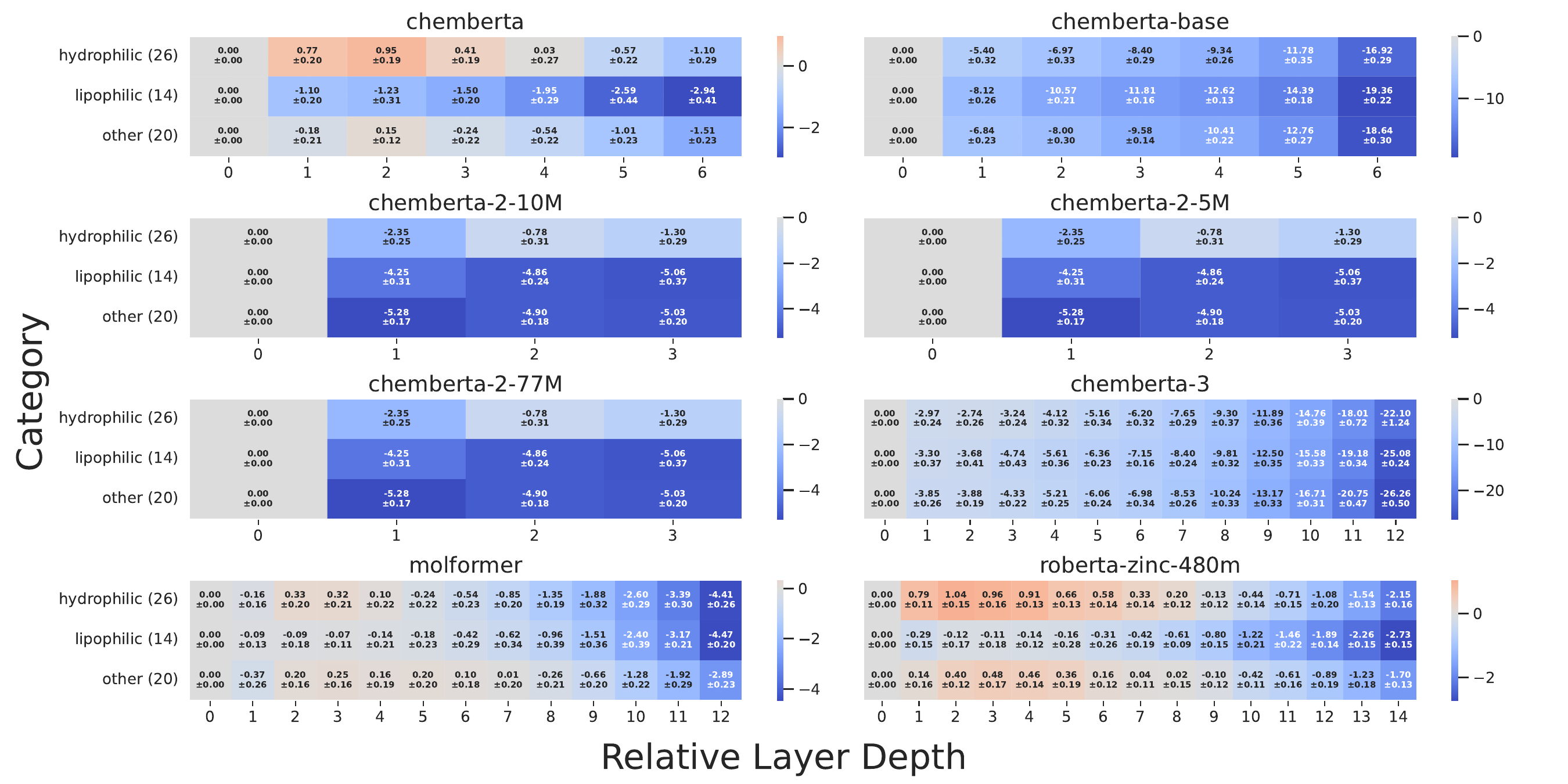}
    \end{subfigure}
    


    
    \caption{Average relative differences in probing performance (\% macro-averaged F1) in probing performance 
    of PT (left) and RI (right) models after fine-tuning on lipophilicity. We group into \emph{hydrophilic} (top), \emph{lipophilic} (middle), and \emph{other} (bottom) groups (see \Cref{tab:lipo_stats}), with numbers indicating the group size. 
    The RI results for \texttt{chemberta-2} are based on a single model (hence, are the same for all three variants) as all models use the same architecture. }
    \label{fig:avg_delta}
\end{figure*}

\subsection{Probing Results} 
\label{sec:ft_probing_results}
We conduct probing experiments similar to \S\ref{sec:pretraining} but with the difference that we now compare the layer-wise representations of molecular substructures in a model before and after fine-tuning.
This is done for both RI and PT models to understand potential differences in their behavior during fine-tuning.
In our analysis, we first inspect lipophilic and hydrophilic molecular substructures and then inspect individual molecular substructures.
Due to a lack of space, we focus our analysis in the main paper on lipophilicity prediction and provide the results and analysis for solubility prediction in \Cref{sec:ft_effect_apx}. 


\paragraph{Group analysis}
\Cref{fig:avg_delta} shows heatmaps for eight pre-trained (left) and six randomly initialized (right) models split into hydrophilic (top), lipophilic (middle), and other (bottom) groups. 
Similar to \S\ref{sec:pretraining}, \textcolor{red}{red} indicates an \textcolor{red}{increase} in probing performance while \textcolor{blue}{blue} indicates a \textcolor{blue}{decrease}.
We observe that the groups that are important for lipophilicity prediction (\textit{hydrophilic} and \textit{lipophilic}) undergo larger changes than the \textit{other} group. 
This indicates that \textbf{molecular substructure learning follows chemical theory}.
Interestingly, we find that \textbf{fine-tuning has a noticeably smaller effect than pre-training} on the molecular substructure representations; and that the changes are mostly \textbf{concentrated in upper layers} which corroborates prior observations in the NLP literature~\citep{mosbach-etal-2020-interplay, merchant-etal-2020-happens, fayyaz-etal-2021-models}.
In contrast, \textbf{RI models behave differently}, as fine-tuning appears to mostly negatively affect the encoding of substructures in upper layers. 

\paragraph{Individual analysis}
A detailed analysis of individual molecular substructures reveals that the effect of fine-tuning varies across models. Furthermore, even among molecular substructures of the same group (lipophilic, hydrophilic, other), the magnitude may vary (we provide detailed heatmaps in \Cref{sec:ft_effect_apx}). 
Nevertheless, there are multiple substructures for which probing performance increases after fine-tuning on lipophilicity prediction.
\JL[We observe that fine-tuning further improves probing performance of pre-trained models on]{These are primarily} \emph{carboxylic acids} (\texttt{COO}, \texttt{COO2}, \texttt{Al\_COO},
and \texttt{Ar\_COO}\footnote{With the exception of \texttt{chemberta2-77M}.}). 
This is \textbf{consistent with observations made by chemists suggesting that carboxylic acids contribute most negatively to the logD value} and are therefore highly indicative~\citep{logdcontrib}.
Interestingly, \texttt{chemberta-2} models consistently improve upon \emph{halogens} after fine-tuning (we study this closer in \S\ref{sec:takeaways}).
Again, we do not observe any consistent trends for RI models, except for a degradation of \texttt{AromaticHeterocycles} and \texttt{Ar\_NH}. 

\subsection{Discussion}
Our probing experiments on models before and after fine-tuning on lipophilicity (\S\ref{sec:ft_probing_results}) and solubility (\Cref{sec:ft_effect_apx}) prediction reveal three major findings.
First, \textbf{molecular substructures that are theoretically more relevant for a downstream task undergo larger changes during fine-tuning}.
Reciprocally, molecular substructures that undergo major changes during fine-tuning for a specific downstream task might indicate a high importance.
This might be especially interesting for tasks with a high variability in terms of important molecular substructures such as toxicity prediction.
Second, \textbf{changes are less pronounced compared to pre-training and occur more frequently in the upper layers}.
Considering the increasing model sizes and consequently, the increasing costs of probing, one way to reduce costs could be to restrict probing to the upper layers as they yield the largest changes. 
Third, \textbf{pre-training increases the robustness of molecular substructures in \clms}, making them more likely to be retained during fine-tuning.

\section{Practical Implications}\label{sec:takeaways}

The varying probing performance across different models (especially for the \texttt{chemberta-2} models with different pre-training data sizes, cf.~\S\ref{sec:pretraining}) suggests that this might be attributed to a lack of molecules containing a specific molecular substructure during pre-training.
To better understand the impact of pre-training data, we further pre-train the models.
In particular, we curate pre-training datasets consisting of molecules with molecular substructures on which a model underperforms.
Our results show that further pre-training on these molecules does indeed improve a model's internal representation of them (\Cref{sec:appendix_takeaway}).
Moreover, we find that models that share the same architecture converge towards the same upper bound shape in terms of probing performance (\Cref{fig:app_further_pt_bp_performance_halogens}). 

Similarly, we conduct experiments for the  \texttt{chemberta-3} model which exhibits a strange dome in the upper layers.
Further pre-training the model on two different datasets indicates that, again, this dome becomes less pronounced as probing performance also increases in the lower layers (cf. \Cref{fig:app_cb3_further_pt_plots}). 
Most notably, we find that the pre-training dataset can make a substantial difference on the resulting probing performance. 

These experiments provide an exploratory demonstration of how probing may be used to identify signs of undertraining for specific molecular substructures. However, our limited-scale experiments also show that improvements in the representations of individual substructures do not necessarily translate into improved predictions on lipophilicity (cf. \Cref{tab:ft_cb_halogens_ft_performance,tab:ft_thiazole_thiophene_furan,tab:ft_cb_phenol_ft_performance}). 

\section{Conclusion}\label{sec:discussion}

Our results provide a systematic analysis of the extent to which chemical language models (\clms) trained on linearized molecular representations encode important molecular substructures. Across eight pre-trained and six randomly initialized models, we find that pre-training generally improves molecular structure awareness, with the most pronounced effects emerging in the upper layers. 
Although certain molecular substructures are unlearned during pre-training, only a small subset exhibits this behavior consistently across all models. 
Notably, we find that ring structures are linearly decodable from representations of randomly initialized models, suggesting that these are surface-level properties of the input representations and do not require pre-training.
Our fine-tuning analysis reveals that, on average, groups of molecular substructures relevant to lipophilicity and solubility prediction 
undergo larger representational changes than others. Extending the analysis to other molecular property prediction tasks, such as toxicity, bioactivity, or permeability, represents an important direction for future work.

Finally, our further pre-training experiments showcase how probing may be used to identify potential signs of undertraining and mitigate the resulting representational gaps using a small and carefully curated dataset.


\section{Limitations}
\paragraph{True complexity of lipophilicity and solubility prediction} 
Although \citet{hansch1977substituent} estimate the lipophilicity of a molecule using a linear combination of its substructures, we note that this approach is an oversimplification of the underlying process. 
The actual lipophilicity of a molecule also depends on the environment a certain substructure is in (i.e., other neighboring substructures) as well as its depth and position in the molecule and is subject to further research in chemistry.

While the logD value (measuring the lipophilicity) is part of the general solubility equation (corrected for ionization at pH 7.4), there are various other factors that influence the logS value~\citep{HILL2010648}. 
Some of these challenges are part of ongoing chemistry research as highlighted by \citet{llompart2024will}, who find that accurately predicting aqueous solubility requires the knowledge of many factors such as the solid-solvated phase transition, solid state, temperature, polymorphism, intermolecular interactions between solute-solvent etc.


\paragraph{Other downstream tasks}
Our experiments focus on lipophilicity and solubility prediction as the downstream tasks.
While there exist other tasks such as predicting the bioactivity and toxicity of molecules, one limiting factors is the number of publicly available molecules that have been studied with the same experimental conditions.
Moreover, many tasks are still subject to ongoing chemical research and are not understood well (yet). 
For instance, a non-trivial challenge in predicting the bioactivity are activity cliffs~\citep{stumpfe2019evolving}, pairs of molecules with highly similar structures---i.e., close proximity in the molecular ``landscape''---but different magnitudes in terms of bioactivity, resulting in a steep ``cliff''. 
A better chemical understanding of the underlying process would allow researchers to build models that capture fine-grained structural differences between molecules.

\paragraph{Effects of random- and downsampling}
We note that due to the random sampling of the three probing datasets and the downsampling for individual molecular substructures, the molecules across different splits may vary with each molecular substructure dataset corresponding to a distinct subset of the original 100k molecules, both in terms of size and composition.

\paragraph{Limitations of probing} 
Probing remains a debated diagnostic method as it does not indicate whether a feature is used during prediction, but only how extractable the property is from the learned representations. 
Using linear probes (as in this work), we can therefore only assess the linear separability of the investigated property in these representations.
Note, that there is no consensus on what probing classifier to use. 
While some works argue that utilizing linear classifiers prevents the possibility of memorization~\citep{belinkov-2022-probing}, others propose to instead consider memorization during evaluation \citep{pimentel-etal-2020-information}.

\paragraph{More complex probing task} 
Our probing tasks focus on detecting the presence of a molecular substructure, rather than counting occurrences or identifying its location.
Consequently, some probing tasks may be easier to learn due to the underlying nature of a molecule.
For instance, a molecule may contain multiple occurrences of a single molecular substructure, making it easier to detect its presence (as is the case for aromatic rings that often occur multiple times in many molecules).
More challenging tasks could offer additional insights but are subject to future investigation; and moreover, also require respective chemistry research.






\section{Impact Statement}\label{sec:impact-statement}
The primary goal of this work is to provide insights on how chemical language models are affected by pre-training and fine-tuning on chemical data. 
While this work falls under the category of fundamental research without direct implications on downstream applications, the authors acknowledge that some of the findings (e.g., fine-tuning mostly aligns with chemical theory) may lead to research which could be abused to identify molecular substructures that are harmful to the human body.
The authors emphasize that the experiments and tasks presented here have no direct connection to harmful applications (including tasks such as predicting the toxicity of molecules).

\section*{Acknowledgments}
We thank Shubham Dokania, Max Rausch-Dupont and Afnan Sultan for their helpful discussions and feedback. 
This work was funded by the Deutsche Forschungsgemeinschaft (DFG, German Research Foundation) -- GRK 2853/1 “Neuroexplicit Models of Language, Vision, and Action” - project number 471607914.
\bibliography{custom}

\appendix

\section{Chemical Background for Molecular Modeling}
\label{sec:appendix-background}
\subsection{Drug Development}

Drug development aims at identifying molecules that are both effective against a target disease and can be safely administered to humans~\citep{JIA2020248}. 
One crucial bottleneck in drug development is synthesizing potential molecules in the laboratory, which is a time- and cost-intensive process. 
Machine learning offers the potential to accelerate this process by improving efficiency and reducing costs: for example, by prioritizing candidate compounds with desirable properties, thereby reducing the need to synthesize nonviable molecules. 
Consequently, \emph{molecular property prediction}---the estimation of physicochemical, biological and functional properties such as lipophilicity, toxicity, permeability, and reactivity---constitutes a core component of this process \citep{Waring2015Attrition, doi:10.1021/acs.jmedchem.5b00104}. 
In particular, \emph{ADMET} properties, which characterize the \emph{drug-likeness} of a compound, are essential for assessing how a drug candidate will interact with the human body. 
In this work, we focus on small molecules (those with molecular weight $\leq$ 1000 Da), which possess better ADMET profiles~\citep{BECK20221560} and constitute most approved pharmaceuticals \citep{MAKURVET2021100075}. 


\paragraph{Molecular property prediction}
Typically, molecular property prediction encompasses a wide range of tasks such as predicting bioactivity, solubility, permeability, and toxicity which are often addressed by different models.
It is important to note that many of these tasks are not classification, but regression tasks.
\JL[Accurate molecular property prediction critically depends on structural awareness of models \textcolor{red}{cite}, as]{Reliably predicting the magnitude of even a single property can be challenging, as} the arrangement and composition of molecular substructures \JL[ directly influence these properties]{play an important role}. 
In many cases, models must capture fine-grained structural differences between molecules that lead to considerably different behavior. One such example is activity cliffs~\citep{stumpfe2019evolving}, which describe pairs of molecules with highly similar structures---i.e., close proximity in the molecular ``landscape''---but different magnitudes in terms of bioactivity, resulting in a steep ``cliff''. 
Activity cliffs have been subject to chemistry research for over a decade, with evolving insights on what molecular substructures constitute them~\citep{maggiora2006outliers, stumpfe2019evolving, dl_limitations_mpp}.

\subsection{Linearized Representations} 
\begin{figure}[htb]
    \centering
    
    \includegraphics[width=\linewidth]
    {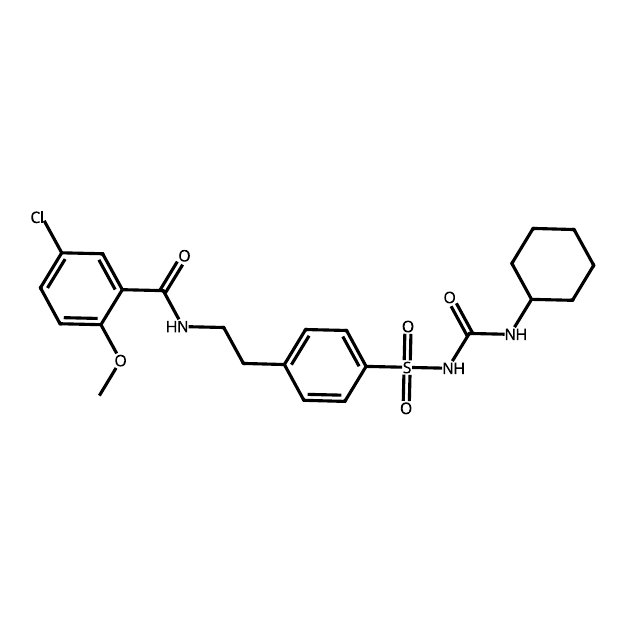}


    \small
    \begin{minipage}{\linewidth}
    \scriptsize
    \raggedright
    \ttfamily
    
    CO\textcolor{ForestGreen}{c1ccc}(Cl)\textcolor{ForestGreen}{cc1}%
    C(=O)NCC%
    \textcolor{orange}{c2ccc}(S(=O)(=O)NC(=O)N%
    \textcolor{purple}{C3CCCCC3})%
    \textcolor{orange}{cc2}
    
    \end{minipage}

    \caption{An example molecule: Glibenclamide and its SMILES representation. The two aromatic rings are highlighted in \textcolor{ForestGreen}{green} and \textcolor{orange}{orange}. Compared to the non-aromatic ring (\textcolor{purple}{purple}), all atoms in the aromatic rings are lowercased.}
    \label{fig:example_molecule}
\end{figure}

In order to process molecules in models or hand-crafted algorithms, works have devised various methods.
One such method called SMILES proposes to represent molecules as linearized representation of molecular graphs~\citep{smiles}. 
The standard SMILES encoding is non-unique, resulting in a one-to-many mapping between a molecule and its possible SMILES representations. However, canonicalization of SMILES enforces a deterministic graph traversal order, resulting in a one-to-one mapping between molecular graphs and SMILES strings.
SMILES encode some graph structural information explicitly. For example, double bonds are represented by $=$, triple bonds by \#. Single bonds, on the other hand, are not explicitly specified. Thus, consecutive atoms are assumed to be connected by either a single or an aromatic bond.
Furthermore, as shown in \Cref{fig:example_molecule}, when SMILES are canonicalized, atoms in aromatic rings (i.e., rings with alternating single and double bonds) are represented with lowercase letters.

\subsection{Lipophilicity Prediction}\label{sec:lipo_prediction}

Lipophilicity refers to the ability of a chemical compound to dissolve in fat-like solvents (lipids, fats, oils; \citealt{ijms24086970}). 
It is an important physicochemical property of molecules which correlates with the (oral) absorption, (tissue) distribution, metabolism, excretion, and toxcicity (ADMET) properties of drugs~\citep{MANNHOLD2009861}, essential in determining how a candidate drug will interact with the human body~\citep{WARING20092844}. 
\JL[However, for most experimentally tested molecules logP or logD values are not available, underlining the need for models that are able to accurately estimate these values]{Compared to other MPP tasks such as toxicity prediction, the impact of specific molecular substructures is better understood for lipophilicity prediction. For instance, \citet{logdcontrib} study how specific molecular substructures affect the lipophilicity of a molecule}. 

The goal of lipophilicity prediction is to estimate the octanol/water distribution coefficient (i.e., logD at pH 7.4) of a specific molecule.
Chemical literature distinguishes between two groups of molecular substructures that are known to affect lipophilicity~\citep{alma99901018739001842}. 
In particular, \emph{hydrophilic}
substructures such as carboxylic acids substantially decrease a molecule's logD value while \emph{lipophilic} \JL[groups make a molecule more lipophile \cite{alma99901018739001842}]{substructures such as aromatic rings increase the logD value}. 
We follow this classification of molecular substructures in our analysis and put all other molecular substructures that do not substantially affect lipophilicity into the \emph{other} group (\Cref{sec:app_ft_setup} provides a list of all molecular substructures). 

\subsection{Aqueous solubility}\label{sec:esol_prediction}

Aqueous solubility refers to the ability of a molecule to dissolve in water. 
Similar to lipophilicity, aqueous solubility is a physicochemical property of molecules which affects their biovailability as well as ADME profiles \citep{llompart2024will}. 
Therefore, predicting a molecule's solubility is of high importance to the development of orally active drugs/compounds \citep{aq_sol_importance}. 
The goal of aqueous solubility prediction is to estimate the logS (log solubility of a molecule in water). 





\subsection{Relation between Lipophilicity and Solubility}

Generally, a high lipophilicity is negatively correlated with aqueous solubility, decreasing oral absorption of a drug~\citep{CURATOLO1998387}. 
Moreover, both tasks share the same categories of important molecular subgroups---i.e., \textit{hydrophilic} substructures increase the hydrogen-bonding ability of a molecule, making it more likely to be soluble in water but decreasing lipophilicity while \textit{lipophilic} substructures increase its hydrophobicity. 
Although both tasks are closely related, computing the logD and logS values requries the consideration of additional, molecule-specific factors such as the melting point~\citep{HILL2010648}.

\section{Model Selection}
\label{sec:app_model_selection}

In our experiments, we consider the following \clms trained with the masked language modeling (MLM) objective on datasets of small molecules. 

\begin{description}[noitemsep,topsep=3pt,itemsep=3pt,itemindent=-1.5em]
\item[Chemberta] \citet{chemberta} release multiple versions of a six-layer model based on RoBERTa~\citep{liu2020roberta} trained on different amounts of data sampled from the ZINC15 database~\citep{ZINC15}.
We use the two publicly available models, namely, \texttt{seyonec/ChemBERTa-zinc-base-v1}~(here, referred to as \texttt{chemberta-base}; trained on 100k molecules) and \texttt{seyonec/chemberta-zinc250k-v1} (\texttt{chemberta}; trained on 250k molecules). Both employ a BPE tokenizer with respective vocabulary sizes of 767 and 52k tokens. While \texttt{Chemberta} has 83,450,880 parameters, \texttt{chemberta-base} has only 44,103,936 parameters.

\item[Chemberta-2] In subsequent work, \citet{chemberta2} release models pre-trained on a larger set of molecules sampled from ZINC15~\citep{ZINC15}. 
We investigate \texttt{DeepChem/chemberta-5M-MLM}, \texttt{DeepChem/chemberta-10M-MLM} and \texttt{DeepChem/chemberta-77M-MLM}, denoted as \texttt{chemberta-2-5M}, \texttt{chemberta-2-10M} and \texttt{chemberta-2-77M} resepectively, as the other models are pre-trained with domain-specific auxiliary objectives\footnote{These are the \texttt{chemberta-5M-MTR}, \texttt{chemberta-10M-MTR} and \texttt{chemberta-77M-MTR} where MTR refers to multitask regression of 200 molecular properties.}. \texttt{chemberta-2-5M/10M/77M} models all share the same number of parameters -- 3,427,440 parameters.
 Compared to the earlier version of Chemberta \cite{chemberta}, these models have only three encoder layers, a hidden layer size of 384, and a BPE tokenizer with a vocabulary size of 600 tokens.
The main difference among the three variants is the training data size: 5M, 10M, and 77M molecules, respectively. We note that \texttt{chemberta2} models exhibit tokenization problems \footnote{see \href{https://github.com/seyonechithrananda/bert-loves-chemistry/issues/60}{Issue 1} and \href{https://github.com/deepchem/deepchem/issues/4253}{Issue 2} and \href{https://discuss.huggingface.co/t/tokenizer-not-recognising-words-in-vocabulary/20140}{Issue 3}  }. In particular, some halogens such as chlorine (Cl) and bromine (Br) are tokenized incorrectly.

\item[Chemberta-3] Most recently, \citet{chemberta3} released their training framework along with a 12-layer model \texttt{DeepChem/ChemBERTa-100M-MLM}~(referred to as \texttt{chemberta-3}) trained on 100M molecules sampled from ZINC20~\citep{zinc20}.
In contrast to the \texttt{chemberta-2} models, \texttt{chemberta-3} is again based on RoBERTa using a self-trained tokenizer with a vocabulary size of 7,924 tokens. Finally, compared to the \texttt{chemberta-2} models, both the size of the hidden vectors and the size of the intermediate FFNNs are larger (3072 and 768 dimensions, respectively). \texttt{chemberta-3} has 92,126,976 parameters.

\item[Molformer] \citet{ross2022large} train a 12 layer RoBERTa-based model using linear attention, rotary positional embeddings and a regex-based tokenizer with a vocabulary size of 2362 tokens. 
We use the publicly available \texttt{ibm-research/MoLFormer-XL-both-10pct} model\footnote{We note that the best model (\texttt{Molformer-XL}) whose results are reported by \citep{ross2022large} is unavailable} trained on 100M molecules sampled equally from ZINC15~\citep{ZINC15} and PubChem~\citep{pubchem} with 44,375,040 parameters.
\item[roberta-zinc-480m]~\citep{heyer2023roberta} is a RoBERTa-based 14 layer, $\sim$102M parameter model trained on 480M molecules sampled from the ZINC20 database \citep{zinc20}. The model is available on Huggingface under \texttt{entropy/roberta\_zinc\_480m}.
\end{description}

For the experiments, we select models that are comparable and only differ with respect to their architecture and training data sizes. We thus omit models trained with domain-specific auxiliary objectives such as MolBERT~\citep{MolBERT}, SELFormer~\citep{Yüksel_2023}, Mol-BERT~\citep{li2021mol}, K-BERT~\citep{kbert} and those that are not publicly available (e.g., SMILES-BERT, \citealt{Wang-2019-SmilesBert}).

\section{Probing Dataset}
\label{sec:probing_set_info}

We derive our probing dataset from PCQM4Mv2, a publicly available dataset designed for predicting the HOMO-LUMO energy gap. 
It is part of the open graph benchmark~\citep{hu2021ogblsc} and is published under an open source license (CC-BY 4.0). 

\subsection{Dataset Preprocessing}
\label{sec:probing_set_creation}

In \Cref{sec:probingdataset}, we described the preprocessing steps used to obtain the final set of probing datasets. 
Specifically, all molecules were preprocessed with RDKIT~\citep{greg_landrum_2020_3732262}\footnote{version 2024.3.6}, and molecules were discarded if they exhibited any of the following issues: 1) invalid SMILES strings: (MolFromSmiles cannot be created), 2) incorrect conformational information (GetConformer().Is3D() returned False), 3) chemistry problems (rdkit.Chem.DetectChemistryProblems is not empty), 4) other processing errors (e.g., due to unrecoverable rotational or double bond information).
Finally, the remaining molecules were annotated with binary labels with 1 denoting the presence of a substructure. To extract functional group information, we used the rdkit.Chem.Fragments module from RDKit. Ring structure labels and bond information were annotated using rdkit.Chem.Lipinski. We then binarized the labels.

\paragraph{Runtime estimates}
The sheer number of instances in the PCQM4Mv2 dataset requires the subsampling of molecules for the probing dataset. 
For reference, running all 78 probing tasks for a single 14-layer model with 100k instances requires $\sim$5.4 hours. 
Scaling this to the whole dataset would require almost 200 hours for only one (out of four) experimental configuration.

\subsection{Probing Tasks}
\label{sec:probing_set_tasks}
Our final probing dataset comprises 78 probing tasks, as summarized in  \Cref{tab:substructures_descriptions}. The \emph{Molecular Substructure} column lists the abbreviations used for each probing task, while \emph{Description} provides a brief explanation of the corresponding task.

\begin{table*}[t]
\centering
\resizebox{0.4\textwidth}{!}{%
    \begin{tabular}{ll}
\toprule
Molecular Substructure & Description \\
\midrule
NHOH & NHs or OHs \\
NO & nitrogens and oxygens \\
AliphaticCarbocycles & aliphatic (containing at least one non-aromatic bond) carbocycles \\
AliphaticHeterocycles & aliphatic (containing at least one non-aromatic bond) heterocycles \\
AliphaticRings & aliphatic (containing at least one non-aromatic bond) rings \\
AromaticCarbocycles & aromatic carbocycles \\
AromaticHeterocycles & aromatic heterocycles \\
AromaticRings & aromatic rings \\
HAcceptors & hydrogen bond acceptors \\
HDonors & hydrogen bond donors \\
Heteroatoms & heteroatoms \\
RotatableBonds & rotatable bonds \\
SaturatedCarbocycles & saturated carbocycles \\
SaturatedHeterocycles & saturated heterocycles  \\
SaturatedRings & saturated rings  \\
Ring & rings \\
Al\_COO & aliphatic carboxylic acids \\
Al\_OH & aliphatic hydroxyl groups \\
Al\_OH\_noTert & aliphatic hydroxyl groups excluding tert-OH \\
ArN & N functional groups attached to aromatics \\
Ar\_COO & aromatic carboxylic acid \\
Ar\_N & aromatic nitrogens \\
Ar\_NH & aromatic amines \\
Ar\_OH & aromatic hydroxyl groups \\
COO & carboxylic acids \\
COO2 & carboxylic acids \\
C\_O & carbonyl O \\
C\_O\_noCOO & carbonyl O excluding COOH \\
C\_S & thiocarbonyl \\
Imine & imines \\
NH0 & tertiary amines \\
NH1 & secondary amines \\
NH2 & primary amines \\
N\_O & hydroxylamine groups \\
Ndealkylation1 & XCCNR groups \\
Ndealkylation2 & tert-alicyclic amines (no heteroatoms, not quinine-like bridged N) \\
Nhpyrrole & H-pyrrole nitrogens \\
SH & thiol groups \\
aldehyde & aldehydes \\
alkyl\_halide & alkyl halides \\
\bottomrule
\end{tabular}
}
\resizebox{0.4425\textwidth}{!}{%
    \begin{tabular}{ll}
\toprule
Molecular Substructure & Description \\
\midrule
allylic\_oxid &  allylic oxidation sites excluding steroid dienone \\
amide & amides \\
amidine & amidine groups \\
aniline & anilines \\
aryl\_methyl & aryl methyl sites for hydroxylation \\
benzene & benzene rings \\
bicyclic & bicyclic \\
diazo & diazo groups \\
ester & esters \\
ether & ether oxygens (including phenoxy) \\
furan & furan  \\
guanido & guanidine groups \\
halogen & halogens \\
hdrzine & hydrazine groups \\
hdrzone & hydrazone groups \\
imidazole & imidazole  \\
imide & imide groups \\
ketone & ketones \\
ketone\_Topliss & ketones excluding diaryl, a,b-unsat. dienones, heteroatom on Calpha \\
lactone & cyclic esters (lactones) \\
methoxy & methoxy groups -OCH3 \\
morpholine & morpholine \\
nitrile & nitriles \\
nitro & nitro groups \\
nitro\_arom & nitro benzene ring substituents \\
nitro\_arom\_nonortho & non-ortho nitro benzene ring substituents \\
oxime & oxime groups \\
para\_hydroxylation & para-hydroxylation sites \\
phenol & phenols \\
phenol\_noOrthoHbond & phenolic OH excluding ortho intramolecular Hbond substituents \\
piperdine & piperdine  \\
piperzine & piperzine  \\
priamide & primary amides \\
pyridine & pyridine  \\
sulfide & thioether \\
thiazole & thiazole  \\
thiophene & thiophene  \\
unbrch\_alkane & unbranched alkanes of at least 4 members (excludes halogenated alkanes) \\
urea & urea groups \\
\bottomrule
\end{tabular}
}
\caption{\textbf{Probing tasks:} Molecular substructure abbreviations and descriptions}
\label{tab:substructures_descriptions}
\end{table*}

\begin{table*}[ht]
\centering
\begin{minipage}[t]{0.45\textwidth}
\resizebox{\textwidth}{!}{%
    \begin{tabular}{lrrr}
\toprule
\textbf{Probing task} & \multicolumn{3}{c}{\textbf{Training Split Size}} \\
\cmidrule(lr){2-4}
 & Seed 42 &  Seed 77 & Seed 4 \\
\midrule
NHOH & 54574 & 54936 & 54628 \\
NO & 5308 & 5374 & 5454 \\
AliphaticCarbocycles & 44174 & 43876 & 44168 \\
AliphaticHeterocycles & 66372 & 66602 & 66996 \\
AliphaticRings & 99440 & 99306 & 99924 \\
AromaticRings & 76818 & 77146 & 77070 \\
AromaticCarbocycles & 75432 & 75536 & 76012 \\
AromaticHeterocycles & 65176 & 64620 & 64144 \\
HAcceptors & 4662 & 4778 & 4826 \\
HDonors & 53418 & 53890 & 53548 \\
Heteroatoms & 2076 & 2060 & 2156 \\
RotatableBonds & 15078 & 15180 & 15118 \\
SaturatedRings & 70664 & 70306 & 71046 \\
SaturatedCarbocycles & 34032 & 33544 & 33866 \\
SaturatedHeterocycles & 42976 & 43060 & 43562 \\
Ring & 23322 & 23484 & 23270 \\
Al\_COO & 14092 & 14152 & 13996 \\
Al\_OH & 52366 & 52306 & 52222 \\
Al\_OH\_noTert & 48914 & 48874 & 48840 \\
ArN & 10872 & 10808 & 11102 \\
Ar\_COO & 3992 & 3956 & 3878 \\
Ar\_N & 56262 & 55878 & 55370 \\
Ar\_NH & 10330 & 10088 & 9820 \\
Ar\_OH & 8548 & 8220 & 8362 \\
COO & 18034 & 18050 & 17816 \\
COO2 & 19182 & 19154 & 18944 \\
C\_O & 68424 & 68080 & 68792 \\
C\_O\_noCOO & 52970 & 52798 & 53526 \\
C\_S & 870 & 812 & 792 \\
Imine & 28506 & 28680 & 28360 \\
ketone & 15014 & 15276 & 15558 \\
NH0 & 75328 & 75390 & 76000 \\
NH1 & 70862 & 70678 & 71058 \\
NH2 & 39752 & 39668 & 40030 \\
N\_O & 3462 & 3568 & 3484 \\
Ndealkylation1 & 14322 & 14354 & 14098 \\
Ndealkylation2 & 13014 & 13062 & 13032 \\
Nhpyrrole & 10330 & 10088 & 9820 \\
SH & 3550 & 3444 & 3436 \\
\bottomrule
\end{tabular}
}
\end{minipage}
\begin{minipage}[t]{0.455\textwidth}
\resizebox{\textwidth}{!}{%
    \begin{tabular}{lrrr}
\toprule
\textbf{Probing task} & \multicolumn{3}{c}{\textbf{Training Split Size}} \\
\cmidrule(lr){2-4}
 & Seed 42 &  Seed 77 & Seed 4 \\
\midrule
aldehyde & 4128 & 4054 & 4228 \\
alkyl\_halide & 12224 & 12168 & 12022 \\
allylic\_oxid & 19950 & 20470 & 20198 \\
amide & 19672 & 19528 & 19746 \\
amidine & 9086 & 9024 & 9054 \\
aniline & 27334 & 27340 & 27794 \\
aryl\_methyl & 42070 & 41994 & 41350 \\
benzene & 75432 & 75530 & 76012 \\
bicyclic & 35830 & 35808 & 35426 \\
ester & 15866 & 15856 & 15890 \\
ether & 63050 & 63536 & 63826 \\
furan & 5358 & 5272 & 5252 \\
guanido & 2322 & 2364 & 2346 \\
halogen & 35422 & 35528 & 35384 \\
hdrzine & 3384 & 3348 & 3280 \\
hdrzone & 1754 & 1792 & 1944 \\
imidazole & 7152 & 7336 & 7074 \\
imide & 1254 & 1166 & 1206 \\
ketone\_Topliss & 13072 & 13170 & 13448 \\
lactone & 1930 & 1940 & 2090 \\
methoxy & 22138 & 22602 & 22642 \\
morpholine & 2316 & 2170 & 2256 \\
nitrile & 11304 & 11414 & 11520 \\
nitro & 4444 & 4384 & 4396 \\
nitro\_arom & 2410 & 2418 & 2368 \\
nitro\_arom\_nonortho & 1182 & 1200 & 1116 \\
oxime & 1570 & 1476 & 1556 \\
para\_hydroxylation & 12858 & 12690 & 12736 \\
phenol & 5432 & 5382 & 5422 \\
phenol\_noOrthoHbond & 5354 & 5284 & 5306 \\
piperdine & 10696 & 10720 & 10628 \\
piperzine & 3064 & 3168 & 3176 \\
pyridine & 18140 & 18244 & 18118 \\
priamide & 576 & 556 & 572 \\
sulfide & 9854 & 9744 & 9946 \\
thiazole & 3874 & 3852 & 3782 \\
thiophene & 5610 & 5394 & 5370 \\
unbrch\_alkane & 12118 & 12130 & 11878 \\
urea & 758 & 752 & 768 \\
\bottomrule
\end{tabular}

}
\end{minipage}
\caption{\textbf{Probing training dataset statistics.} Each row shows the size of the undersampled training split for each probing task across three random samples. Some substructures occur infrequently in the original PCQM4Mv2 dataset, resulting in smaller training sets for certain tasks (e.g., \emph{priamide}, \emph{urea}).}
\label{tab:probing_train_sizes}
\end{table*}

\begin{table*}[ht]
\centering
\begin{minipage}[t]{0.45\textwidth}
\resizebox{\textwidth}{!}{%
    \begin{tabular}{lrrr}
\toprule
\textbf{Probing task} & \multicolumn{3}{c}{\textbf{\% of molecules w/substructure}} \\
\cmidrule(lr){2-4}
 & Seed 42 &  Seed 77 & Seed 4 \\
\midrule
NHOH & 61.20 & 61.51 & 60.86 \\
NO & 95.38 & 95.48 & 95.42 \\
AliphaticCarbocycles & 17.38 & 17.34 & 17.29 \\
AliphaticHeterocycles & 29.50 & 29.54 & 30.05 \\
AliphaticRings & 42.71 & 42.77 & 43.22 \\
AromaticRings & 64.66 & 64.69 & 64.67 \\
AromaticCarbocycles & 49.26 & 48.84 & 49.48 \\
AromaticHeterocycles & 27.84 & 28.32 & 27.61 \\
HAcceptors & 95.94 & 96.14 & 96.03 \\
HDonors & 62.17 & 62.59 & 61.91 \\
Heteroatoms & 97.82 & 97.92 & 97.84 \\
RotatableBonds & 88.03 & 88.13 & 87.86 \\
SaturatedRings & 26.55 & 26.68 & 26.68 \\
SaturatedCarbocycles & 11.27 & 11.21 & 11.04 \\
SaturatedHeterocycles & 17.39 & 17.46 & 17.59 \\
Ring & 85.53 & 85.84 & 86.00 \\
Al\_COO & 7.17 & 7.14 & 7.42 \\
Al\_OH & 21.76 & 21.98 & 21.84 \\
Al\_OH\_noTert & 20.09 & 20.36 & 20.27 \\
ArN & 4.68 & 4.50 & 4.27 \\
Ar\_COO & 2.06 & 2.09 & 2.11 \\
Ar\_N & 23.54 & 23.74 & 22.96 \\
Ar\_NH & 5.59 & 5.83 & 5.48 \\
Ar\_OH & 5.36 & 5.11 & 4.88 \\
COO & 9.20 & 9.19 & 9.48 \\
COO2 & 9.24 & 9.24 & 9.54 \\
C\_O & 42.93 & 43.13 & 43.42 \\
C\_O\_noCOO & 35.73 & 36.13 & 36.20 \\
C\_S & 1.06 & 1.01 & 1.01 \\
Imine & 12.85 & 12.77 & 13.32 \\
ketone & 10.04 & 9.84 & 9.61 \\
NH0 & 57.19 & 57.01 & 57.49 \\
NH1 & 28.09 & 28.98 & 28.75 \\
NH2 & 12.85 & 12.63 & 12.24 \\
N\_O & 1.97 & 1.95 & 2.06 \\
Ndealkylation1 & 3.51 & 3.48 & 3.54 \\
Ndealkylation2 & 3.62 & 3.61 & 3.87 \\
Nhpyrrole & 5.59 & 5.83 & 5.48 \\
SH & 3.23 & 3.26 & 3.12 \\
\bottomrule
\end{tabular}

}
\end{minipage}
\begin{minipage}[t]{0.455\textwidth}
\resizebox{\textwidth}{!}{%
    \begin{tabular}{lrrr}
\toprule
\textbf{Probing task} & \multicolumn{3}{c}{\textbf{\% of molecules w/substructure}} \\
\cmidrule(lr){2-4}
 & Seed 42 &  Seed 77 & Seed 4 \\
\midrule
aldehyde & 2.04 & 2.02 & 2.00 \\
alkyl\_halide & 6.27 & 6.12 & 6.12 \\
allylic\_oxid & 12.45 & 12.53 & 12.56 \\
amide & 14.00 & 14.27 & 14.69 \\
amidine & 5.04 & 5.41 & 5.44 \\
aniline & 13.87 & 13.58 & 13.42 \\
aryl\_methyl & 17.68 & 17.71 & 18.09 \\
benzene & 49.20 & 48.78 & 49.43 \\
bicyclic & 24.58 & 24.25 & 24.30 \\
ester & 10.78 & 11.16 & 11.09 \\
ether & 30.86 & 31.51 & 31.12 \\
furan & 2.37 & 2.66 & 2.63 \\
guanido & 1.56 & 1.64 & 1.59 \\
halogen & 18.18 & 17.68 & 17.57 \\
hdrzine & 2.75 & 2.65 & 2.61 \\
hdrzone & 3.01 & 3.13 & 3.33 \\
imidazole & 3.17 & 3.29 & 3.16 \\
imide & 1.52 & 1.57 & 1.62 \\
ketone\_Topliss & 7.95 & 7.81 & 7.54 \\
lactone & 1.38 & 1.44 & 1.39 \\
methoxy & 10.91 & 11.30 & 11.19 \\
morpholine & 1.12 & 1.18 & 1.14 \\
nitrile & 5.19 & 5.31 & 5.36 \\
nitro & 4.65 & 4.46 & 4.46 \\
nitro\_arom & 2.81 & 2.69 & 2.75 \\
nitro\_arom\_nonortho & 1.77 & 1.71 & 1.71 \\
oxime & 1.71 & 1.76 & 1.74 \\
para\_hydroxylation & 10.64 & 10.49 & 10.81 \\
phenol & 3.87 & 3.75 & 3.58 \\
phenol\_noOrthoHbond & 3.75 & 3.65 & 3.50 \\
piperdine & 3.48 & 3.38 & 3.58 \\
piperzine & 1.17 & 1.23 & 1.21 \\
pyridine & 7.32 & 7.14 & 6.98 \\
priamide & 1.42 & 1.49 & 1.47 \\
sulfide & 6.33 & 6.22 & 6.25 \\
thiazole & 1.45 & 1.55 & 1.46 \\
thiophene & 2.36 & 2.44 & 2.44 \\
unbrch\_alkane & 7.11 & 7.14 & 7.00 \\
urea & 1.76 & 1.70 & 1.71 \\
\bottomrule
\end{tabular}
}
\end{minipage}
\caption{\textbf{Class distributions in the probing test set.} Each row shows the percentage of molecules containing a molecular substructure probed for in each of the three randomly drawn test samples.}
\label{tab:probing_test_distributions}
\end{table*}

\subsection{Dataset Statistics}
\label{sec:probing_set_stats}
\Cref{tab:probing_train_sizes} summarizes the sizes of the downsampled training sets for each probing task. \Cref{tab:probing_test_distributions} reports the class distribution in the corresponding test sets. Note that many molecular substructures are highly infrequent, resulting in significant class imbalance. To address this, we evaluate probing performance using the macro-averaged F1 score, which weighs each class equally.
For training the probing classifier, we also downsample instances in the training set (i.e., the probing classifier is always trained on a balanced dataset).

\section{Probing Setup and Compute Infrastructure}\label{sec:appendix_experimental_setup}

\subsection{Probing Setup}
\label{sec:app_probing_setup}
For each probing task and each encoder layer of a \clm, we train a logistic regression\ classifier using scikit learn\footnote{sklearn.linear\_model.LogisticRegression} with default parameters and set \emph{max\_iter}=2000. The input features are the hidden representations from the corresponding layer. 
We apply padding to the longest sequence in each batch.
Probing experiments were conducted on CPU-only nodes on the same cluster as described in \Cref{sec:appendix-compute-env}. 
As an indicative reference, a single probing run for one probing task on a 14-layer model (e.g., \texttt{roberta-zinc-480m}) typically completes in approximately 4.14 minutes on an Intel Xeon E5-2620 CPU. 
Runtime varies slightly with the number of layers and probing tasks.

\subsection{Computing Environment}\label{sec:appendix-compute-env}
All experiments were performed on a high performance computing cluster with varying CPU and GPU architectures. 
Fine-tuning experiments were conducted on a single GPU; either NVIDIA Titan X (12GB), NVidia V100 (32GB), or A100 (40GB). 
Probing experiments were executed on 2 CPUs equipped with AMD EPYC 7662 (64 cores, 512 GB RAM) and did not require a GPU. 

\subsection{Compute Time}
Overall, $\sim677.9$ CPU-only hours were spent for probing.
For fine-tuning, the experiments required $\sim57.9$ GPU hours for lipophilicity prediction and $77.3$ GPU hours for solubility prediction. 
The downstream task experiments for the baselines required $692.5$ GPU hours for \texttt{gpt-oss-20b} and $7.2$ GPU hours for  \texttt{Llama-3.2-3B-Instruct}.
All baseline experiments using the feature-based models using fingerprints required less than 1 CPU-only hour in total. 
Finally, the further pre-training experiments required $\sim32.5$ GPU hours.
In total, this amounts to approximately 678.9 CPU-only hours and 867.4 GPU hours. 

\section{Effect of Pre-training (RQ1)}
\label{sec:pt_effect_apx}

We provide further evidence and analysis for our findings regarding the effect of pre-training on ring structures (\S\ref{sec:pre-training-rings}) and individual molecular substructures (\S\ref{sec:pre-training-molecules}).
Moreover, we perform analysis excluding the strangely behaving \texttt{chemberta-3} model which we study more extensively in \Cref{sec:appendix-chemberta-3}.
Excluding \texttt{chemberta-3} this model, we observe multiple additional patterns which we did not discuss in the main paper (as the analysis in the main paper includes \texttt{chemberta-3}). 


\subsection{Ring Structures}
\begin{figure*}[htb]
        \includegraphics[width=1.0\linewidth]
        {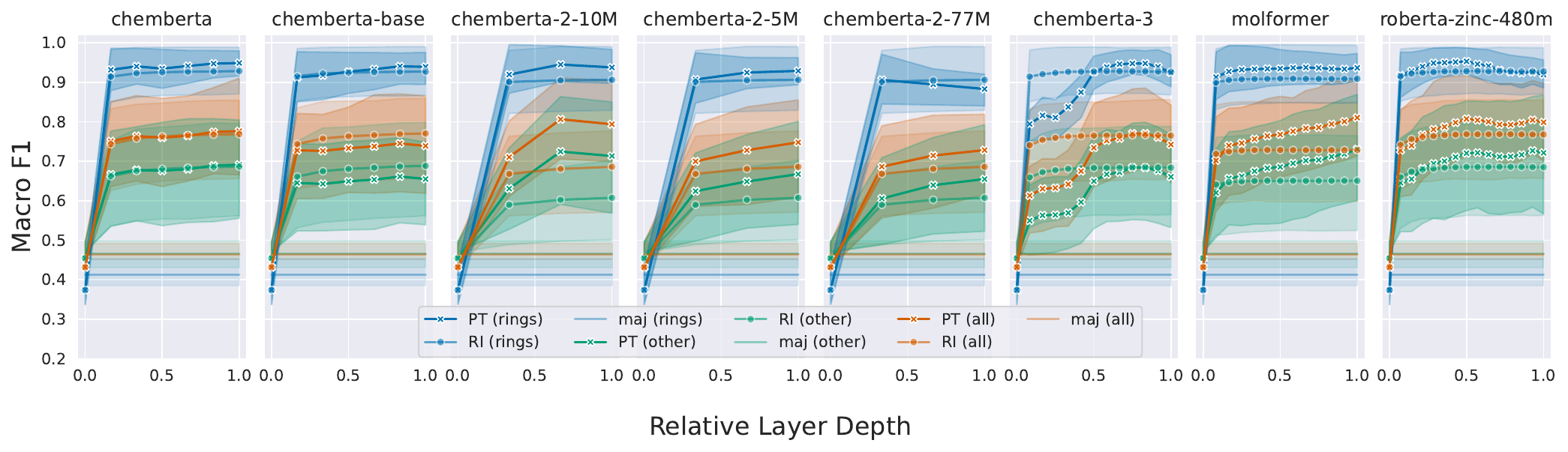} 
        \vspace{-2em}
    \caption{Average probing performance (macro-averaged F1 score $\uparrow$) of pre-trained (PT), randomly initialized (RI) models, and the majority class prediction (maj) on molecular substructures with \textbf{shading between the upper and lower quartiles}.
    We report average performance on all 12 classes of rings (avg rings), all other 66 substructures (avg other) and all 78 substructures (avg all).}
    \label{fig:probing_plots_iqr}
\end{figure*}

For ease of visualization, we omitted the shading that shows the upper and lower quartiles in \Cref{fig:bp}. 
\Cref{fig:probing_plots_iqr} presents the average probing performance along with its variability for \textcolor{blue}{rings}, \textcolor{ForestGreen}{other} and \textcolor{orange}{all}. 
Notably, \textcolor{blue}{rings} have the lowest variability across both PT and RI models, while probing performance of \textcolor{ForestGreen}{other} shows greated variability.
We provide the 12 ring types used to compute \textcolor{blue}{rings} in \Cref{tab:ring_structures_avg}.

We further analyze the representations of rings and other molecular substructures, finding that the representations of both RI (\oblue) and PT (\xblue) models perform exceptionally well at identifying ring structures compared to all other groups (\textbf{\textcolor{ForestGreen}{$\blackcircle{3pt}$}} and \xgreen). 
Moreover, both RI and PT models \textbf{encode ring structures well already at the first encoder layer}, suggesting that these surface-level patterns are easy for the models to extract directly from the input, even without pre-training. 
We also find that the benefit of pre-training diminishes for ring structures compared to that of other molecular substructures (i.e., the gap between \xblue~and \oblue~is much smaller compared to the gap between \xgreen~and \ogreen).

\paragraph{Different rings}
For different ring types, we find that---in contrast to aliphatic and aromatic rings---pre-training does improve the probing performance for saturated rings for all models except for \texttt{chemberta-3}.

\begin{figure}[htb]
    \centering
    \includegraphics[width=0.7\linewidth]{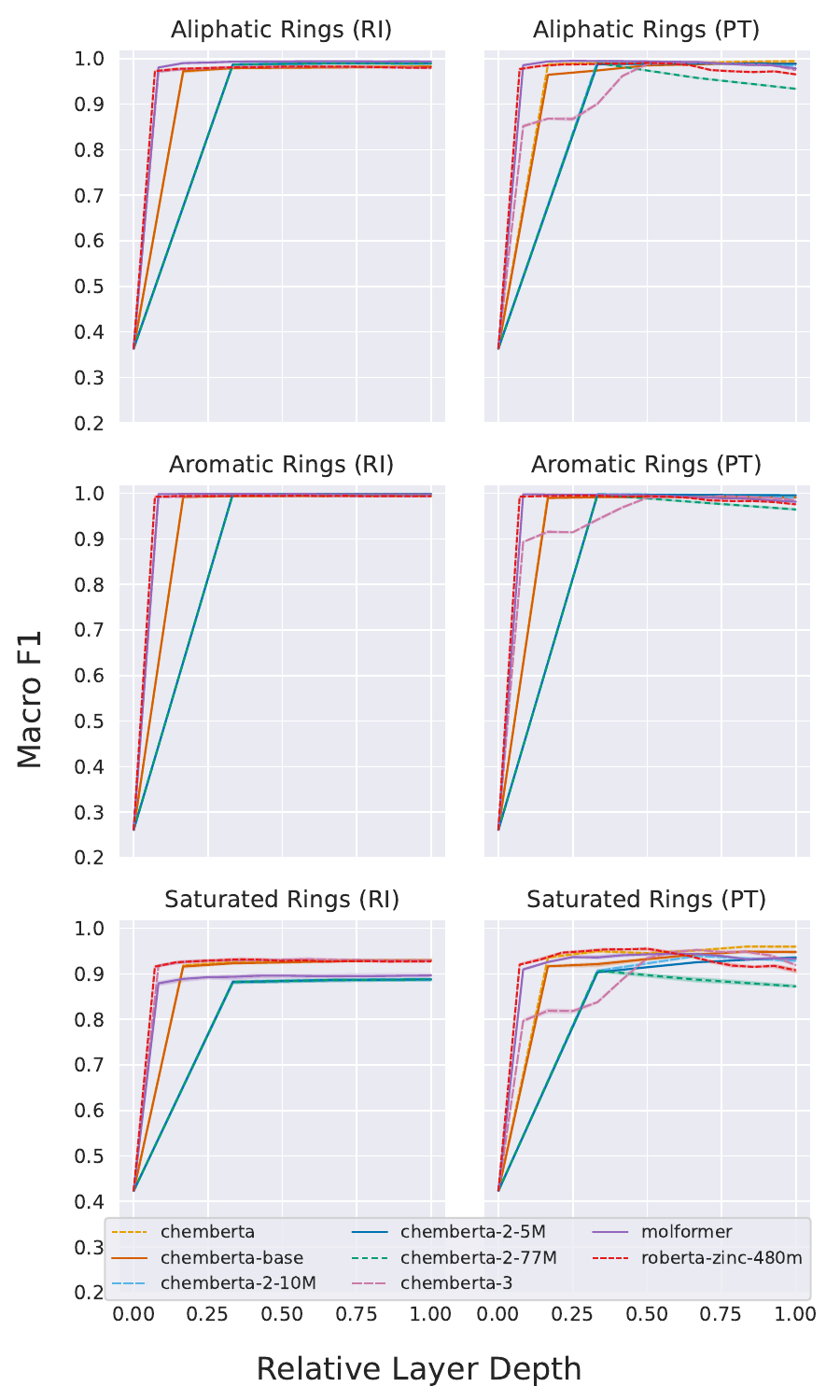} 
    \caption{Probing performance on aliphatic (top), aromatic (mid), and saturated (bottom) rings for randomly initialized (left) and pre-trained (right) models. As can be seen, the differences between randomly initialized and pre-trained models are minimal for aliphatic and aromatic rings (except for \texttt{chemberta-3}, for which we provide an explanation in \Cref{sec:appendix-chemberta-3}). This shows that random initializations already encode these ring structures very well.}
    \label{fig:performance-rings}
\end{figure}

\begin{table*}[htb]
\centering
\begin{tabular}{l}
\toprule
\textbf{Molecular Substructure} \\
\midrule
AliphaticCarbocycles \\
AliphaticHeterocycles \\
AliphaticRings\\
AromaticRings  \\
AromaticCarbocycles \\
AromaticHeterocycles\\
SaturatedCarbocycles\\
SaturatedHeterocycles \\
SaturatedRings \\
benzene\\
bicyclic\\
Ring \\
\bottomrule
\end{tabular}

\caption{List of the 12 ring types used for computing the average probing performance for \textcolor{blue}{rings} shown in \Cref{fig:bp}.}
\label{tab:ring_structures_avg}
\end{table*}

\begin{figure*}[ht]

  \begin{subfigure}[t]{0.5\textwidth}
        \centering
         \includegraphics[width=1.0\linewidth]{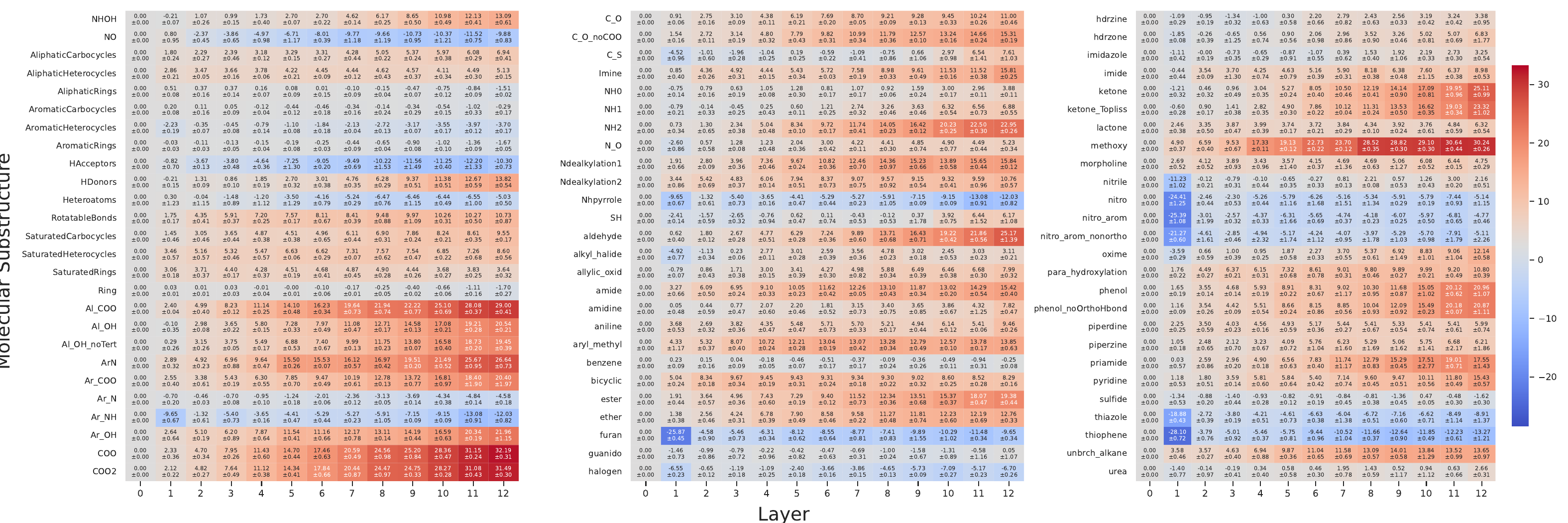}
        \caption{\scriptsize Molformer}
        \label{fig:delta_bp_molformer}
    \end{subfigure}
  \medskip
  \begin{subfigure}[t]{0.5\textwidth}
        \centering
         \includegraphics[width=1.0\linewidth]{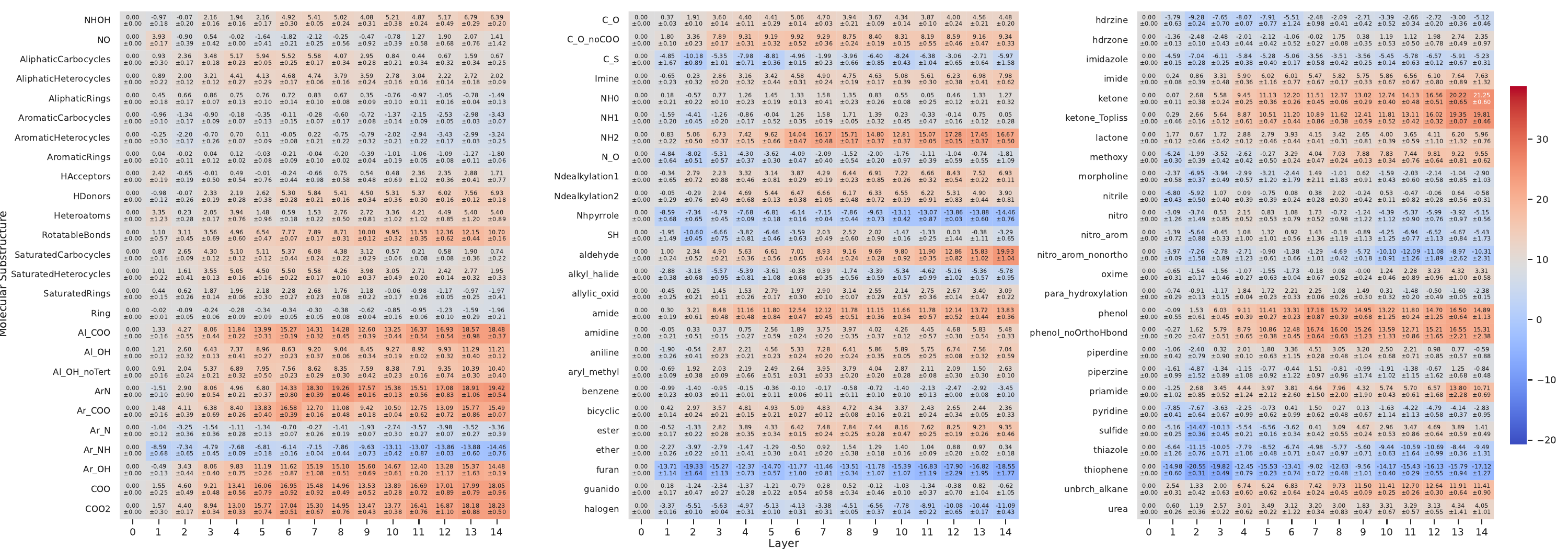}
        \caption{\scriptsize Roberta-zinc-480m}
        \label{fig:delta_bp_robertaz}
    \end{subfigure}
    \medskip
  \begin{subfigure}[t]{0.5\textwidth}
        \centering
         \includegraphics[width=1.0\linewidth]{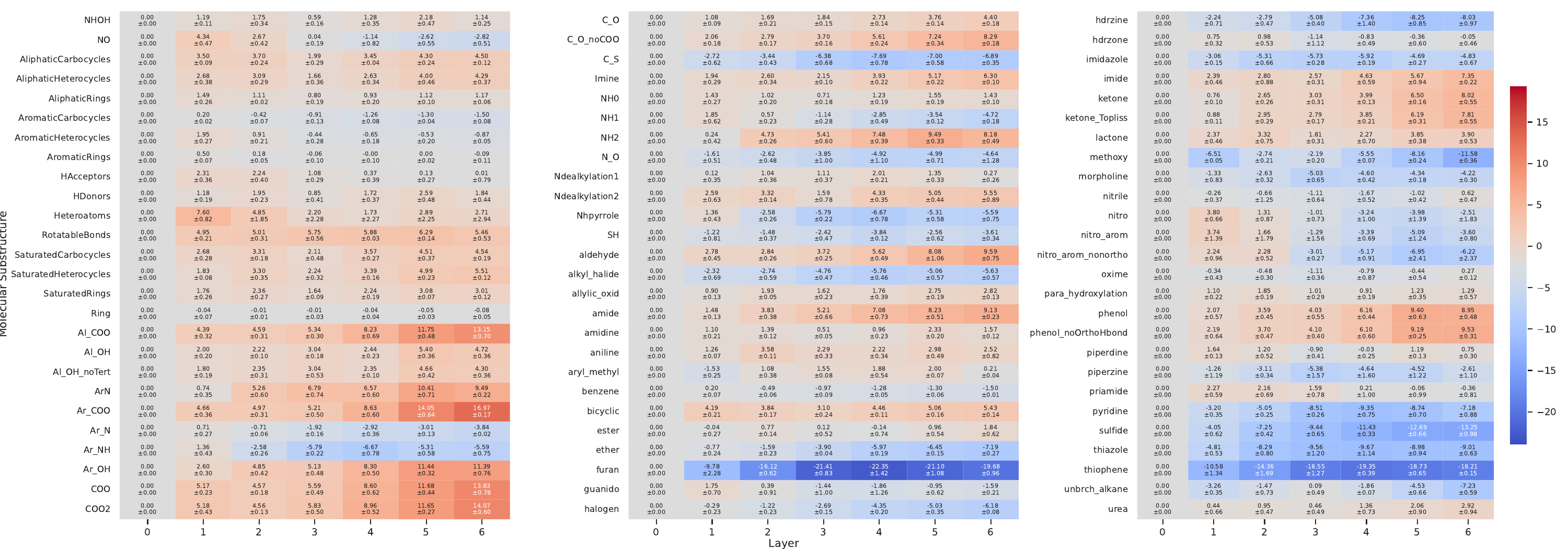}
        \caption{\scriptsize Chemberta}
        \label{fig:delta_bp_chemberta}
    \end{subfigure}
  \hfill
  \begin{subfigure}[t]{0.5\textwidth}
        \centering
         \includegraphics[width=1.0\linewidth]{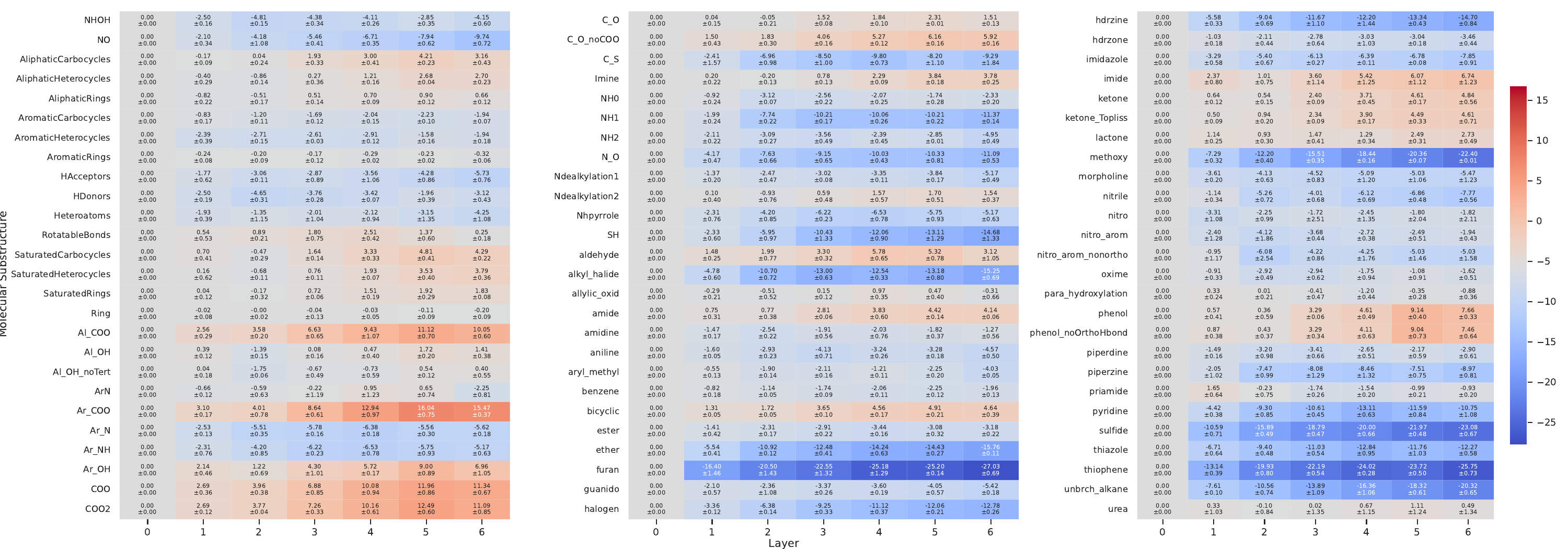}
        \caption{\scriptsize Chemberta-base}
        \label{fig:delta_bp_chemberta_base}
    \end{subfigure}
    \medskip
  \begin{subfigure}[t]{0.5\textwidth}
        \centering
         \includegraphics[width=1.0\linewidth]{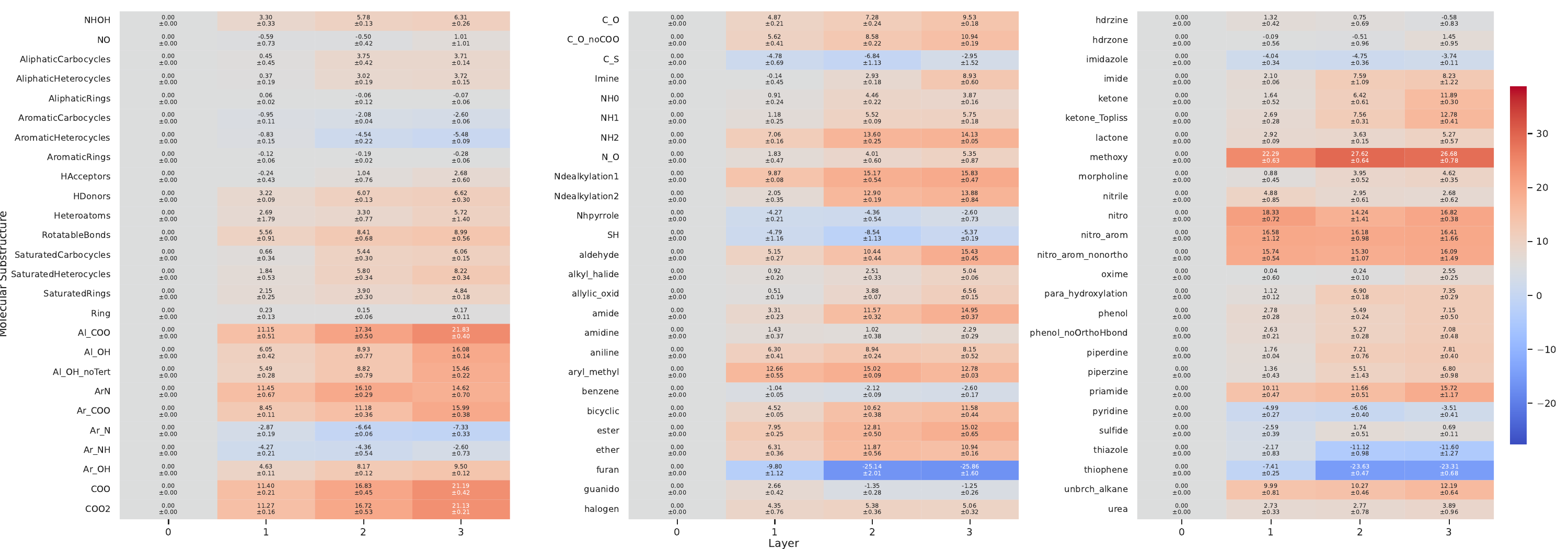}
        \caption{\scriptsize Chemberta-2-5M}
        \label{fig:delta_bp_chemberta2_5M}
    \end{subfigure}
  \hfill
  \begin{subfigure}[t]{0.5\textwidth}
        \centering
         \includegraphics[width=1.0\linewidth]{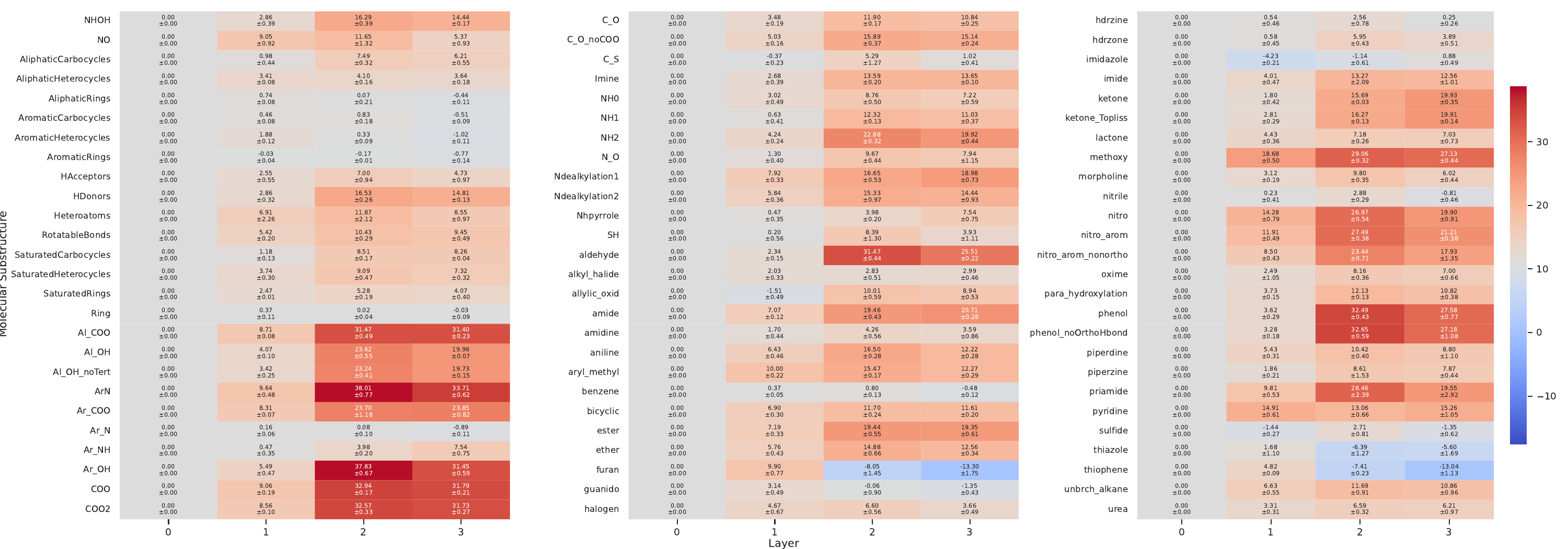}
        \caption{\scriptsize Chemberta-2-10M}
        \label{fig:delta_bp_chemberta2_10M}
    \end{subfigure}
     \medskip
  \begin{subfigure}[t]{0.5\textwidth}
        \centering
         \includegraphics[width=1.0\linewidth]{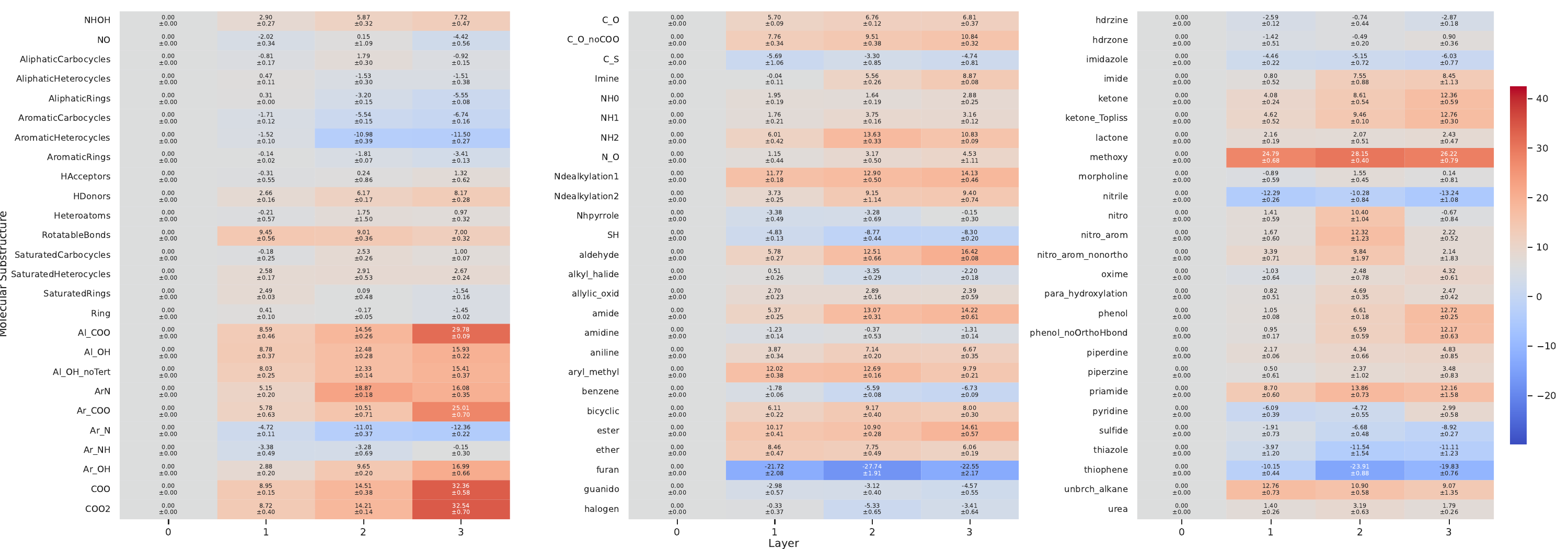}
        \caption{\scriptsize Chemberta-2-77M}
        \label{fig:delta_bp_chemberta2_77M}
    \end{subfigure}
     \begin{subfigure}[t]{0.5\textwidth}
        \centering
         \includegraphics[width=1.0\linewidth]{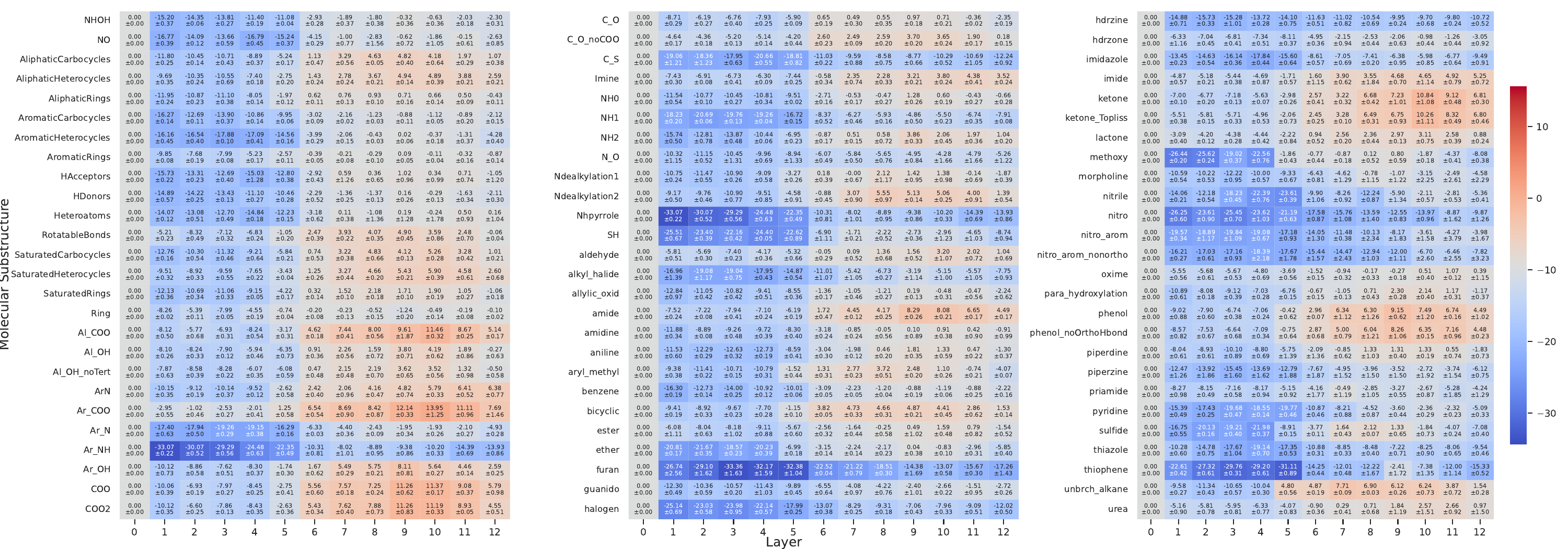}
        \caption{\scriptsize Chemberta-3}
        \label{fig:delta_bp_chemberta-3}
    \end{subfigure}
  \hfill
\caption{\textbf{Effect of pre-training on molecular substructure encoding in \clms}. We plot the layer-wise difference in performance (\% macro-averaged F1) on each probing task between each pre-trained model and its randomly initialized counterpart. \textcolor{red}{Red} denotes improvement in probing performance after pre-training. \textcolor{blue}{Blue} designates degradation in probing performance after pre-training}
\label{fig:app_delta_bp}
\end{figure*}

\subsection{Individual Molecular Substructures}

\Cref{fig:app_delta_bp} illustrates the layer-wise differences in probing performance after pre-training for individual molecular substructures. Results for \texttt{molformer} are shown in \Cref{fig:delta_bp_molformer}, while \texttt{roberta-zinc-480m} is presented in \Cref{fig:delta_bp_robertaz}. The performance of \texttt{chemberta} and \texttt{chemberta-base} is shown in \Cref{fig:delta_bp_chemberta} and \Cref{fig:delta_bp_chemberta_base}, respectively. The three \texttt{chemberta-2} models trained on 5M, 10M, and 77M molecules are shown in \Cref{fig:delta_bp_chemberta2_5M}, \Cref{fig:delta_bp_chemberta2_10M}, and \Cref{fig:delta_bp_chemberta2_77M}, respectively. Finally, the latest \texttt{chemberta-3} model's performance is shown in  \Cref{fig:delta_bp_chemberta-3}.
We additionally note the following for analysis:

\paragraph{\textcolor{red}{PT > RI}} 
In addition to the molecular substructures discussed in \S\ref{sec:pretraining}, we further identify consistent improvements across following substructures (when excluding \texttt{chemberta-3}). 
For \emph{hydroxy groups} (\texttt{Al\_OH}) and \emph{aldehyde} as well as \texttt{C\_O}, although to a lesser degree.

\section{Fine-Tuning Setup}
\label{sec:app_ft_setup}

In the following, we provide details about the two downstream tasks, namely, lipophilicity prediction and solubility prediction.
For each task, we provide a dataset description, report the hyperparameters considered for tuning, and elaborate the dataset pre-processing along with the final molecular substructures considered for probing.
For both tasks, we use the splits provided by \citet{ross2022large}.\footnote{Available at \url{https://github.com/IBM/molformer}}

\begin{table*}[t]
\centering
\resizebox{0.45\textwidth}{0.45\textheight}{%
\begin{tabular}{clccc}
\toprule
\multirow{14}{*}{\rotatebox{90}{\textbf{lipophilic}}} & \textbf{Molecular Substructure} & \multicolumn{3}{c}{\textbf{\% of molecules w/substructure}}\\
& & train & val & test \\

\midrule
&AliphaticCarbocycles & 14.79 & 16.67 & 19.29 \\
&AliphaticHeterocycles & 48.87 & 44.29 & 48.33 \\
&AliphaticRings & 56.58 & 55.24 & 61.43 \\
&AromaticCarbocycles & 89.70 & 90.00 & 90.48 \\
&AromaticHeterocycles & 73.42 & 76.43 & 74.05 \\
&AromaticRings & 99.02 & 98.33 & 98.33 \\
&SaturatedCarbocycles & 12.59 & 14.05 & 17.14 \\
&SaturatedHeterocycles & 40.27 & 35.48 & 40.00 \\
&SaturatedRings & 47.74 & 45.48 & 52.86 \\
&benzene & 89.70 & 90.00 & 90.48 \\
&halogen & 43.51 & 40.00 & 40.95 \\
&pyridine & 29.64 & 28.57 & 29.29 \\
&thiazole & 6.31 & 6.43 & 7.14 \\
&thiophene & 6.04 & 4.76 & 6.90 \\
\midrule
\multirow{23}{*}{\rotatebox{90}{\textbf{hydrophilic}}} &NHOH & 86.37 & 88.57 & 85.00 \\
&NO & 99.97 & 100.00 & 100.00 \\
&Heteroatoms & 99.97 & 100.00 & 100.00 \\
&HAcceptors & 99.94 & 100.00 & 100.00 \\
&HDonors & 86.37 & 88.81 & 85.00 \\
&Al\_COO & 7.71 & 5.48 & 6.90 \\
&Al\_OH & 13.30 & 16.43 & 10.00 \\
&Al\_OH\_noTert & 11.19 & 14.05 & 8.81 \\
&Ar\_COO & 3.36 & 5.71 & 4.29 \\
&Ar\_NH & 16.85 & 19.05 & 19.52 \\
&Ar\_OH & 8.66 & 11.19 & 7.86 \\
&COO & 11.07 & 11.19 & 10.95 \\
&COO2 & 11.07 & 11.19 & 10.95 \\
&NH0 & 87.68 & 83.81 & 84.76 \\
&NH1 & 65.48 & 68.57 & 66.90 \\
&NH2 & 19.55 & 20.71 & 20.71 \\
&amide & 51.25 & 49.29 & 52.38 \\
&aniline & 53.01 & 58.33 & 53.33 \\
&ether & 45.92 & 44.05 & 42.38 \\
&morpholine & 6.22 & 7.38 & 6.90 \\
&para\_hydroxylation & 16.93 & 18.81 & 21.19 \\
&phenol & 6.70 & 8.81 & 6.67 \\
&phenol\_noOrthoHbond & 6.58 & 8.81 & 6.19 \\
&piperdine & 18.21 & 15.48 & 17.62 \\
&piperzine & 9.64 & 8.81 & 10.00 \\
&priamide & 4.76 & 4.29 & 6.90 \\
\midrule
\multirow{21}{*}{\rotatebox{90}{\textbf{other}}} 
&RotatableBonds & 97.65 & 96.19 & 96.90 \\
&Ring & 99.94 & 99.76 & 99.52 \\
&ArN & 9.97 & 12.38 & 8.10 \\
&Ar\_N & 69.43 & 71.90 & 69.05 \\
&C\_O & 63.15 & 61.43 & 63.57 \\
&C\_O\_noCOO & 56.43 & 55.71 & 57.38 \\
&Imine & 1.93 & 1.67 & 2.62 \\
&Ndealkylation1 & 9.43 & 7.14 & 11.67 \\
&Ndealkylation2 & 16.37 & 11.19 & 14.29 \\
&Nhpyrrole & 16.85 & 19.05 & 19.52 \\
&alkyl\_halide & 8.39 & 6.19 & 7.14 \\
&aryl\_methyl & 25.77 & 27.14 & 25.00 \\
&bicyclic & 53.57 & 54.29 & 53.33 \\
&imidazole & 9.26 & 8.10 & 7.86 \\
&ketone & 4.85 & 6.90 & 7.38 \\
&ketone\_Topliss & 4.08 & 5.71 & 6.43 \\
&methoxy & 16.88 & 15.48 & 17.62 \\
&nitrile & 7.83 & 7.62 & 6.67 \\
&sulfide & 4.97 & 6.90 & 4.76 \\
&urea & 4.35 & 4.05 & 3.81 \\
\bottomrule
\end{tabular}

}
\caption{\textbf{Molecular substructure frequencies in the lipophilicity dataset}: Percentage of molecules in the training, validation and test splits of the lipophilicity dataset containing each of the 60 specific molecular substructures.}
\label{tab:lipo_stats}
\end{table*}

\begin{table*}[t]
\centering
\resizebox{0.6\textwidth}{0.4\textheight}{%
\begin{tabular}{clccc}
\toprule
\multirow{14}{*}{\rotatebox{90}{\textbf{lipophilic}}} & \textbf{Molecular Substructure} & \multicolumn{3}{c}{\textbf{\% of molecules w/substructure}}\\
& & train & val & test \\

\midrule
&AliphaticCarbocycles & 12.99 & 13.27 & 15.04 \\
&AliphaticHeterocycles & 14.65 & 9.73 & 16.81 \\
&AliphaticRings & 25.08 & 22.12 & 26.55 \\
&AromaticCarbocycles & 49.72 & 48.67 & 53.98 \\
&AromaticHeterocycles & 15.32 & 10.62 & 19.47 \\
&AromaticRings & 58.60 & 55.75 & 61.95 \\
&SaturatedCarbocycles & 9.66 & 8.85 & 12.39 \\
&SaturatedHeterocycles & 9.32 & 7.96 & 12.39 \\
&SaturatedRings & 17.20 & 16.81 & 21.24 \\

&benzene & 49.72 & 48.67 & 53.98 \\
&halogen & 29.41 & 30.97 & 27.43 \\
\midrule
\multirow{23}{*}{\rotatebox{90}{\textbf{hydrophilic}}}

&NHOH & 43.17 & 34.51 & 53.98 \\
&NO & 71.37 & 67.26 & 83.19 \\
&Heteroatoms & 86.13 & 84.07 & 89.38 \\
&HAcceptors & 72.25 & 68.14 & 83.19 \\
&HDonors & 43.62 & 34.51 & 53.98 \\
&Al\_OH & 13.98 & 15.93 & 11.50 \\
&Al\_OH\_noTert & 11.54 & 12.39 & 9.73 \\
&Ar\_OH & 7.88 & 4.42 & 13.27 \\
&NH0 & 26.75 & 27.43 & 33.63 \\
&NH1 & 18.87 & 14.16 & 30.97 \\
&amide & 18.87 & 15.93 & 28.32 \\
&aniline & 15.76 & 10.62 & 23.89 \\
&ester & 8.88 & 7.08 & 8.85 \\
&ether & 21.31 & 14.16 & 22.12 \\
&para\_hydroxylation & 10.43 & 7.96 & 15.04 \\
&phenol & 6.99 & 4.42 & 11.50 \\
&phenol\_noOrthoHbond & 6.88 & 4.42 & 11.50 \\
\midrule
\multirow{21}{*}{\rotatebox{90}{\textbf{other}}} 
&RotatableBonds & 68.04 & 67.26 & 68.14 \\
&Ring & 71.81 & 69.03 & 75.22 \\
&Ar\_N & 14.21 & 8.85 & 15.93 \\
&C\_O & 34.63 & 31.86 & 43.36 \\
&C\_O\_noCOO & 34.63 & 31.86 & 43.36 \\
&allylic\_oxid & 11.43 & 12.39 & 12.39 \\
&aryl\_methyl & 12.32 & 14.16 & 15.04 \\
&bicyclic & 22.09 & 18.58 & 28.32 \\
&imide & 4.99 & 7.08 & 11.50 \\
&unbrch\_alkane & 10.65 & 12.39 & 9.73 \\
&urea & 6.44 & 6.19 & 12.39 \\
\bottomrule
\end{tabular}

}
\caption{\textbf{Molecular substructure frequencies in the ESOL dataset}: Percentage of molecules in the training, validation and test splits of the ESOL dataset containing each of the 39 specific molecular substructure.}
\label{tab:esol_stats}
\end{table*}

\subsection{Lipophilicity} \label{sec:app_ft_setup_lipo}

\paragraph{Lipophilicity dataset}

\begin{figure*}[htb]

  \begin{subfigure}[t]{0.5\textwidth}
         \centering
    \includegraphics[width=1.0\linewidth]
    {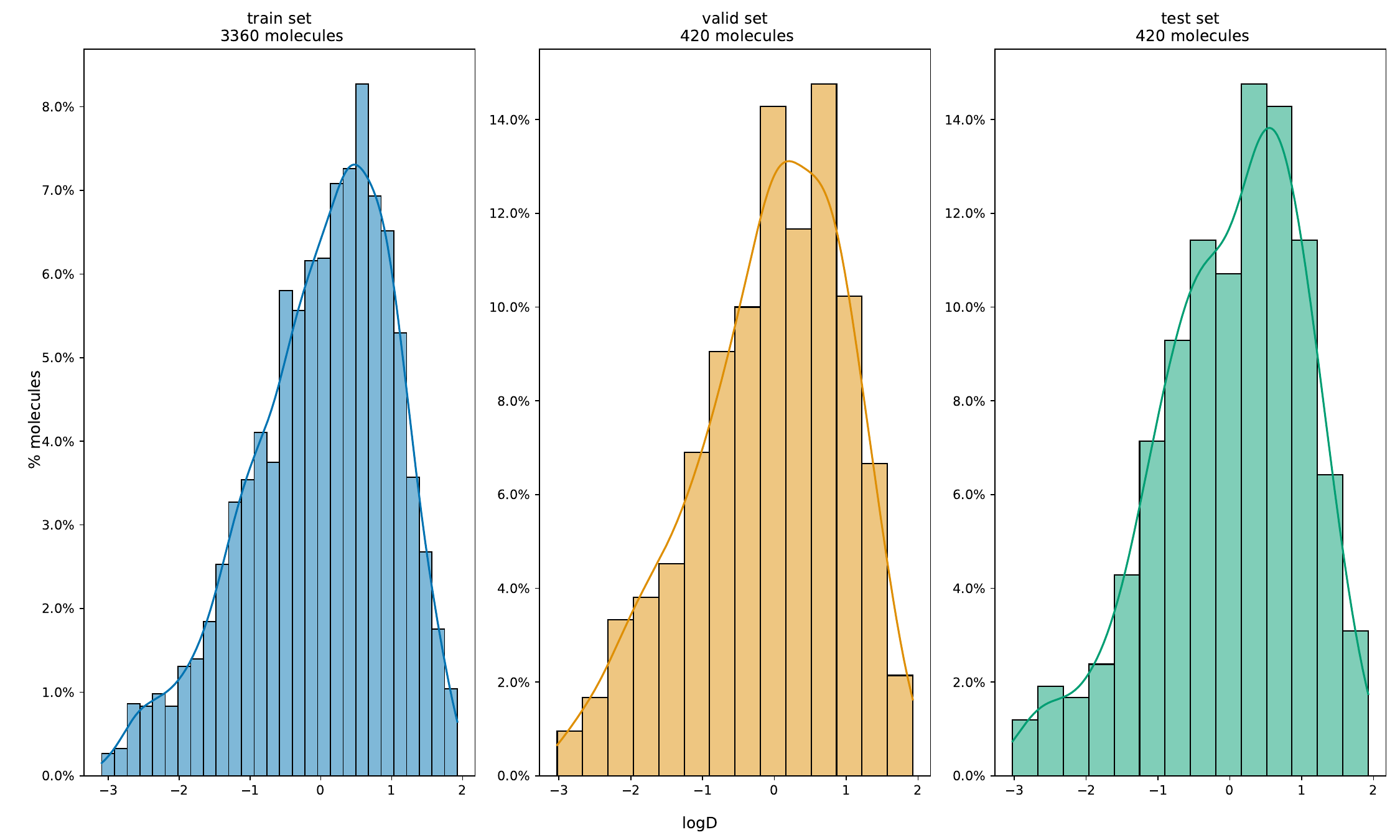}
    \caption{logD}
    \label{fig:app_plot_logd_distr}
    \end{subfigure}
  \medskip
  \begin{subfigure}[t]{0.5\textwidth}
    \centering
    \includegraphics[width=1.0\linewidth]   {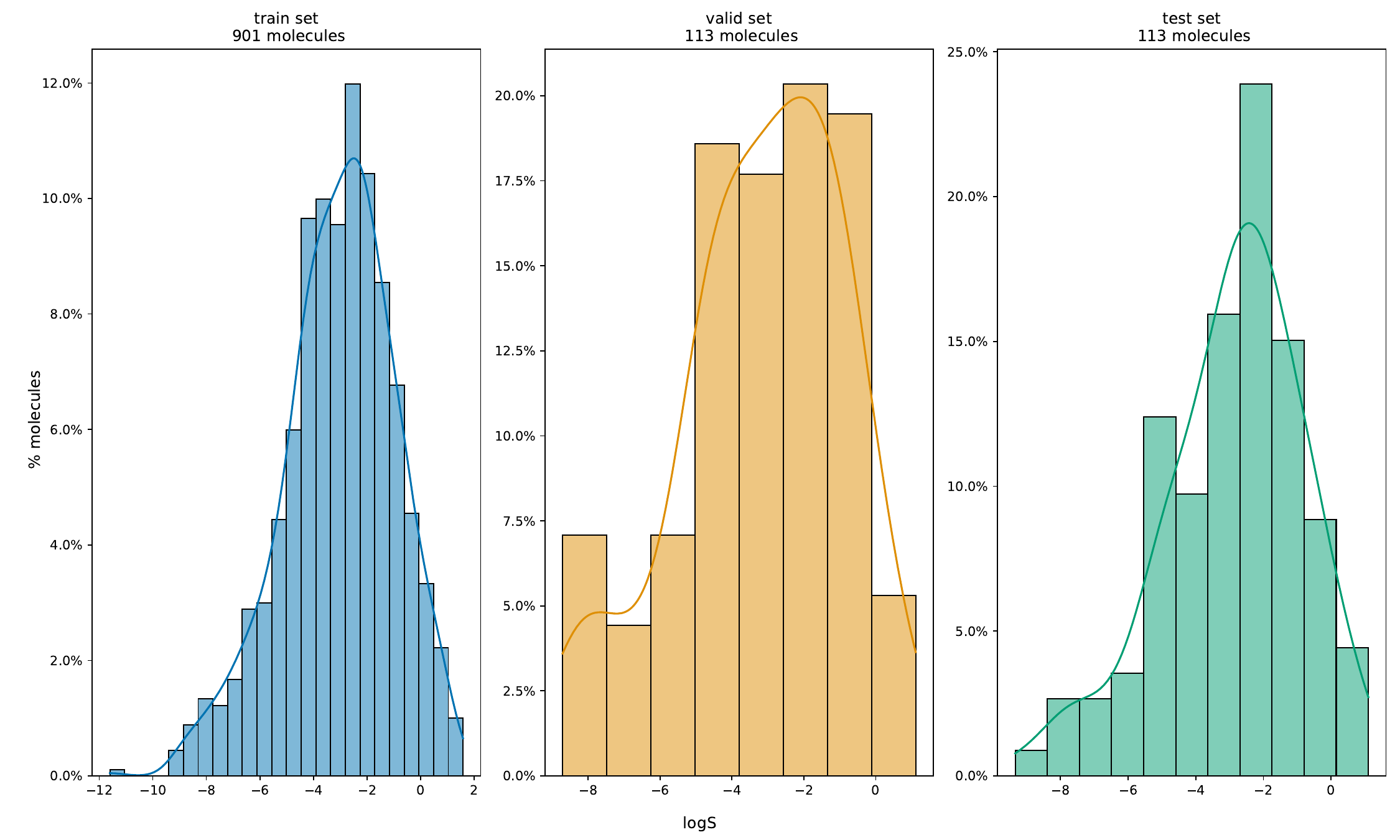}
    \caption{logS }
    \label{fig:app_plot_logs_distr}
\end{subfigure}

  \hfill
\caption{The distribution of predicted values across the training, validation and test splits of the \textbf{lipophilicity} (\ref{fig:app_plot_logd_distr}) and \textbf{solubility (ESOL)} (\ref{fig:app_plot_logs_distr}) datasets.} 
\label{fig:app_distr_logd_logs}
\end{figure*}

The lipophilicity dataset originates from MoleculeNet~\citep{wu2018moleculenetbenchmarkmolecularmachine}, an open source dataset (MIT license) and comprises 4,200 molecules and their corresponding logD values. 
\Cref{fig:app_plot_logd_distr} shows the distribution of logD values in the training, validation and test splits. 
As can be seen, the distributions of the logD value follow a similar shape across the training, validation, and test data with peaks around 0--1. 
\Cref{tab:logs_logd_stats} shows that the validation and test sets have very similar distributions, with means close to $-0.079$ and $-0.004$, while the training set is centered around 0. The validation set values are more dispersed ($\sigma$= 1.044) compared to the similar variability of the training ($\sigma$= 1.0 and 1.01, respectively). The more negative logD values mean that the validation and test sets are centered slightly towards molecules with lower lipophilicity.

\begin{table*}[ht]
\centering
\begin{minipage}[t]{0.45\textwidth}
\resizebox{\textwidth}{!}{%
    \begin{tabular}{lrrr}
\toprule
&\multicolumn{3}{c}{\textbf{logD}} \\
\cline{2-4}
 & \textbf{train} & \textbf{valid} & \textbf{test} \\
\midrule
\textbf{mean }& 0.000000 & -0.078882 & -0.004254 \\
\textbf{std }& 1.000149 & 1.043810 & 1.014424 \\
\textbf{min} & -3.091151 & -3.024248 & -3.024248 \\
\textbf{25\% } & -0.632465 & -0.718184 & -0.665917 \\
\textbf{50\%} & 0.145283 & 0.061654 & 0.157827 \\
\textbf{75\%} & 0.755773 & 0.707686 & 0.755773 \\
\textbf{max} & 1.926575 & 1.926575 & 1.926575 \\
\bottomrule
\end{tabular}

}
\end{minipage}
\begin{minipage}[t]{0.455\textwidth}
\resizebox{\textwidth}{!}{%
    \begin{tabular}{lrrr}
\toprule
&\multicolumn{3}{c}{\textbf{logS}} \\
\cline{2-4}
 & \textbf{train} & \textbf{valid} & \textbf{test} \\
\midrule

\textbf{mean} & -3.047764 & -3.186230 & -2.919593 \\
\textbf{std }& 2.076592 & 2.294987 & 2.061854 \\
\textbf{min }& -11.600000 & -8.710000 & -9.332000 \\
\textbf{25\%} & -4.300000 & -4.570000 & -3.955000 \\
\textbf{50\%} & -2.900000 & -3.000000 & -2.630000 \\
\textbf{75\%} & -1.614000 & -1.456000 & -1.600000 \\
\textbf{max} & 1.580000 & 1.110000 & 1.100000 \\
\bottomrule
\end{tabular}

}
\end{minipage}
\caption{Summary statistics for the distribution of predicted values -- logD (lipophilicity) and logS (solubility)-- in the training, validation and test splits of the lipophilicity and solubility datasets, respectively.}
\label{tab:logs_logd_stats}
\end{table*}

\paragraph{Hyperparameters}
We used the AdamW optimizer with the default settings of $\beta=0.9$ and $\beta=0.999$, $\epsilon=\num{1e-8}$ and weight decay of 0.01, together with a linear decay of the learning rate every 10 epochs with $\gamma=0.1$. 
We performed grid search across different batch sizes and learning rates (presented in \Cref{tab:app_hyperparam_ranges}).
We trained the pre-trained for up to 10 epochs. 
Note, that we did not conduct a separate hyperparameter search for randomly initialized models and instead, used the optimal hyperparameter values obtained for the corresponding pre-trained models. 
However, we adjusted the maximum number of training epochs for randomly initialized models to 20 epochs. 
In addition, because the pre-trained \texttt{chemberta-2} models share the same architecture, we fine-tuned only a single randomly initialized \texttt{chemberta-2} model, marked with * in \Cref{tab:app_final_hyperparams}. 
The final hyperparameters are presented in \Cref{tab:app_final_hyperparams}. 


\begin{table*}[ht]
\centering
\small
\begin{tabular}{lllllll}
\toprule
\textbf{Task} & \textbf{Regression Head} & \textbf{Batch size} & \textbf{Learning rate} & \multicolumn{2}{c}{\textbf{\# Epochs}} \\ 
\cmidrule(lr){3-4}

& & PT & RI \\
\midrule
Lipo & LR & $[8, 16, 32, 64]$ & $[0.01,  \num{1e-3}, \num{1e-4}, \num{5e-4}, \num{1e-5}, \num{5e-5}]$ & 10 & 20\\
ESOL & LR, 2L-MLP & $[8, 16, 32, 64]$ & $[0.01,  \num{1e-3}, \num{1e-4}, \num{5e-4}, \num{1e-5}, \num{5e-5}]$ & 20 & 20\\
\bottomrule
\end{tabular}
\caption{Hyperparameter ranges explored. For randomly initialized models, we did not perform extensive hyperparameter tuning, instead adopting the optimal batch size and learning rates from their pre-trained counterparts. However, we increased the number of epochs. For ESOL, we evaluate both, linear regression (LR) and 2-layer MLP (2L-MLP) heads.} 
\label{tab:app_hyperparam_ranges}
\end{table*}

\begin{table*}[h]
\centering
\begin{tabular}{lllllll}
\toprule
\textbf{Model} & \multicolumn{2}{c}{\textbf{Batch size}} & \multicolumn{2}{c}{\textbf{Learning rate}} & \multicolumn{2}{c}{\textbf{Best Epoch}}\\ 
\cmidrule(lr){2-3}\cmidrule(lr){4-5}\cmidrule(lr){6-7}
&PT & RI &PT & RI & PT & RI \\
\midrule
\texttt{chemberta-base} &16 &16 & \num{1e-4} & \num{1e-4} & 8 & 18\\
\texttt{chemberta} & 16 &16   & \num{1e-5} & \num{1e-5} & 9 & 10 \\
\midrule
\texttt{chemberta-2-5M} & 8  & 32* & \num{5e-5}& \num{5e-4}* & 8 & 20* \\
\texttt{chemberta-2-10M} & 32 & 32* & \num{5e-4}& \num{5e-4}* & 10 & 20*\\
\texttt{chemberta-2-77M} &  32 & 32* & \num{5e-4}& \num{5e-4}* & 7 & 20*\\

\midrule
\texttt{molformer} & 16 &16  &  \num{1e-4} & \num{1e-4}  & 10 & 20 \\
\texttt{roberta-zinc-480m}  &  32 & 32  & \num{1e-5} & \num{1e-5}  & 7 & 17 \\
\texttt{chemberta-3} & 64 & 64    & \num{1e-4}  & \num{1e-4}  & 10 & 10  \\
\bottomrule
\end{tabular}
\caption{Final hyperparameter values for pre-trained and randomly initialized models fine-tuned for \textbf{lipophilicity}. Both pre-trained and ranodmly initalized models share the hyperparameters except for the best epoch. Note that \texttt{chemberta-2-5M}, \texttt{chemberta-2-10M}, \texttt{chemberta-2-77M} models share the same architecture. Thus, we fine-tune a single randomly initialized model for all of them (\texttt{chemberta-2}), selecting the average optimal hyperparameters across all three pre-trained models (*).}
\label{tab:app_final_hyperparams}
\end{table*}

\paragraph{Probing Setup}

To conduct meaningful analyses of the effect of fine-tuning on molecular substructures, we excluded any molecular substructures that appear fewer than ten times in either the training or test split of the lipophilicity dataset. 
\Cref{tab:lipo_stats} summarizes the final set of 60 molecular substructures, along with their relative frequencies in the training, validation and test splits of the lipophilicity dataset.
Because our fine-tuning analyses categorize molecular substructures as either relevant (hydrophilic and lipophilic) or non-relevant (other) (\Cref{sec:lipo_prediction}), \Cref{tab:lipo_stats} also indicates the group assignment (hydrophilic, lipophilic, other) for each molecular substructures.

\subsection{Solubility}

\paragraph{Solubility dataset}
We fine-tune all models on the ESOL solubility dataset from the MoleculeNet benchmark~\citep{wu2018moleculenetbenchmarkmolecularmachine}. 
The dataset comprises 1,127 molecules and their corresponding logS values (log solubility in mols per liter). 
Notably, the dataset is considerably smaller than the lipophilicity one (1,128 vs 4,200 molecules). 
We use the train-validation-test splits (80/10/10) provided by \citet{ross2022large}. 
\Cref{fig:app_plot_logs_distr} illustrates the distribution of logS values across the training, validation and test splits of the dataset. 

As can be seen, the distribution of logS values differs substantially between different splits of the datasets; considerably more than in the lipophilicity prediction dataset.

Most notable is the tail difference around a logS value of -8, which occur more frequently in the validation set ($\sim7.5\%$)  compared to the training and the test sets ($\leq2.5\%$).
As shown in \Cref{fig:app_distr_logd_logs}, while the  train and test means are centered around -3 (-3.05 and -2.92, respectively), the validation set has a slightly lower mean of -3.19. Furthermore, the validation set exhibits greater dispersion (standard deviation $\sigma=2.30$) compared to the training ($\sigma=2.08$) and test ($\sigma=2.06$) sets. 
This might explain lower performance of almost all models on the ESOL dataset, as they were selected on a validation set with a different distribution compared to the training and test sets.

\paragraph{Hyperparameters}
We largely followed the fine-tuning setup described for lipophilicity in the previous section and evaluate the same hyperparameter ranges (\Cref{tab:app_hyperparam_ranges}).
We only adjusted the number of training epochs to 20, as our initial experiments showed a slower convergence of models on this task.
We furthermore used two types of prediction heads: a linear one and a two-layer MLP.
\Cref{tab:app_final_hyperparams_esol} lists the final hyperparameter values for fine-tuning the models on the ESOL dataset.

\begin{table*}[ht]
\centering
\begin{tabular}{llllllll}
\toprule
\textbf{Model} & \textbf{Head type} & \multicolumn{2}{c}{\textbf{Batch size}} & \multicolumn{2}{c}{\textbf{Learning rate}} & \multicolumn{2}{c}{\textbf{Best Epoch}}\\ 
 \cmidrule(lr){3-4}\cmidrule(lr){5-6}\cmidrule(lr){7-8}
& &PT & RI &PT & RI & PT & RI \\
\midrule
\texttt{chemberta-base}& linear & 16 & 16& \num{1e-4} & \num{1e-4} & 19 & 18\\
\texttt{chemberta}& linear & 8 &  8 & \num{5e-5} & \num{5e-5} & 12 &  17\\
\midrule
\texttt{chemberta-2-5M}& linear & 64  & 8* & \num{0.001}&  \num{5e-4}* & 19 & 19* \\
\texttt{chemberta-2-10M}& MLP & 8  & 8* & \num{5e-4}&  \num{5e-4}* & 20& 19*\\
\texttt{chemberta-2-77M}& linear & 8  & 8* & \num{5e-4}&  \num{5e-4}* & 6 & 19*\\

\midrule
\texttt{molformer} & linear& 16& 16&  \num{5e-5}  & \num{5e-5}  & 18 & 19\\
\texttt{roberta-zinc-480m} & linear & 32  & 32 & \num{5e-5} & \num{5e-5}  & 16 &  20\\
\texttt{chemberta-3} & linear& 32 & 32 & \num{1e-4}  & \num{1e-4}  & 19 & 20  \\
\bottomrule
\end{tabular}
\caption{Final hyperparameter values for pre-trained and randomly initialized models fine-tuned on \textbf{ESOL}. Both pre-trained and ranodmly initalized models share the hyperparameters except for the best epoch. Note that \texttt{chemberta-2-5M}, \texttt{chemberta-2-10M}, \texttt{chemberta-2-77M} models share the same architecture. Thus, we fine-tune a single randomly initialized model for all of them (\texttt{chemberta-2}), selecting the average optimal hyperparameters across all three pre-trained models (*).}
\label{tab:app_final_hyperparams_esol}
\end{table*}

\paragraph{Probing Setup}
Our probing setup for solubility follows the general probing setup described in \Cref{sec:experimental-setup} and \Cref{sec:app_probing_setup}. 
Similar to lipophilicity prediction (cf. \Cref{sec:app_ft_setup_lipo}), we exclude probing tasks for substructures appearing fewer than ten times in either the training or test split of the ESOL dataset. 
As a result, our ESOL fine-tuning analyses are based on 39 molecular substructures listed in \Cref{tab:esol_stats}.

\subsection{Lipophilicity and Solubility: Differences}\label{sec:task_differences_app}
After preprocessing, we analyze both datasets in terms of the distribution of molecular substructures as well as SMILES strings (i.e., molecules). 

\paragraph{Overlap in molecular substructures} We observe that the distributions of substructures between the lipophilicity and ESOL datasets are quite different, making a direct comparison on a common set of substructures difficult. 
For example, $\sim90\%$ molecules in the lipophilicity dataset contain \texttt{AromaticHeterocycles}, whereas \texttt{AromaticHeterocycles} are found in only 10\%-15\% of molecules in ESOL. 
Some substructures appearing frequently in the lipophilicity dataset (e.g. \texttt{pyridine}, \texttt{COO} and \texttt{methoxy}) are absent from the ESOL dataset altogether, and vice versa (\texttt{unbrch-alkane}, \texttt{imide} and \texttt{allylic-oxid}). 
We note that this also results in slight differences for groups of hydrophilic, lipophilic and other molecular substructures (cf. \Cref{tab:lipo_stats} and \Cref{tab:esol_stats}).

\paragraph{Overlap in SMILES} We further inspected the lipophilicity and solubility datasets for overlapping SMILES strings, which would have enabled us to investigate how the two tasks influence the encoding of substructures present in a molecule. 
In particular, if the effects are reversed. 
However, we identified only 37 overlapping molecules in the training sets and a single one in the test sets across the two datasets.
In light of this, we decide against a direct comparison between the effect of fine-tuning on lipophilicity and ESOL on substructure encoding. 
Instead, we conduct separate analyses for the two datasets, based on 60 and 39 substructures, respectively.

\section{Effect of Fine-Tuning (RQ2)}
\label{sec:ft_effect_apx}

Similar to lipophilicity prediction (\S\ref{sec:finetuning}), we analyse the probing performance on solubility prediction for different groups and individual molecular substructures.

\subsection{Solubility: Group Analysis}


\begin{figure*}[htb]
  \begin{subfigure}[t]{0.5\textwidth}
        \centering
         \includegraphics[width=1.0\linewidth]
         {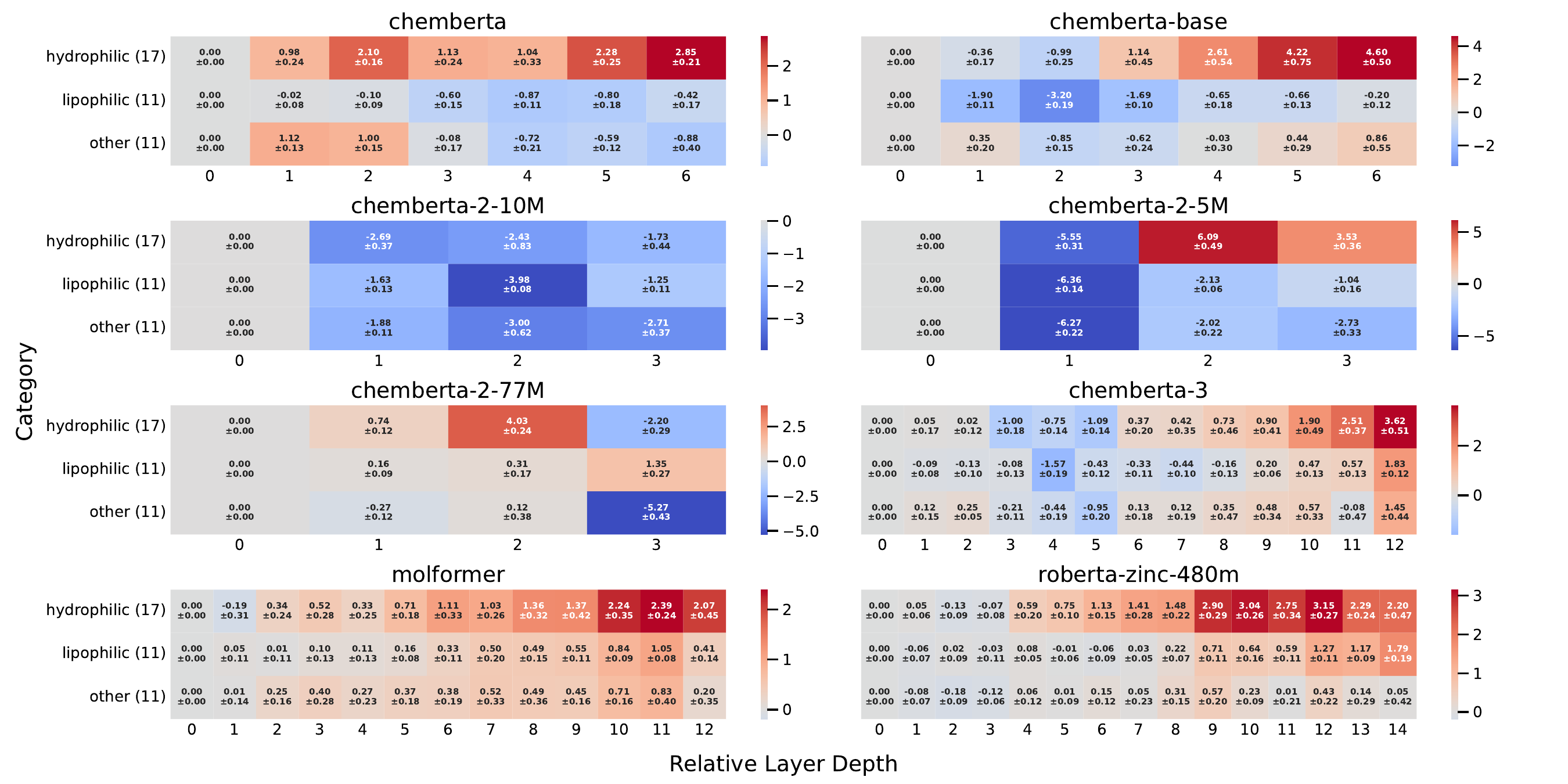}

    \end{subfigure}
  \hfill
  \begin{subfigure}[t]{0.5\textwidth}
        \centering
         \includegraphics[width=1.0\linewidth]
         {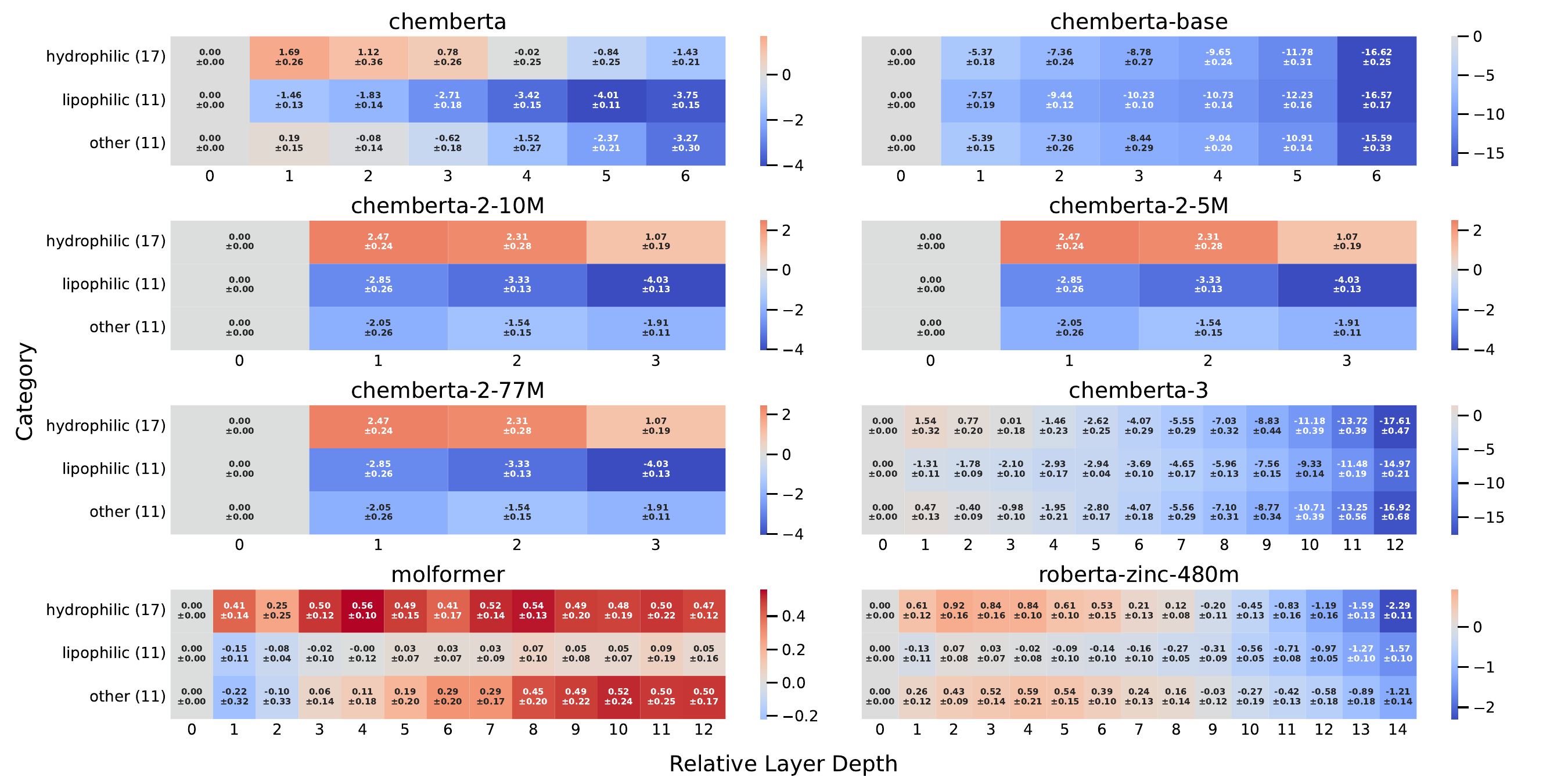}

    \end{subfigure}

    \caption{Average differences in probing performance 
    of PT (left) and RI (right) models after fine-tuning on \textbf{solubility} (ESOL). We group into \emph{hydrophilic} (top), \emph{lipophilic} (middle), and \emph{other} (bottom) groups (see \Cref{tab:esol_stats}). The number of substructures in the corresponding group is indicated next to it. The RI results for \texttt{chemberta-2} are based on a single model (hence, are the same for all three variants) as all models use the same architecture. }
    \label{fig:avg_delta_esol}
\end{figure*}

\Cref{fig:avg_delta_esol} shows the changes (before and after fine-tuning on solubility) in terms of probing performance.
We observe similar trends: fine-tuning leads to larger changes in task-relevant groups (lipophilic and hydrophilic), concentrated in the upper layers. 
In contrast to lipophilicity, fine-tuning on solubility has a more pronounced effect on hydrophilic substructures, as highlighted by the darker \textcolor{red}{red} shading.
This corroborates our lipophilicity results that molecular substructure learning is 
consistent with chemical theory. 
This still holds when we observe \texttt{chemberta-2-10M}, which exhibits a consistent unlearning effect on both downstream tasks:
For solubility prediction, the unlearning effect is stronger for the lipophilic groups while for lipophilicity prediction it is stronger for the hydrophilic groups. 

We can furthermore observe that the effect of fine-tuning on the probing performance is noticeably smaller compared to pre-training (similar as we observed for lipophilicity prediction).
Interestingly, we see several common patterns with respect to how fine-tuning on solubility affects RI and PT models. 
Namely, we find that all RI models improve on hydrophilic substructures during fine-tuning.  
However, in contrast to PT models, these changes are more prominent in lower and middle layers. 

\subsection{Solubility: Individual Analysis}

\begin{figure*}[ht]

  \begin{subfigure}[t]{0.5\textwidth}
        \centering
         \includegraphics[width=1.0\linewidth]{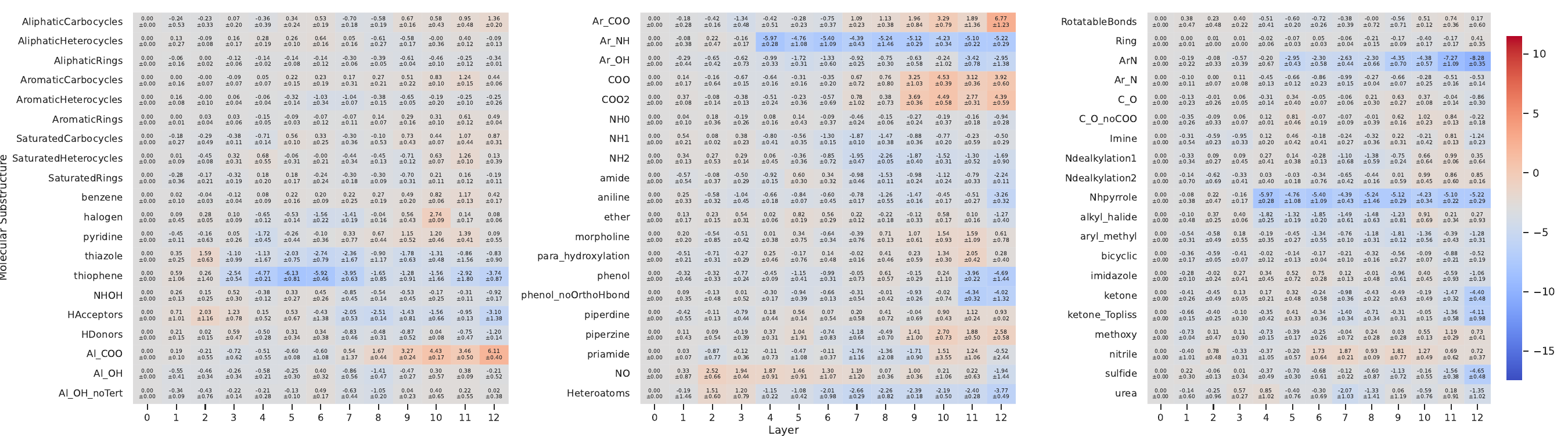}
        \caption{\scriptsize Molformer (PT)}
        \label{fig:molformer-pt}
    \end{subfigure}
    \begin{subfigure}[t]{0.5\textwidth}
        \centering
         \includegraphics[width=1.0\linewidth]{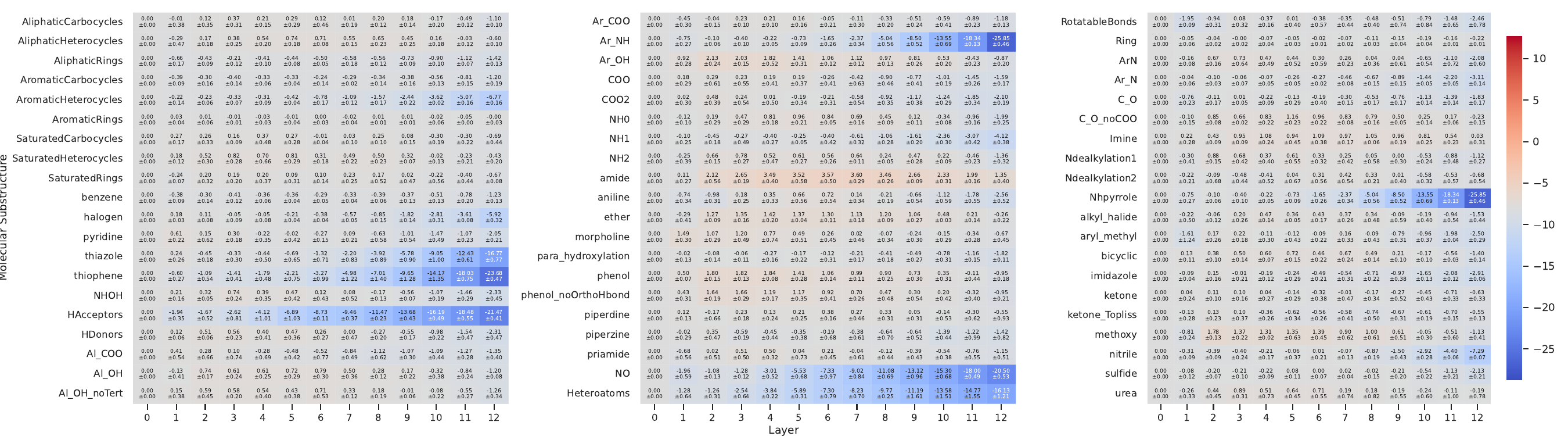}
        \caption{\scriptsize Molformer(RI)}
        \label{fig:molformer-ri}
    \end{subfigure}
  \medskip
  \begin{subfigure}[t]{0.5\textwidth}
        \centering
         \includegraphics[width=1.0\linewidth]{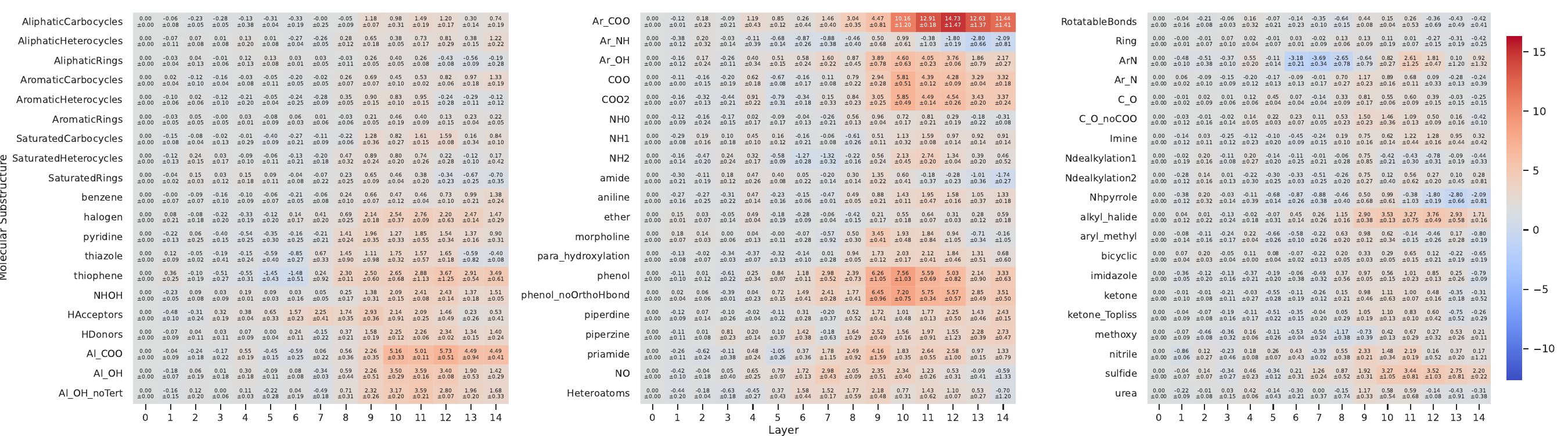}
        \caption{\scriptsize Roberta-zinc-480m (PT)}
        \label{fig:roberta-pt}
    \end{subfigure}
    \begin{subfigure}[t]{0.5\textwidth}
        \centering
         \includegraphics[width=1.0\linewidth]{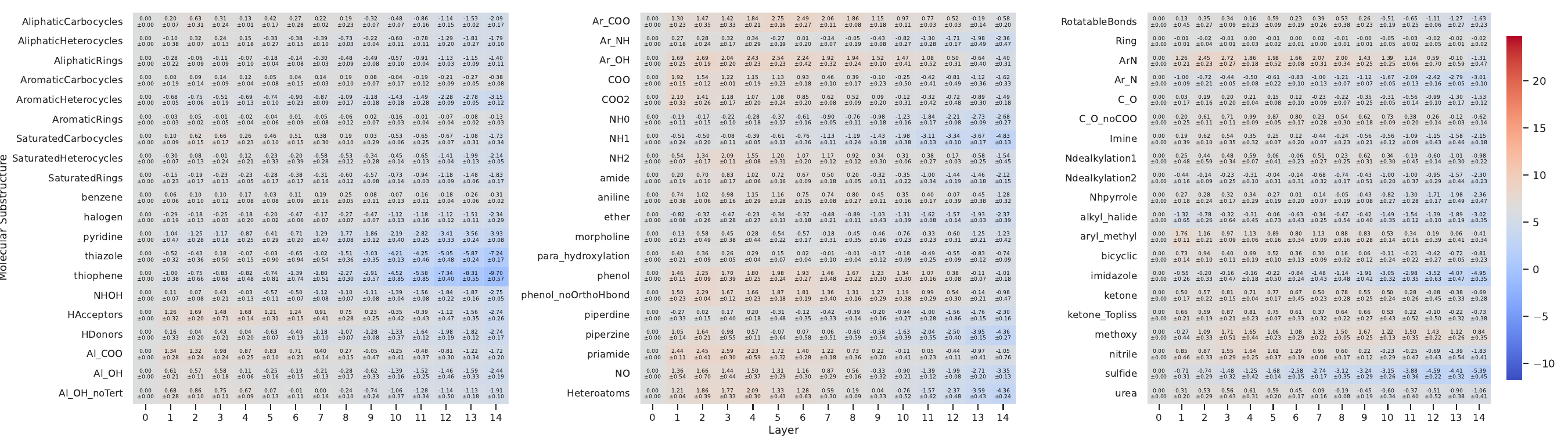}
        \caption{\scriptsize Roberta-zinc-480m (RI)}
        \label{fig:roberta-ri}
    \end{subfigure}
\medskip
  \begin{subfigure}[t]{0.5\textwidth}
        \centering
         \includegraphics[width=1.0\linewidth]{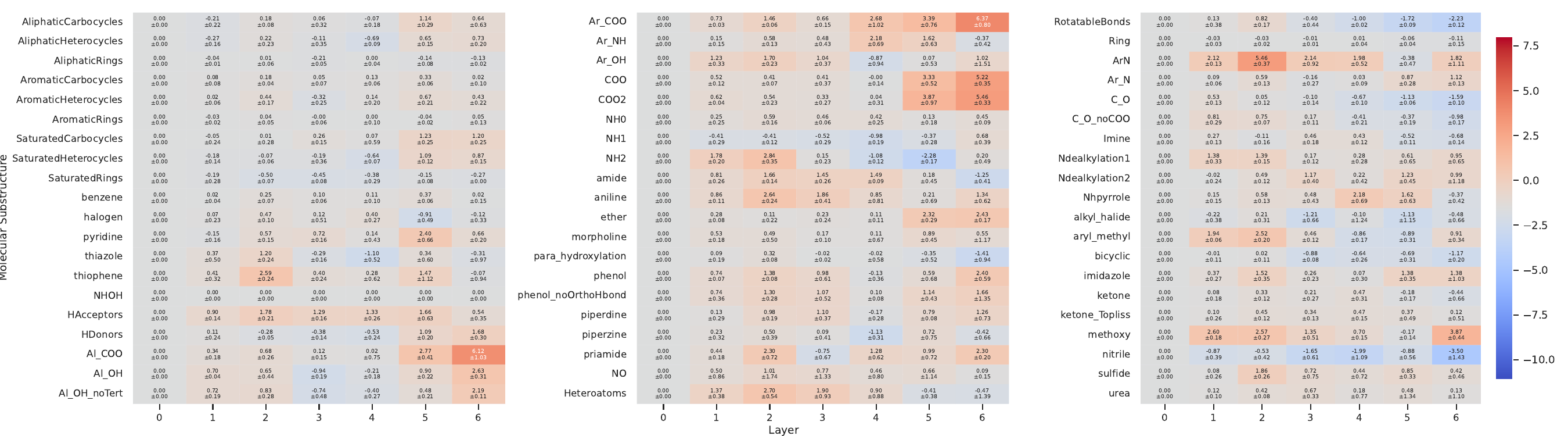}
        \caption{\scriptsize Chemberta (PT)}
        \label{fig:chemberta-pt}
    \end{subfigure}
  \hfill
    \begin{subfigure}[t]{0.5\textwidth}
        \centering
         \includegraphics[width=1.0\linewidth]{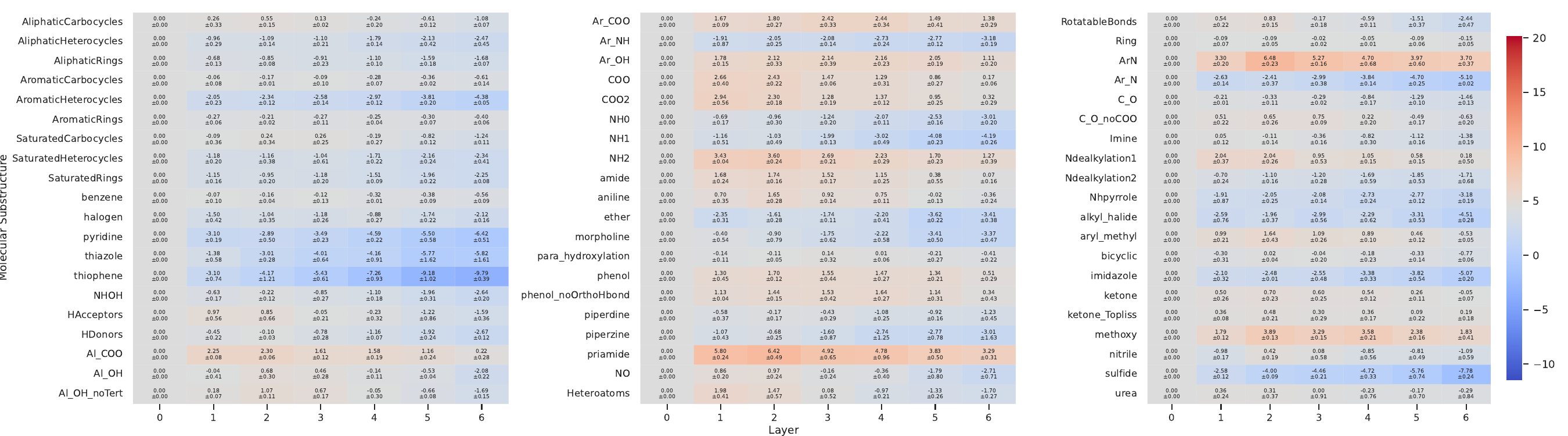}
        \caption{\scriptsize Chemberta (RI)}
        \label{fig:chemberta-ri}
    \end{subfigure}
  \hfill
  \medskip
  \begin{subfigure}[t]{0.5\textwidth}
        \centering
         \includegraphics[width=1.0\linewidth]{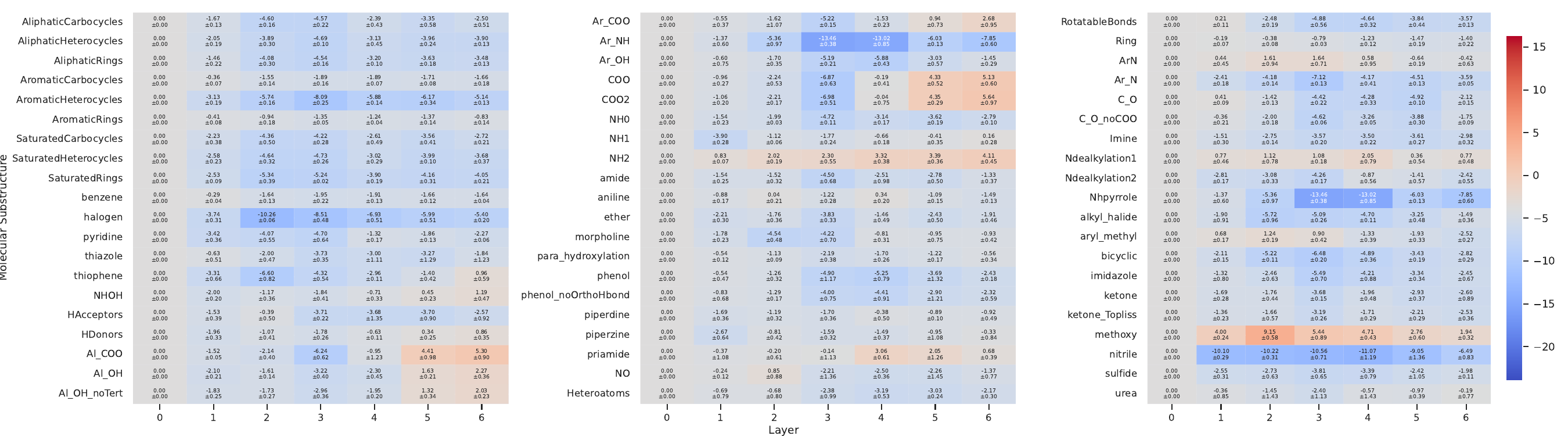}
        \caption{\scriptsize Chemberta-base (PT)}
        \label{fig:chemberta-base-pt}
    \end{subfigure}
     \begin{subfigure}[t]{0.5\textwidth}
        \centering
         \includegraphics[width=1.0\linewidth]{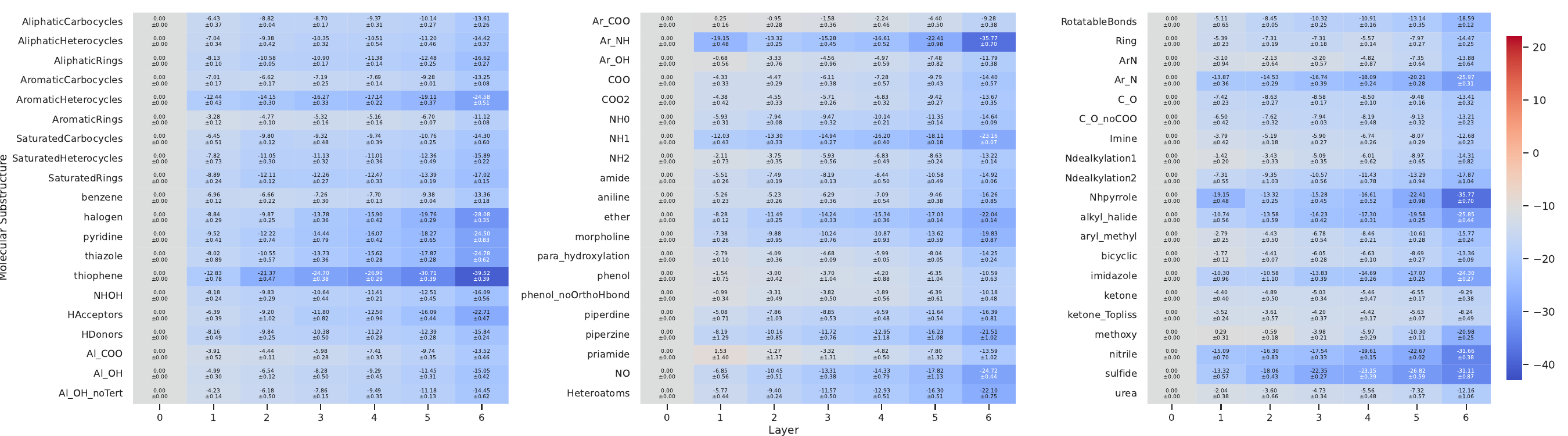}
        \caption{\scriptsize Chemberta-base (RI)}
        \label{fig:chemberta-base-ri}
    \end{subfigure}
    \medskip
  \begin{subfigure}[t]{0.5\textwidth}
        \centering
         \includegraphics[width=1.0\linewidth]{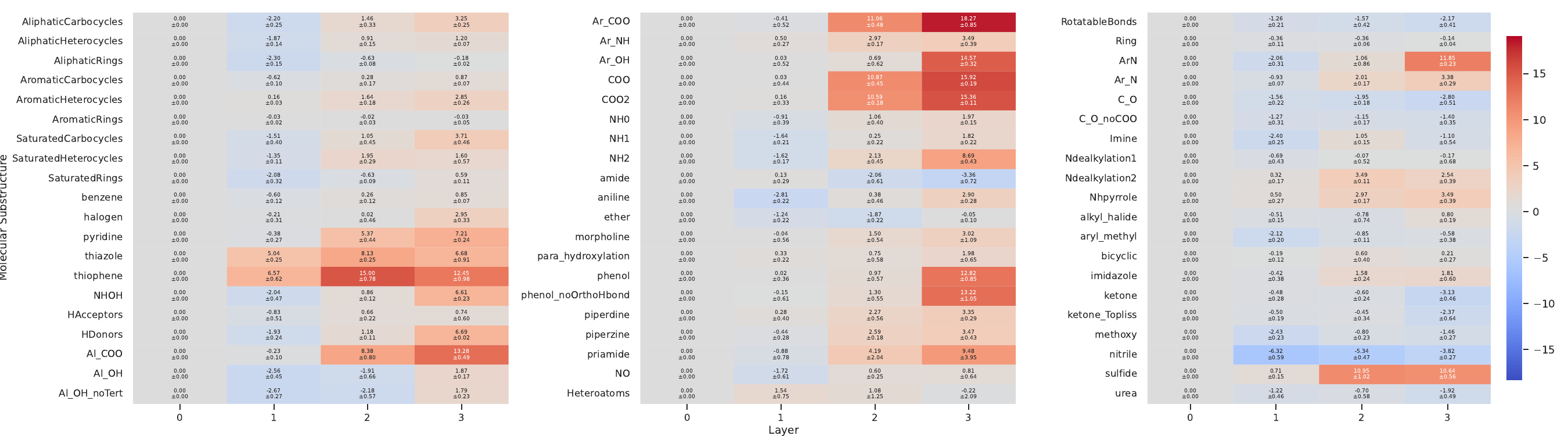}
        \caption{\scriptsize Chemberta-2-5M (PT)}
        \label{fig:ft_delta_cb2_5_PT}
    \end{subfigure}
  \hfill
   \begin{subfigure}[t]{0.5\textwidth}
        \centering
         \includegraphics[width=1.0\linewidth]{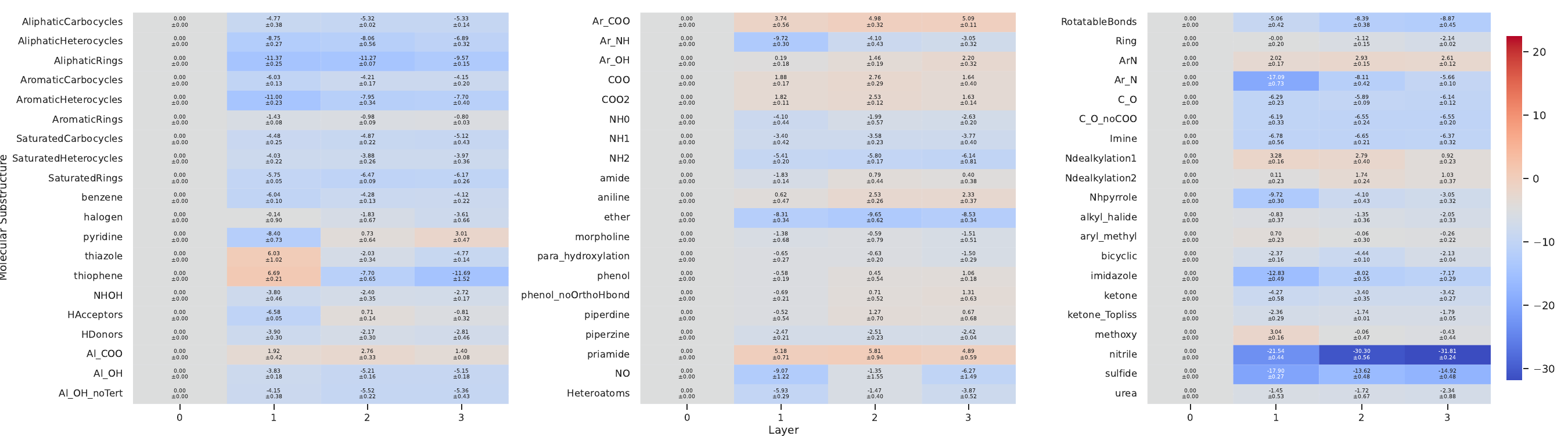}
        \caption{\scriptsize Chemberta-2 (RI)}
        \label{fig:ft_delta_cb2_5_RI}
    \end{subfigure}
  \hfill
  \begin{subfigure}[t]{0.5\textwidth}
        \centering
         \includegraphics[width=1.0\linewidth]{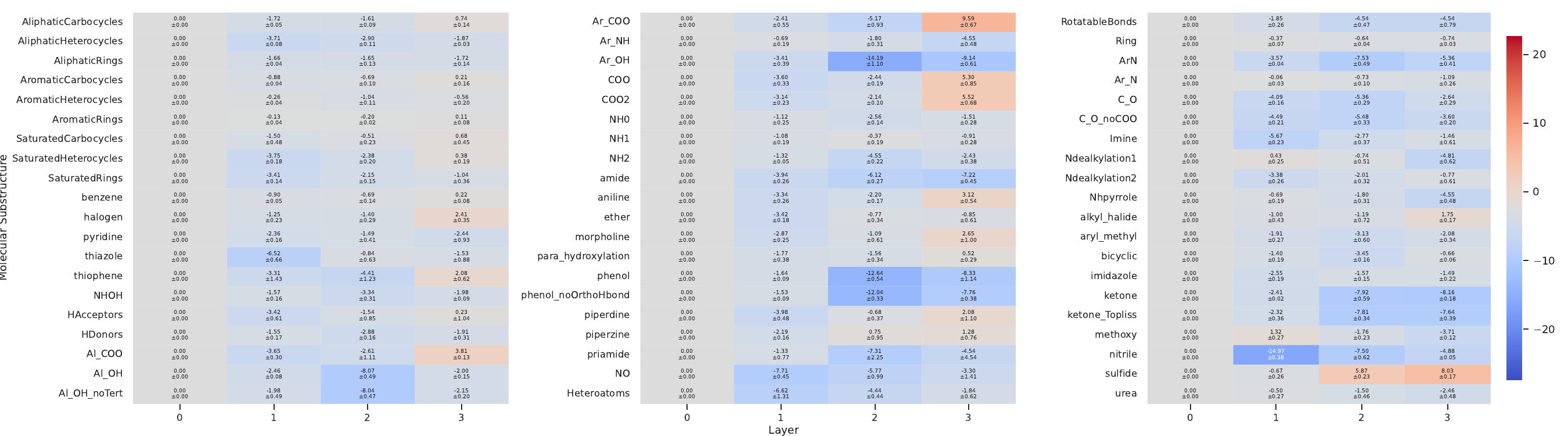}
        \caption{\scriptsize Chemberta-2-10M (PT)}
        \label{fig:ft_delta_cb2_10_PT}
    \end{subfigure}
\hfill
    \begin{subfigure}[t]{0.5\textwidth}
        \centering
         \includegraphics[width=1.0\linewidth]{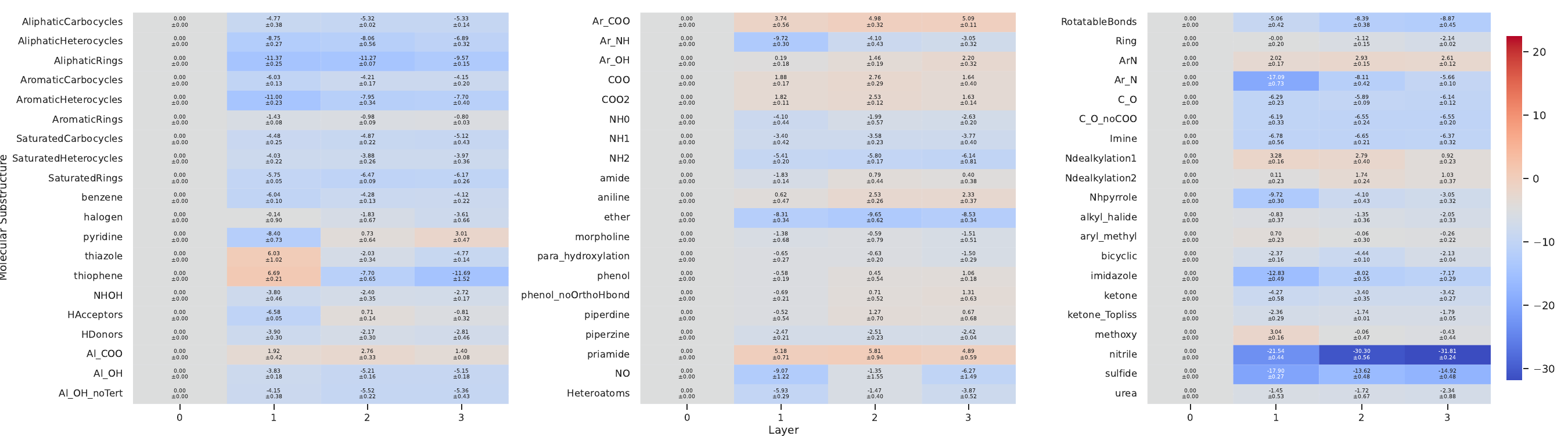}
        \caption{\scriptsize Chemberta-2 (RI)}
        \label{fig:ft_delta_cb2_10_RI}
    \end{subfigure}
     \medskip
  \begin{subfigure}[t]{0.5\textwidth}
        \centering
         \includegraphics[width=1.0\linewidth]{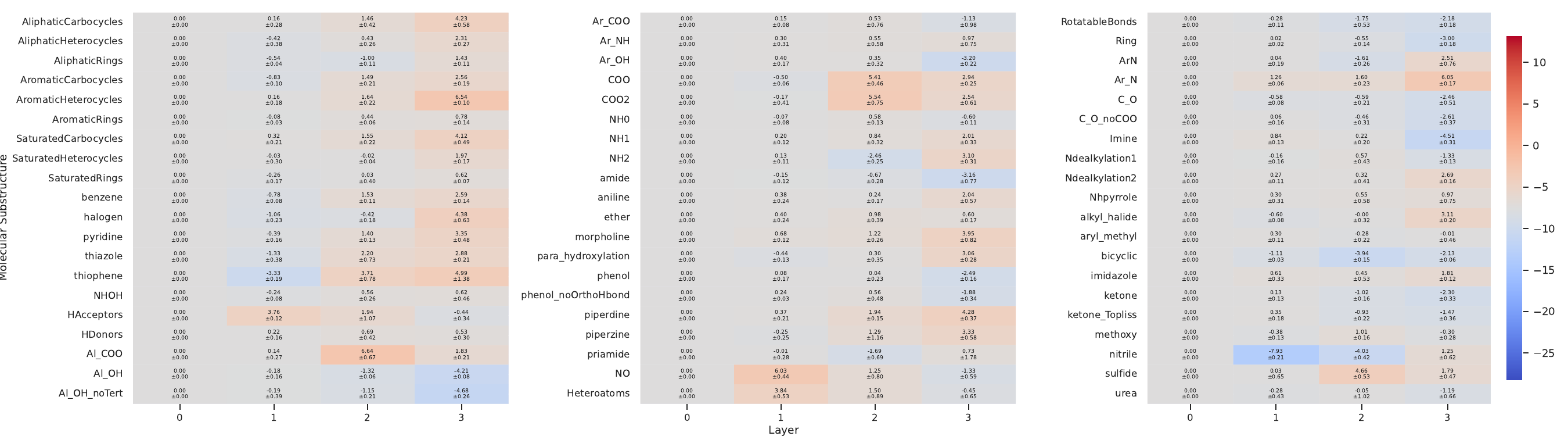}
        \caption{\scriptsize Chemberta-2-77M (PT)}
        \label{fig:ft_delta_cb2_77_PT}
    \end{subfigure}
      \begin{subfigure}[t]{0.5\textwidth}
        \centering
         \includegraphics[width=1.0\linewidth]{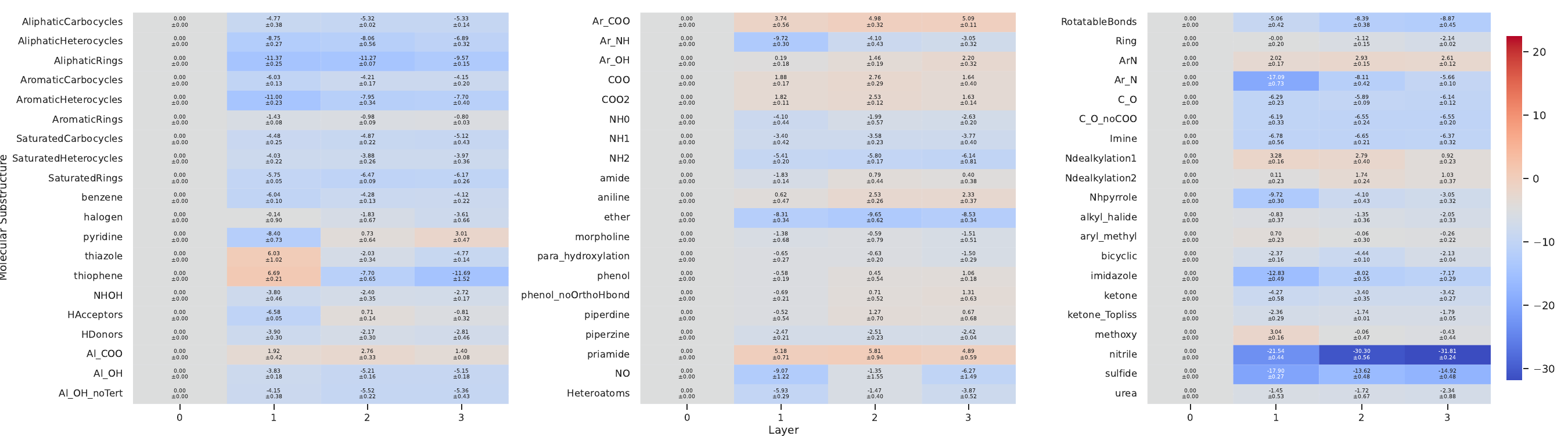}
        \caption{\scriptsize Chemberta-2 (RI)}
        \label{fig:ft_delta_cb2_77_RI}
    \end{subfigure}
       \medskip
  \begin{subfigure}[t]{0.5\textwidth}
        \centering
         \includegraphics[width=1.0\linewidth]{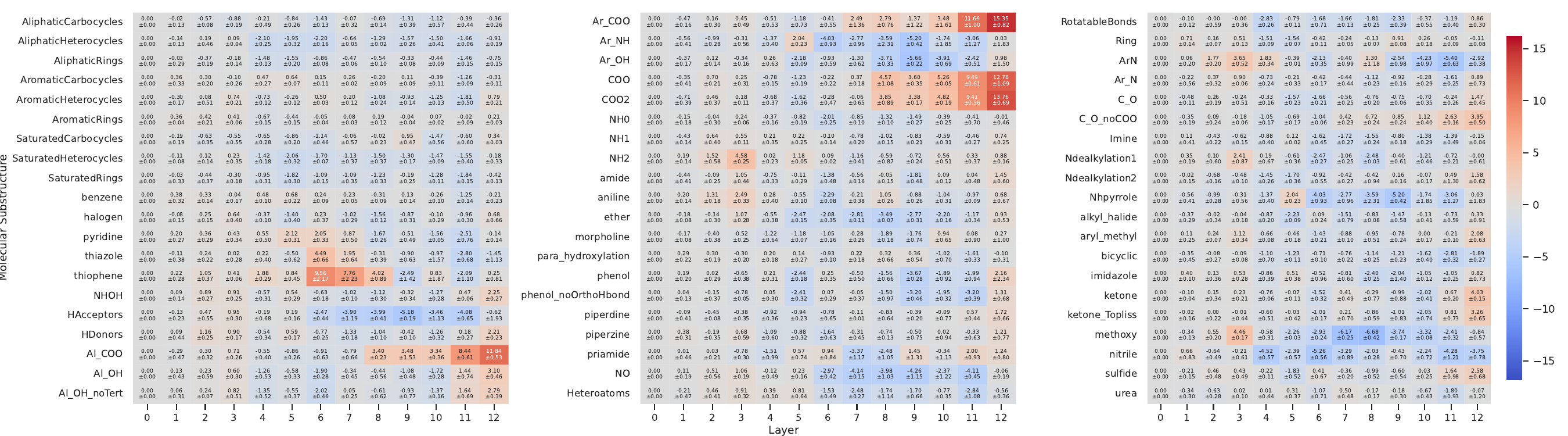}
        \caption{\scriptsize Chemberta-3 (PT)}
        \label{fig:ft_delta_cb3_PT}
    \end{subfigure}
      \begin{subfigure}[t]{0.5\textwidth}
        \centering
         \includegraphics[width=1.0\linewidth]{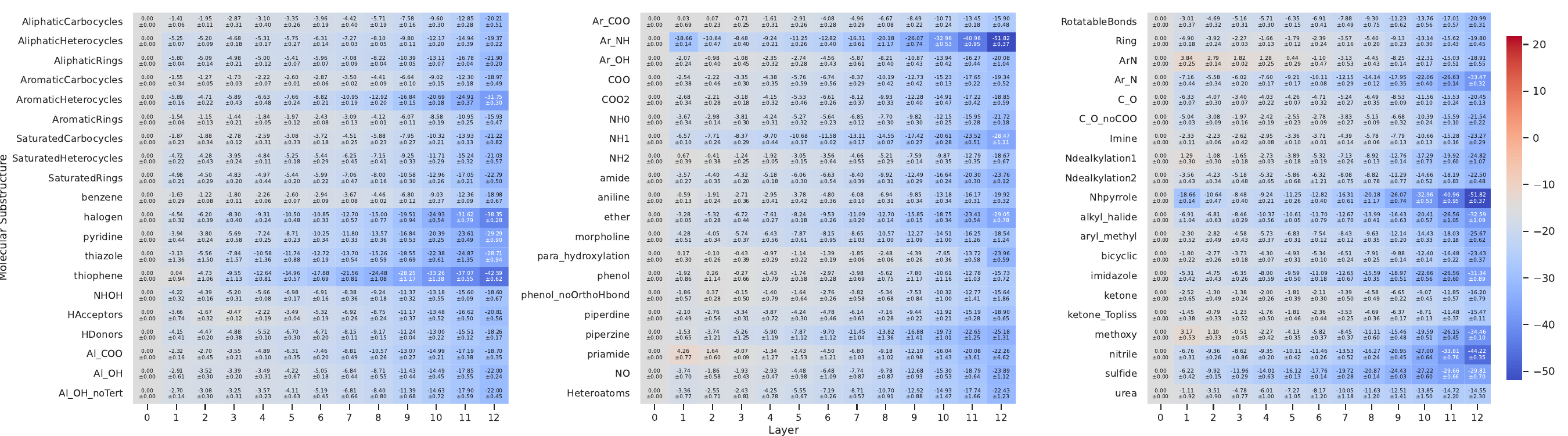}
        \caption{\scriptsize Chemberta-3 (RI)}
        \label{fig:ft_delta_cb3_RI}
    \end{subfigure}
  \hfill
\caption{\textbf{Effect of fine-tuning on lipophilicity on molecular substructure encoding in \clms}. We report the layer-wise difference in probing performance (macro-averaged F1, in \% ) after fine-tuning across 60 tasks (cf. \Cref{tab:lipo_stats}).  \Cref{fig:ft_delta_cb2_5_RI,fig:ft_delta_cb2_10_RI,fig:ft_delta_cb2_77_RI} correspond to the same model architecture. PT denotes to the pre-trained models (left), while RI refers the same models but with randomly initialized weights (right). Improvements in probing performance after fine-tuning are shown in \textcolor{red}{red}, while degradation is indicated in \textcolor{blue}{blue}.}
\label{fig:app_delta_pt}
\end{figure*}

\begin{figure*}[ht]

  \begin{subfigure}[t]{0.5\textwidth}
        \centering
         \includegraphics[width=1.0\linewidth]{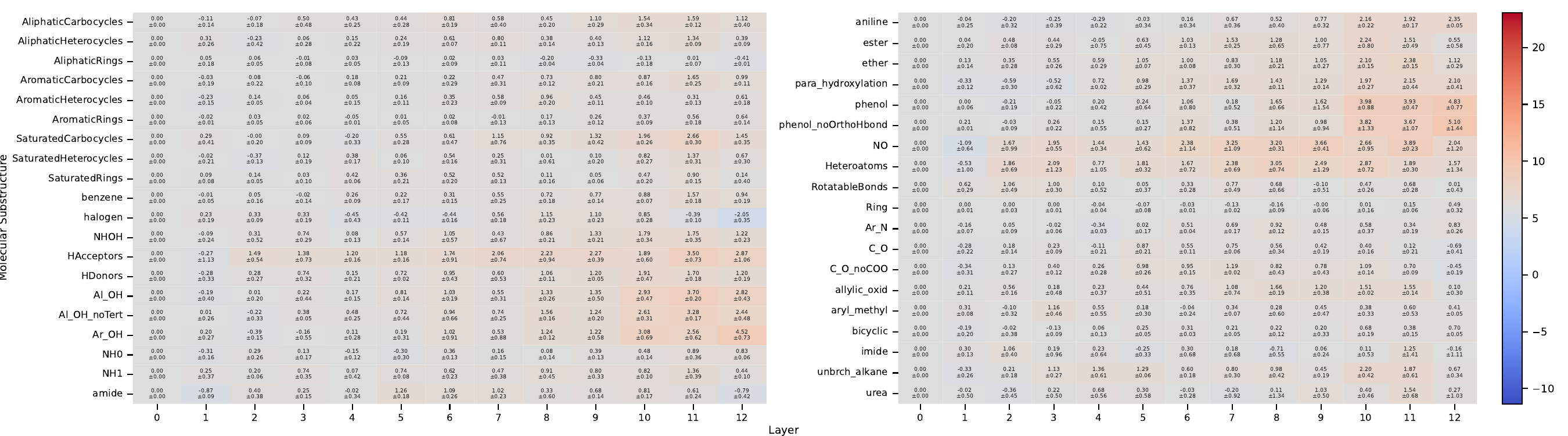}
        \caption{\scriptsize Molformer (PT)}
        \label{fig:molformer-pt_esol}
    \end{subfigure}
    \begin{subfigure}[t]{0.5\textwidth}
        \centering
         \includegraphics[width=1.0\linewidth]{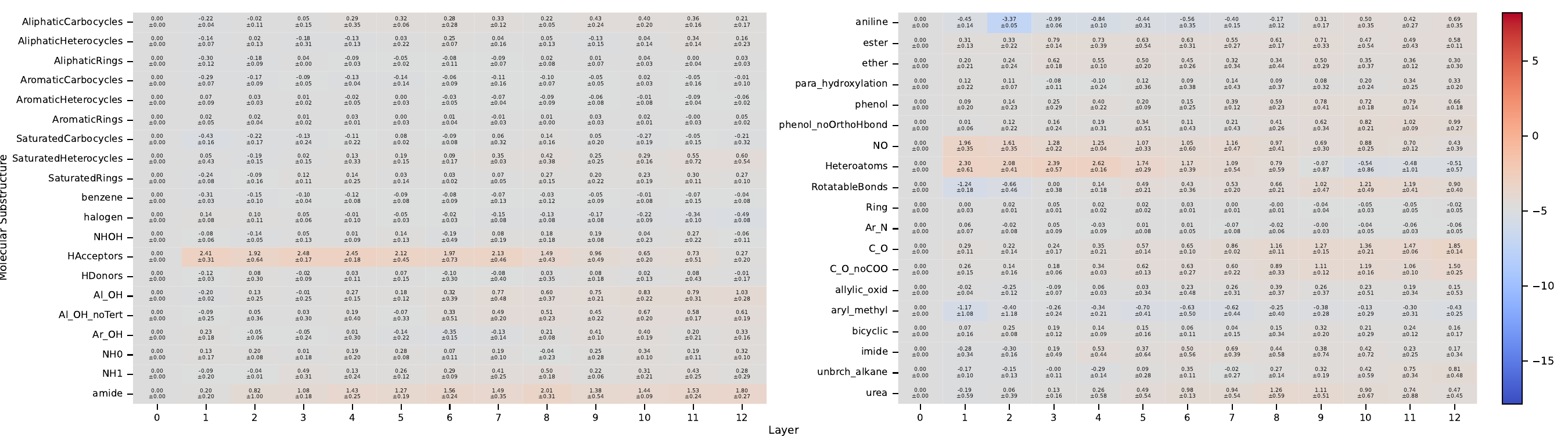}
        \caption{\scriptsize Molformer(RI)}
        \label{fig:molformer-ri_esol}
    \end{subfigure}
  \medskip
  \begin{subfigure}[t]{0.5\textwidth}
        \centering
         \includegraphics[width=1.0\linewidth]{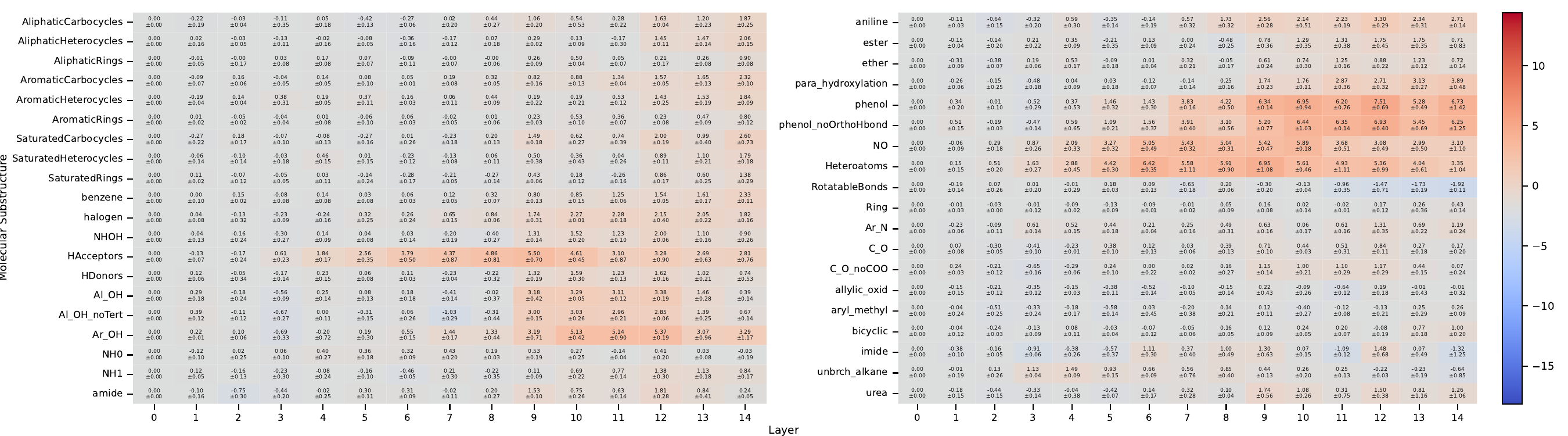}
        \caption{\scriptsize Roberta-zinc-480m (PT)}
        \label{fig:roberta-pt_esol}
    \end{subfigure}
    \begin{subfigure}[t]{0.5\textwidth}
        \centering
         \includegraphics[width=1.0\linewidth]{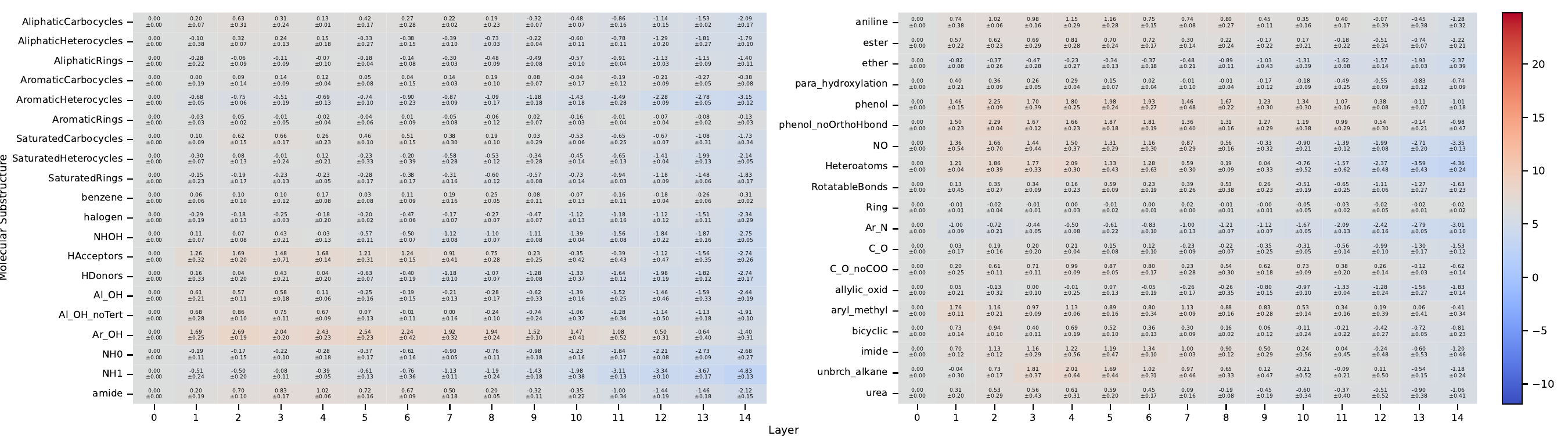}
        \caption{\scriptsize Roberta-zinc-480m (RI)}
        \label{fig:roberta-ri_esol}
    \end{subfigure}
\medskip
  \begin{subfigure}[t]{0.5\textwidth}
        \centering
         \includegraphics[width=1.0\linewidth]{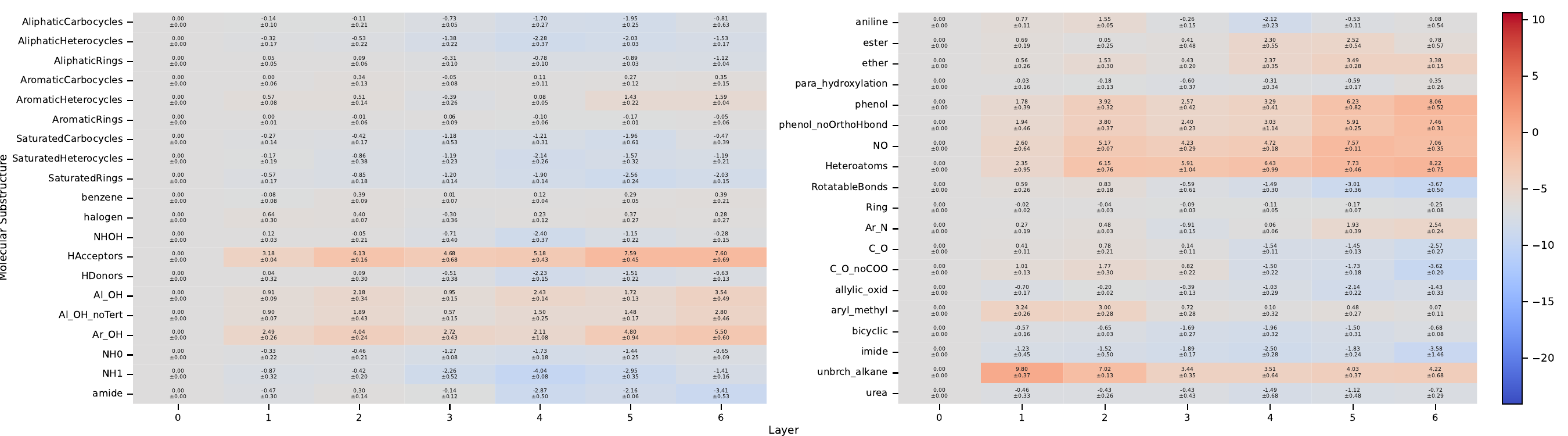}
        \caption{\scriptsize Chemberta (PT)}
        \label{fig:chemberta-pt_esol}
    \end{subfigure}
  \hfill
    \begin{subfigure}[t]{0.5\textwidth}
        \centering
         \includegraphics[width=1.0\linewidth]{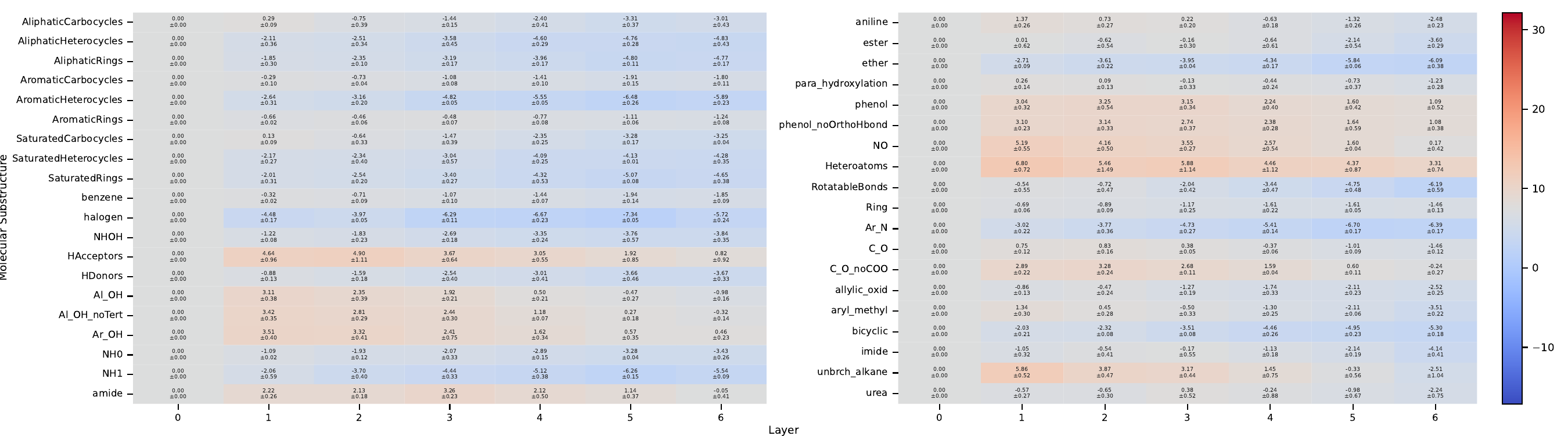}
        \caption{\scriptsize Chemberta (RI)}
        \label{fig:chemberta-ri_esol}
    \end{subfigure}
  \hfill
  \medskip
  \begin{subfigure}[t]{0.5\textwidth}
        \centering
         \includegraphics[width=1.0\linewidth]{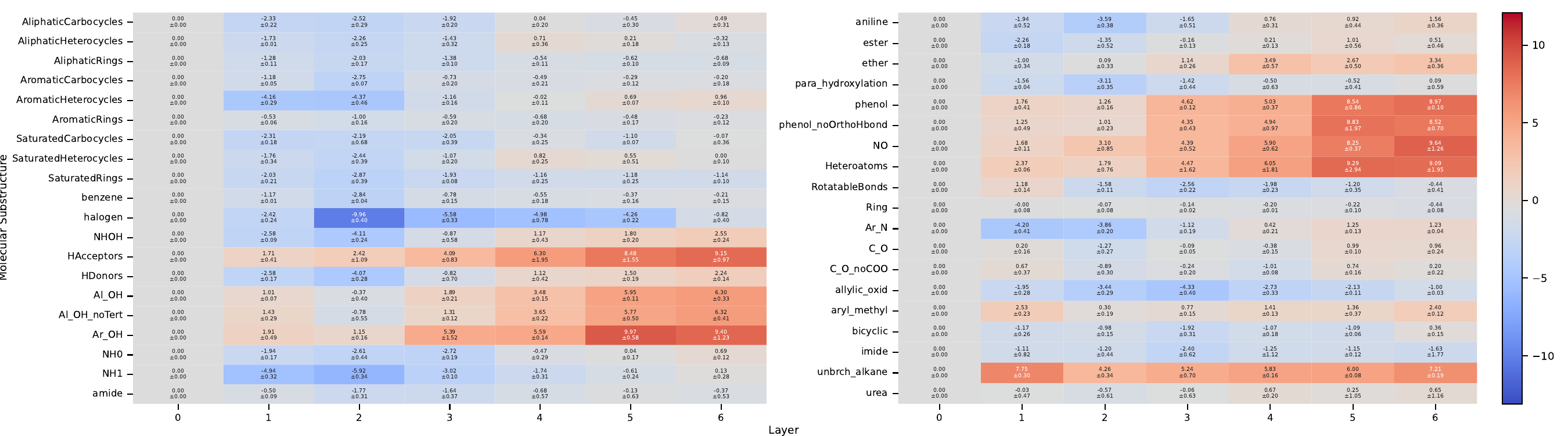}
        \caption{\scriptsize Chemberta-base (PT)}
        \label{fig:chemberta-base-pt_esol}
    \end{subfigure}
     \begin{subfigure}[t]{0.5\textwidth}
        \centering
         \includegraphics[width=1.0\linewidth]{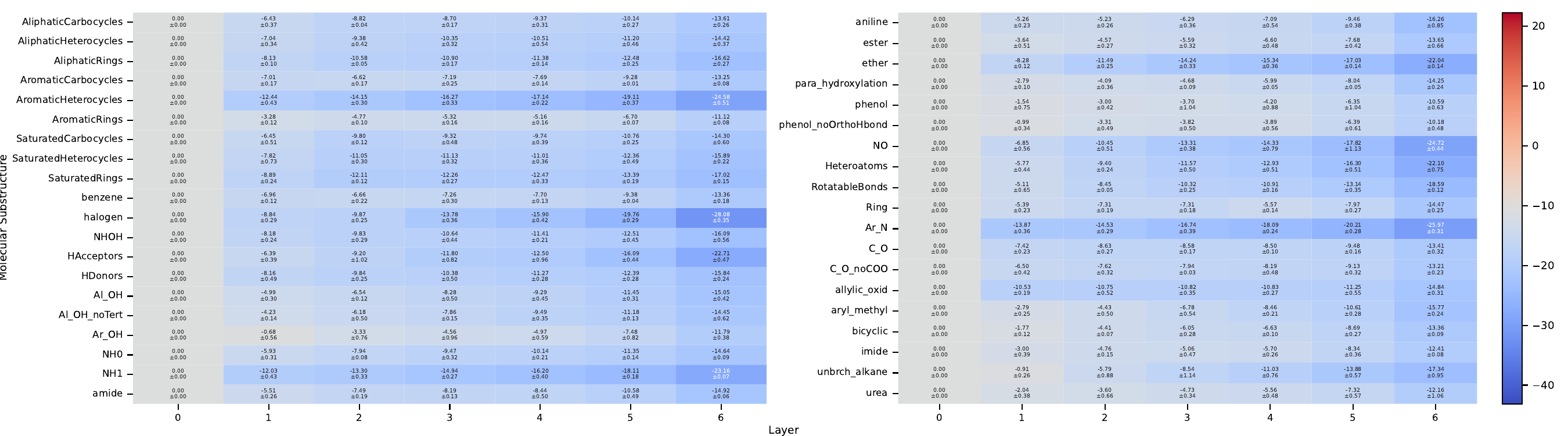}
        \caption{\scriptsize Chemberta-base (RI)}
        \label{fig:chemberta-base-ri_esol}
    \end{subfigure}
    \medskip
  \begin{subfigure}[t]{0.5\textwidth}
        \centering
         \includegraphics[width=1.0\linewidth]{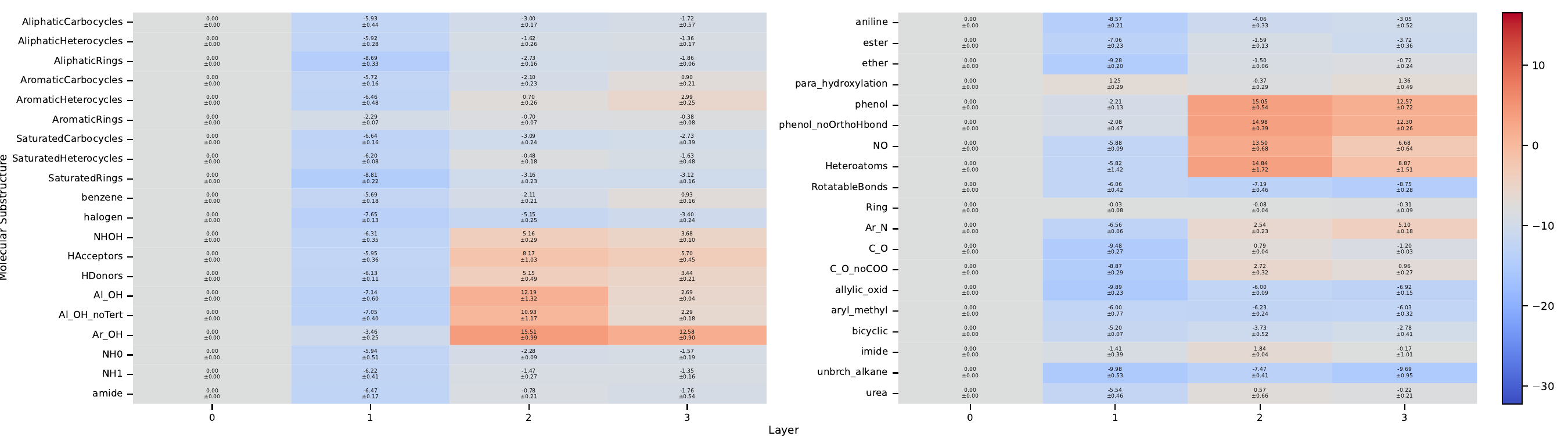}
        \caption{\scriptsize Chemberta-2-5M (PT)}
        \label{fig:ft_delta_cb2_5_PT_esol}
    \end{subfigure}
  \hfill
   \begin{subfigure}[t]{0.5\textwidth}
        \centering
         \includegraphics[width=1.0\linewidth]{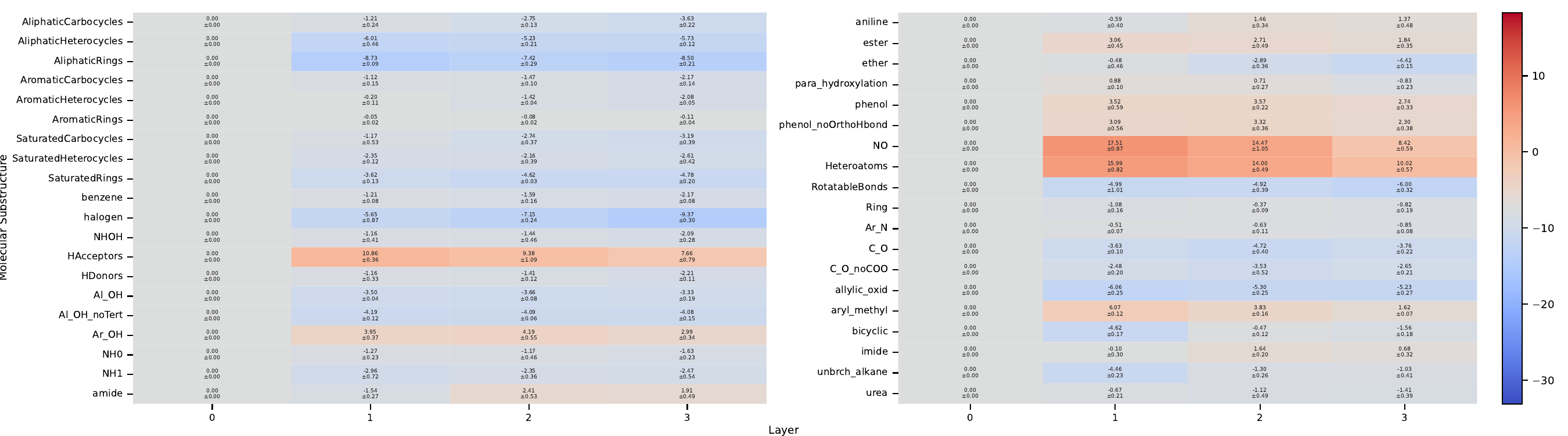}
        \caption{\scriptsize Chemberta-2 (RI)}
        \label{fig:ft_delta_cb2_5_RI_esol}
    \end{subfigure}
  \hfill
  \begin{subfigure}[t]{0.5\textwidth}
        \centering
         \includegraphics[width=1.0\linewidth]{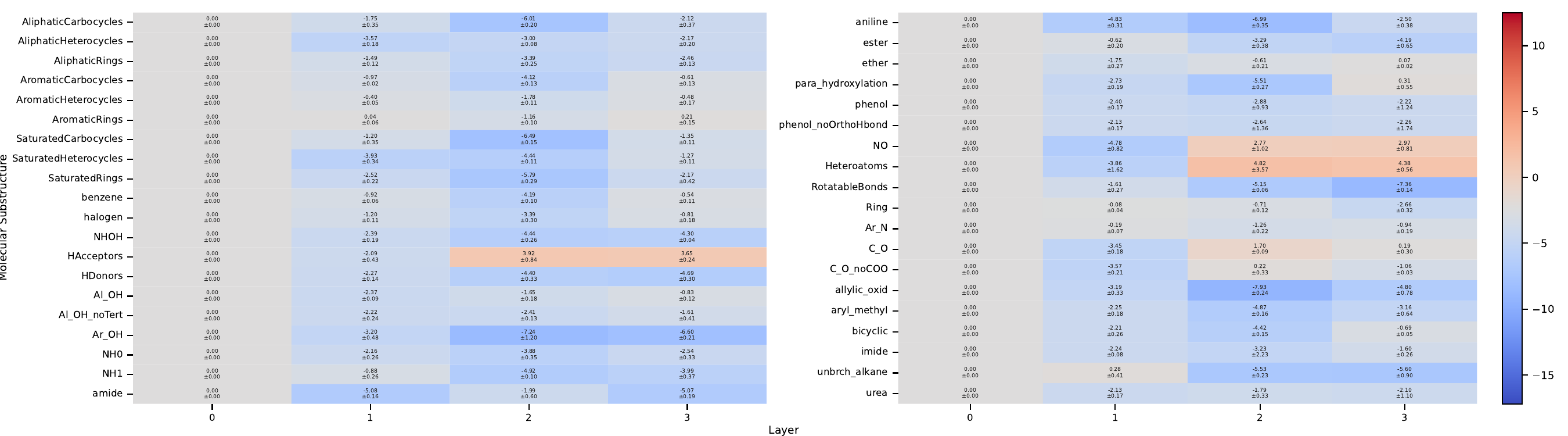}
        \caption{\scriptsize Chemberta-2-10M (PT)}
        \label{fig:ft_delta_cb2_10_PT_esol}
    \end{subfigure}
\hfill
    \begin{subfigure}[t]{0.5\textwidth}
        \centering
         \includegraphics[width=1.0\linewidth]{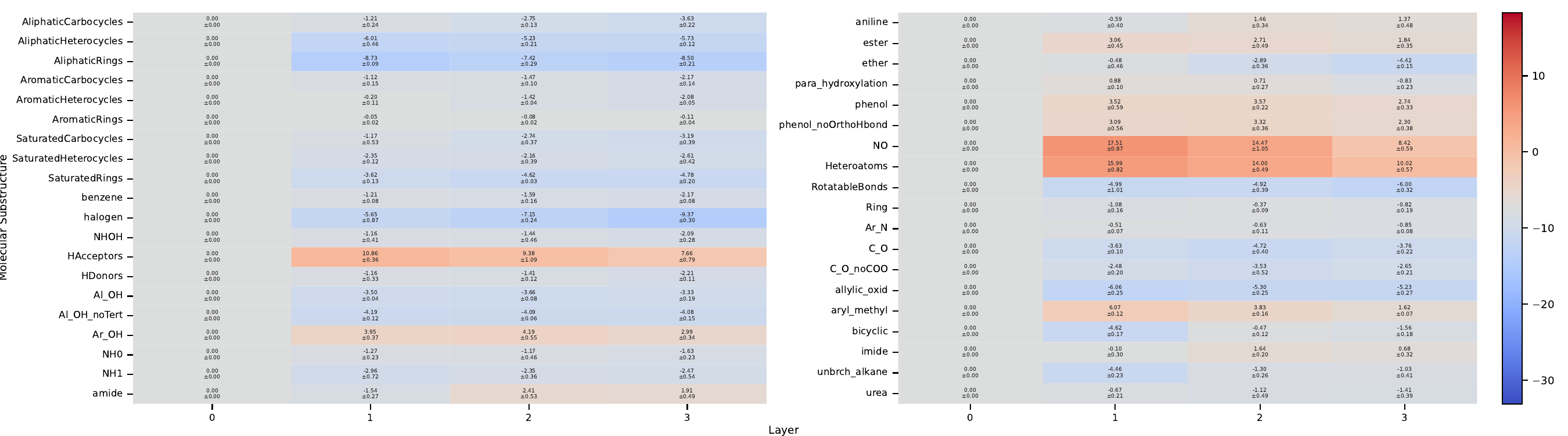}
        \caption{\scriptsize Chemberta-2 (RI)}
        \label{fig:ft_delta_cb2_10_RI_esol}
    \end{subfigure}
     \medskip
  \begin{subfigure}[t]{0.5\textwidth}
        \centering
         \includegraphics[width=1.0\linewidth]{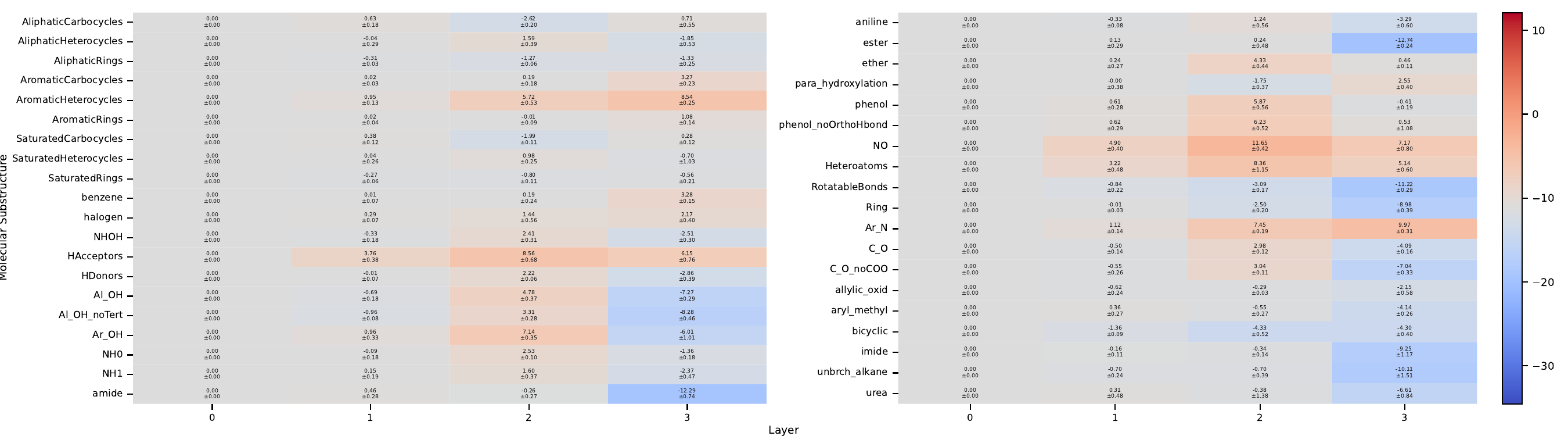}
        \caption{\scriptsize Chemberta-2-77M (PT)}
        \label{fig:ft_delta_cb2_77_PT_esol}
    \end{subfigure}
      \begin{subfigure}[t]{0.5\textwidth}
        \centering
         \includegraphics[width=1.0\linewidth]{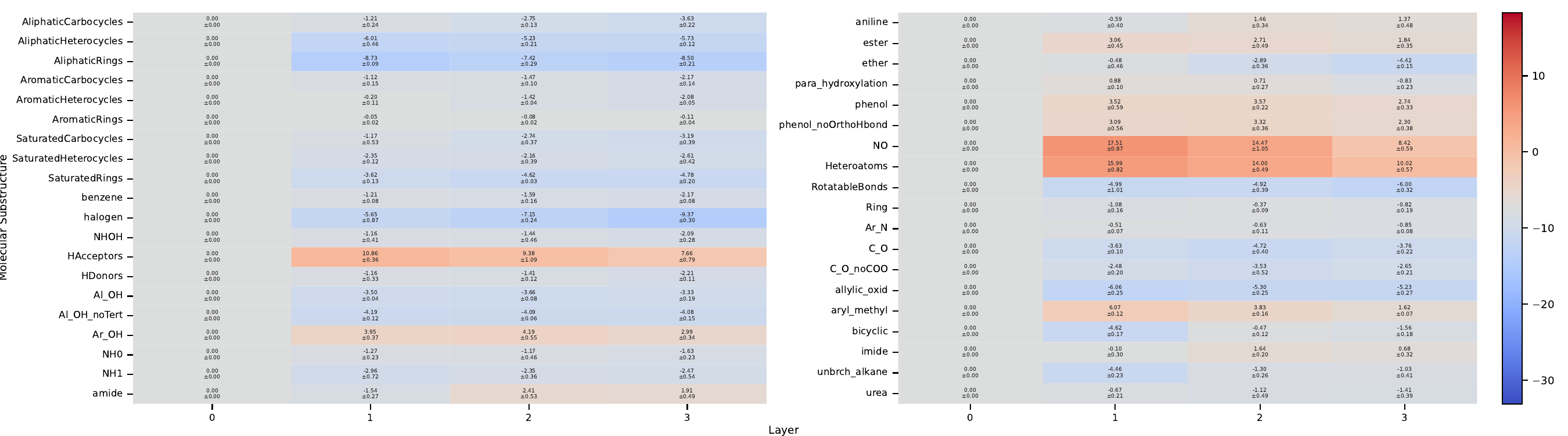}
        \caption{\scriptsize Chemberta-2 (RI)}
        \label{fig:ft_delta_cb2_77_RI_esol}
    \end{subfigure}
       \medskip
  \begin{subfigure}[t]{0.5\textwidth}
        \centering
         \includegraphics[width=1.0\linewidth]{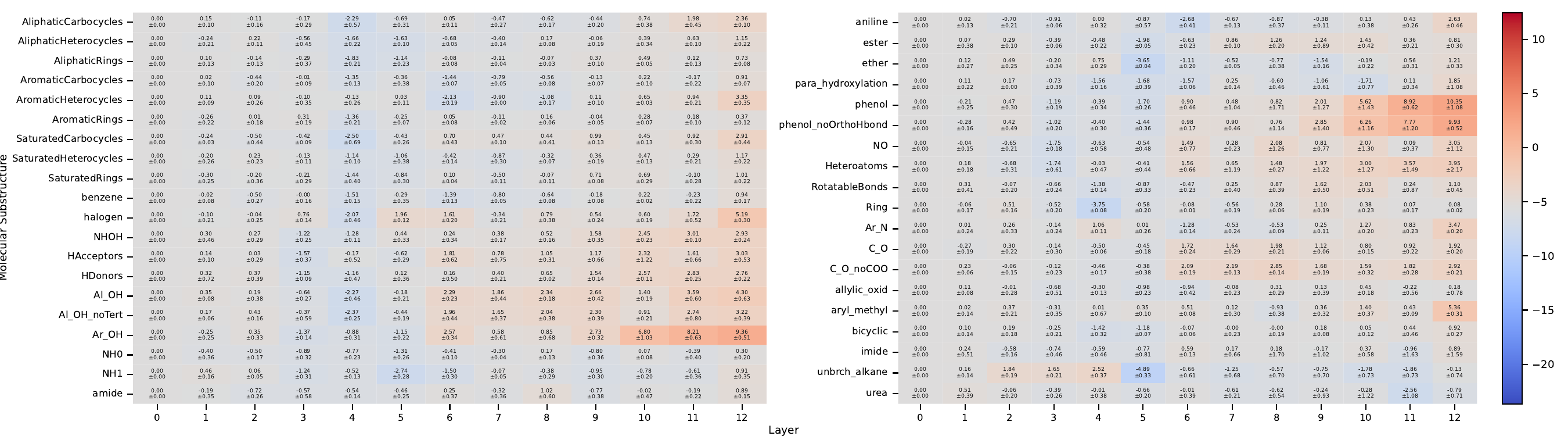}
        \caption{\scriptsize Chemberta-3 (PT)}
        \label{fig:ft_delta_cb3_PT_esol}
    \end{subfigure}
      \begin{subfigure}[t]{0.5\textwidth}
        \centering
         \includegraphics[width=1.0\linewidth]{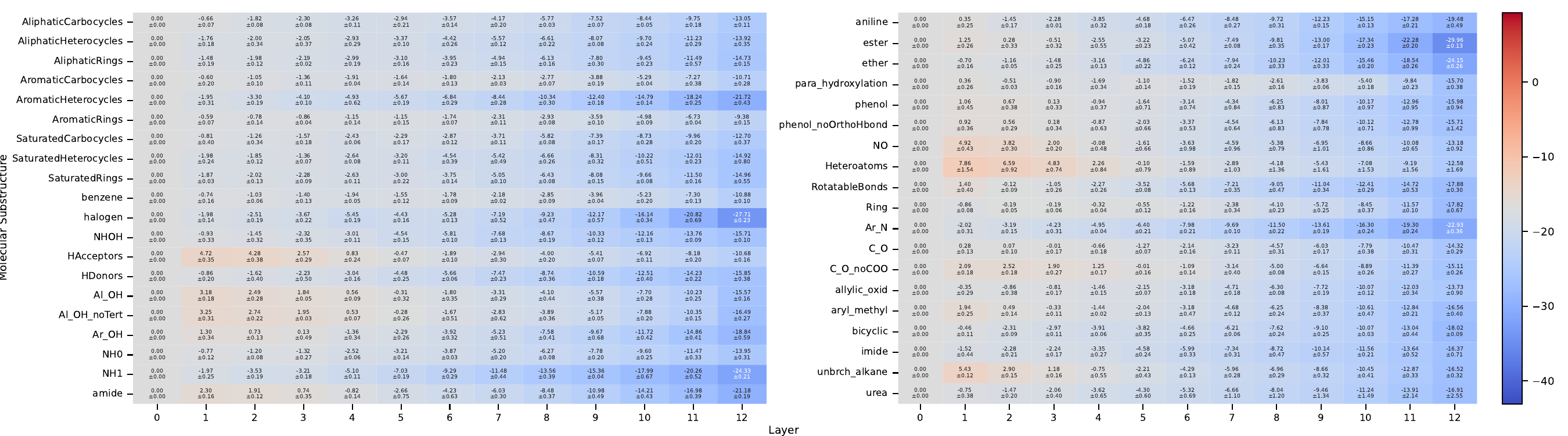}
        \caption{\scriptsize Chemberta-3 (RI)}
        \label{fig:ft_delta_cb3_RI_esol}
    \end{subfigure}
  \hfill
\caption{\textbf{Effect of fine-tuning on solubility (ESOL) on molecular substructure encoding in \clms}. We report the layer-rwise difference in probing performance (macro-averaged F1, in \% ) after fine-tuning across 39 tasks (cf. \Cref{tab:esol_stats}).  \Cref{fig:ft_delta_cb2_5_RI,fig:ft_delta_cb2_10_RI,fig:ft_delta_cb2_77_RI} correspond to the same model architecture. PT denotes to the pre-trained models (left), while RI refers the same models but with randomly initialized weights (right). Improvements in probing performance after fine-tuning are shown in \textcolor{red}{red}, while degradations are indicated in \textcolor{blue}{blue}.}
\label{fig:app_delta_pt_esol}
\end{figure*}

The heatmaps in \Cref{fig:app_delta_pt_esol} present the layer-wise differences in probing performance after fine-tuning on solubility prediction for each substructure (\Cref{fig:app_delta_pt} presents the respective heatmaps for lipophilicity prediction).

First, our results show that fine-tuning on solubility improves encoding of \texttt{NO} (nitrogens and oxygens), \texttt{Heteroatoms} and \texttt{HAcceptors} across all PT models. 
In addition, all models except for \texttt{chemberta-2-10M} exhibit improved probing performance on \textit{phenols} (\texttt{phenol}, \texttt{phenol\_noOrthoHbond}), various \textit{hydroxyl groups} (i.e., aliphatic hydroxyl groups (\texttt{Al\_OH}), aromatic hydroxyl groups (\texttt{Ar\_OH}), aliphatic hydroxyl groups excluding tert-OH (\texttt{Ar\_OH\_noTert})), as well as \textit{aromatic nitrogens} (\texttt{Ar\_N}). 
These results are consistent with chemical theory. 
More specifically, \citet{WALKER20175100} state that replacing a carbon atom with polar heteroatoms N or O is one of the most common approaches to increasing solubility. 
Furthermore, \citet{alma99901018739001842} note that the ability of polar hydroxyl groups (\texttt{OH}) to form hydrogen bonds leads to an increase in aqueous solubility. 
\texttt{Phenols} on the other hand have a moderate effect on solubility due to the presence of a lipophilic aromatic ring and a hydrophilic hydroxyl group. 
In summary, after fine-tuning on solubility, we observe improvements on hydrophilic substructures known to positively contribute to the solubility of a molecule. 
For decreased performance, we do not observe such systematic patterns.


While in general there are no consistent trends for RI models, fine-tuning on solubility prediction leads to an improvement of RI models on \texttt{NO}, \texttt{Heteroatoms}, \texttt{HAcceptors} in lower layers. 
While this might suggest that RI models might be also capable of capturing task-relevant substructures, we do not observe such a trend for lipophilicity prediction.

\section{Practical Implications: Further Pre-Training}\label{sec:appendix_takeaway}

Our probing experiments have allowed us to identify three interesting patterns.
First, we saw that pre-training leads to worse encodings of \texttt{thiophene}, \texttt{thiazole} and \texttt{furan} in all pre-trained models compared to the randomly initialized ones. 
Second, all \texttt{chemberta-2} models use the same architecture and exhibit consistent improvements in probing performance on halogens while we observe the reverse for all other models. 
Finally, we observed diverging patterns in substructure encoding for \texttt{chemberta-2-5M} and \texttt{chemberta-2-10M} which differ only in terms of pre-training data.
We conjecture that these differences might stem from different compositions of the pre-training data and conduct experiments to see if further pre-training models on specific datasets can mitigate the low probing performance.

\paragraph{PT<RI}
In \Cref{sec:pre-training-molecules}, we saw that pretraining degraded probing performance on substructures such as \texttt{thiophene}, \texttt{thiazole} and \texttt{furan} across all models. 
We hypothesize that the models were undertrained on these substructures. 
While we cannot conduct a frequency analysis of these substructures due to the unavailability of the pre-training data, further pre-training models on molecules that contain these molecular substructures should allow models to somewhat recover from the low performance.
For this experiment, we select three models---\texttt{chemberta-2-5M} (3 layers), \texttt{chemberta} (6 layers) and \texttt{molformer} (12 layers)---each representing a small, mid-sized, and large model. 



\paragraph{Chemberta-2 and Halogens}
Our analysis in \Cref{sec:pre-training-molecules} showed that \texttt{chemberta-2-5M} and \texttt{chemberta-2-10M} improved on halogens after pretraining, whereas performance degraded for all other models (which generally showed a very high probing performance). 
Moreover, in \Cref{sec:ft_probing_results} we found that, unlike the other models, fine-tuning on lipophilicity led to further improvements on halogens for \texttt{chemberta-2-5M} and \texttt{chemberta-2-10M}.
We thus hypothesize the following:
\begin{enumerate}

    \item Further pre-training on halogen-containing molecules will improve \texttt{chemberta-2-5M} and \texttt{chemberta-2-10M}'s performance on halogens. 
    \item As further pre-training improves probing performance for halogens, we expect smaller probing performance gains on halogens after fine-tuning on lipophilicity.
\end{enumerate}

\paragraph{Same architecture, different performance}
We conjecture that probing can furthermore be used to identify molecular substructures that a model has seen less during pre-training and that further pre-training can be used to mitigate this gap.
We study this hypothesis on two models with the same architecture (\texttt{chemberta-2-5M} and \texttt{chemberta-2-10M}) which exhibit large differences in probing performance on \texttt{phenol} (\Cref{fig:app_further_pt_bp_performance_PHENOL}, left). 
We conjecture that this is a result of the difference in the pre-training data (and its phenol distribution).

\subsection{Pre-training Dataset}
For further pre-training, we subsample from the Guacamol dataset~\citep{guacamol}, a dataset that is designed for benchmarking de novo molecular design. 
It comes with pre-defined train, validation and test splits with 1,273,104, 79,568 and 238,706 molecules, respectively. 

\paragraph{Pre-processing}
We first canonicalize all SMILES strings and discard 3,217 molecules which appear in the downstream lipophilicity and solubility datasets. 
We then annotate the canonicalized and cleaned training and validation splits of Guacamol with binary labels (where 1 denotes the presence of a substructure) using RDKIT.
For each substructure of interest, we select SMILES strings containing this substructure from the training and validation sets. 
Having obtained the substructure-containing subsets, we randomly sample 50,000 and 3,000 molecules from them.

\subsection{Experimental Setup} \label{ssec:app_further_pt_setup}

\paragraph{Pre-training}
We conducted pre-training for all models on the subset of data sampled from Guacamol for 20 epochs (15,625 steps) with a batch size of 64. 
We used the AdamW optimizer with the default settings of $\beta_1=0.9$ and $\beta_2=0.999$, $\epsilon=\num{1e-8}$ and weight decay of 0.01.
We use a scheduler with a linear learning rate decay of $0.1$ and 100 warm-up steps. 


\paragraph{Probing Setup}
Our probing setup is identical to that summarized in \Cref{sec:app_probing_setup}.

\paragraph{Finetuning Setup}
For fine-tuning on lipophilicity prediction, we follow the setup described in \Cref{sec:app_ft_setup_lipo} and use the best hyperparameters for pre-trained models listed in \Cref{tab:app_final_hyperparams}.

\subsection{Results}

\begin{figure*}[!th]
\centering
        \includegraphics[width=1.0\linewidth]
{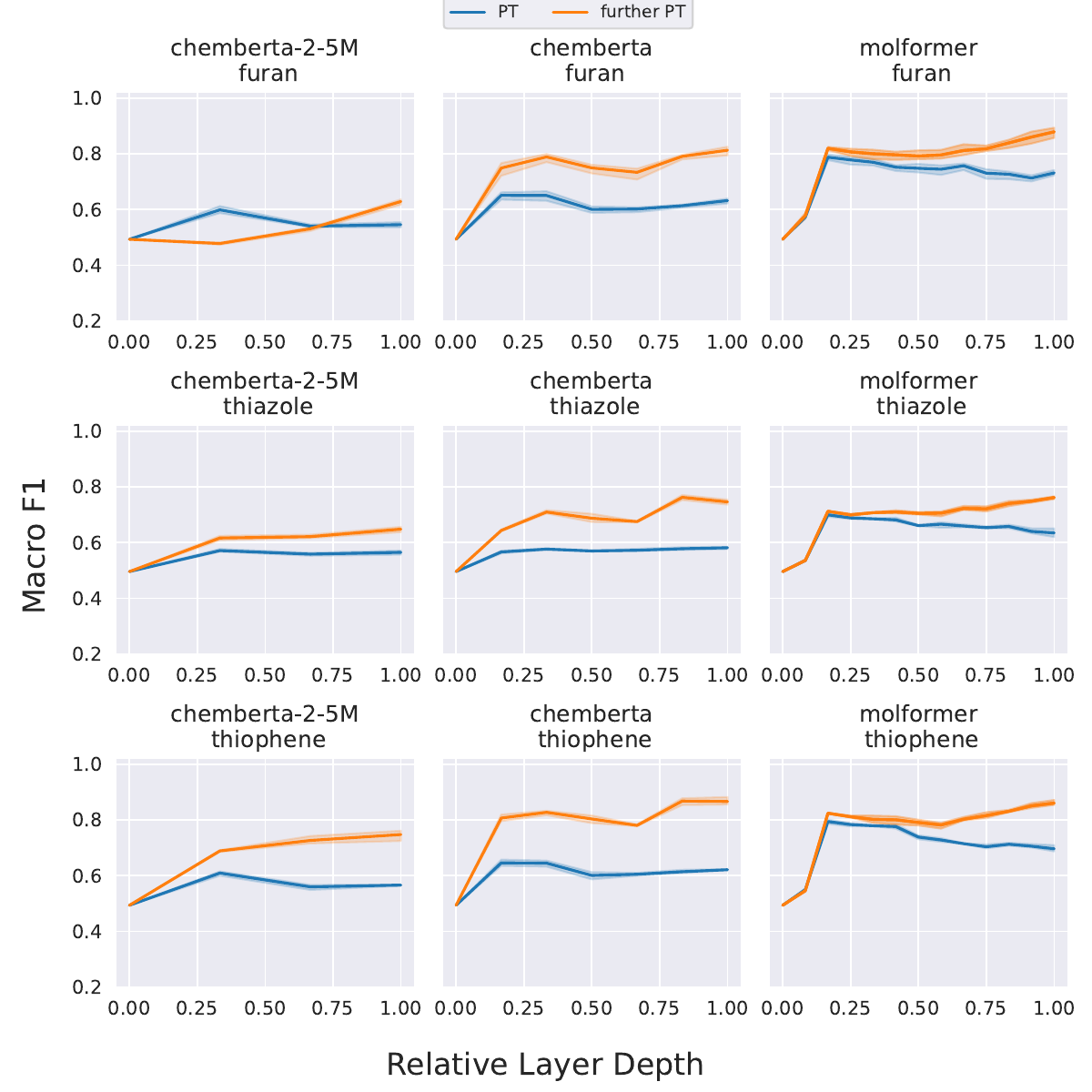}
    \caption{Probing performance of pre-trained models (PT) and those further pre-trained (further PT) on data containing \texttt{furan} (row 1), \texttt{thiazole} (row 2) and \texttt{thiophene} (row 3) on the respective substructure. Across all models, we observe an improvement in probing performance on the specific substructure after additional pre-training with data containing the corresponding substructure.}
    \label{fig:app_further_pt_bp_performance}
\end{figure*}

\begin{table*}[htb]
\centering
\begin{tabular}{@{}llll@{}}
\toprule
\textbf{Model}  &  \multicolumn{3}{c}{\textbf{RMSE}}      \\
 \cmidrule(lr){2-4}

& \textbf{furan} & \textbf{thiazole} & \textbf{thiophene} \\ \midrule
\texttt{molformer}     & 0.596 \small(\textcolor{red}{+0.031}) & 0.597  \small(\textcolor{red}{+0.032})  & 0.594  \small(\textcolor{red}{+0.029})   \\
\texttt{chemberta}     & 0.641 \small(\textcolor{ForestGreen}{-0.034}) & 0.669 \small(\textcolor{Gray}{-0.006})   & 0.667 \small(\textcolor{Gray}{-0.008})    \\
\texttt{chemberta-2-5M} & 0.661 \small(\textcolor{Gray}{-0.003})& 0.631  \small(\textcolor{ForestGreen}{-0.033})  & 0.646 \small(\textcolor{ForestGreen}{-0.018})    \\ \bottomrule
\end{tabular}
\caption{Fine-tuning performance (RMSE, lower is better) of models further pre-trained on data containing \textbf{\texttt{furan}}, \textbf{\texttt{thiazole}}, \textbf{\texttt{thiophene}} on the lipophilicity dataset. \textcolor{ForestGreen}{green} indicates improvement in RMSE (lower is better) while \textcolor{red}{red} denotes increased RMSE. \textcolor{Gray}{Gray} indicates negligible changes (<0.01) in RMSE.}
\label{tab:ft_thiazole_thiophene_furan}
\end{table*}

\paragraph{PT<RI}
\Cref{fig:app_further_pt_bp_performance} shows the impact of further pre-training on more data with specific substructures (\texttt{furan}, \texttt{thiazole}, \texttt{thiophene}) on the probing performance on the corresponding substructure for \texttt{chemberta-2-5M}, \texttt{chemberta} and \texttt{molformer}. 

First, we observe an \textbf{improvement in probing performance across all three substructures and models}. 
While further pre-training benefits the \texttt{chemberta} model most (originally with a pre-training dataset of only 250k molecules), large improvements for \texttt{molformer} are generally observed in the upper layers. 
Compared to the other two models, \texttt{chemberta-2-5M} exhibits more variation in the magnitude of improvement.

Second, \textbf{further pre-training has a mixed effect on the downstream performance on lipophilicity prediction.}
Whereas \texttt{molformer}'s downstream performance is negatively affected in all three cases, \texttt{chemberta-2-5M} mostly benefits from further pre-training. 
In contrast, \texttt{chemberta}'s performance improves when further pre-trained on furan and is negligibly affected by pre-training on \texttt{thiazole} and \texttt{thiophene}. 
This may change under extensive hyperparameter tuning.

\begin{table*}[ht]
\centering

    \begin{tabular}{llcc}
\toprule
\textbf{Model} & \textbf{Layer} & \parbox{2cm}{$\Delta$ \textbf{macro F1 (PT)}} & \parbox{2cm}{$\Delta$ \textbf{macro F1 (halogens)}} \\
\midrule
\multirow[t]{4}{*}{\texttt{chemberta-2-5M}} & 0 & 0.000 &0.000\\
 & 1 & -0.209 &-0.700\\
 & 2 & 0.015 &0.499\\
 & 3 & 2.955 &0.573\\
\cline{1-4}
\multirow[t]{4}{*}{\texttt{chemberta-2-10M}} & 0 & 0.000 & 0.000\\
 & 1 & -1.252& -2.341\\
 & 2 & -1.395 &  -0.212\\
 & 3 & 2.407 & -1.257\\

\cline{1-4}
\multirow[t]{4}{*}{\texttt{chemberta-2-77M}} & 0 & 0.000 &0.000\\
 & 1 & -1.062 &-2.606\\
 & 2 & -0.419& -1.297\\
 & 3 & 4.381 & -2.869\\

\bottomrule
\end{tabular}

\caption{\textbf{Further pretraining on halogens}: Layer-wise difference in probing performance on \textbf{\texttt{halogens}} for \texttt{chemberta-2} models (left) and their counterparts further pre-trained on halogens (right) after fine-tuning.}
\label{tab:app_delta_pt_cb2_halogens}
\end{table*}

\begin{figure*}[hb]
\centering
        \includegraphics[width=0.5\linewidth]
{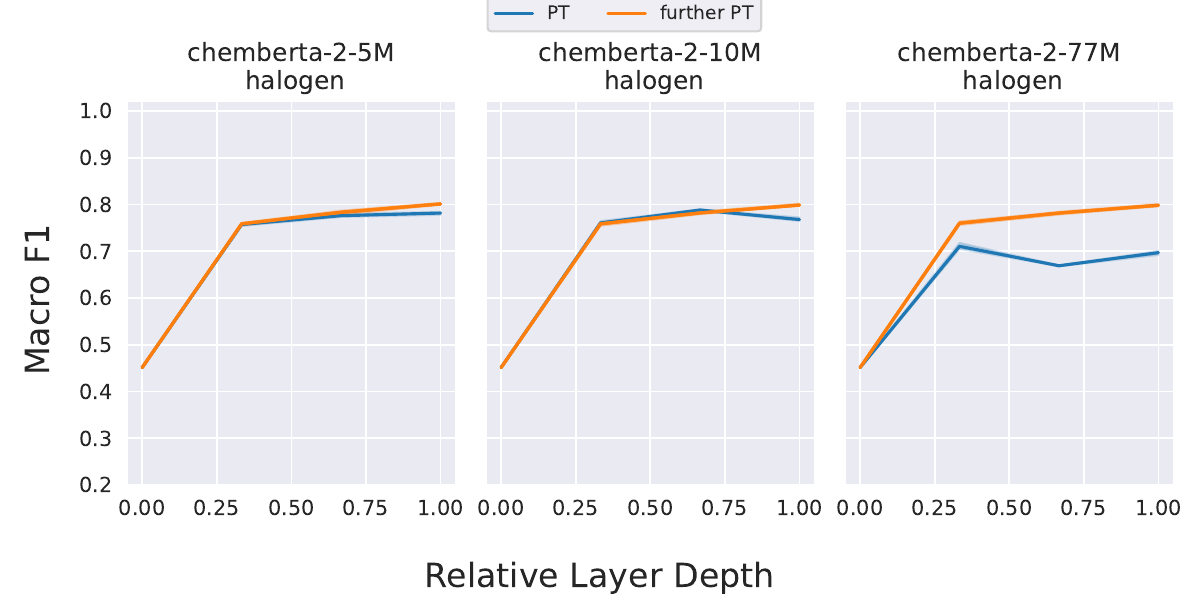}

    \caption{Probing performance of pre-trained models (PT) \texttt{chemberta-2-5M}, \texttt{chemberta-2-10M} and \texttt{chemberta-2-77M} and their counterparts further pre-trained (further PT) on data containing \texttt{halogens} on the \texttt{halogen} substructures.}
    \label{fig:app_further_pt_bp_performance_halogens}
\end{figure*}

\paragraph{Chemberta-2 and Halogens}
\Cref{fig:app_further_pt_bp_performance_halogens} presents the results of further pre-training of models from the \texttt{chemberta-2} family on halogen data. 

\begin{figure*}
    \centering
    \includegraphics[width=0.7\linewidth]{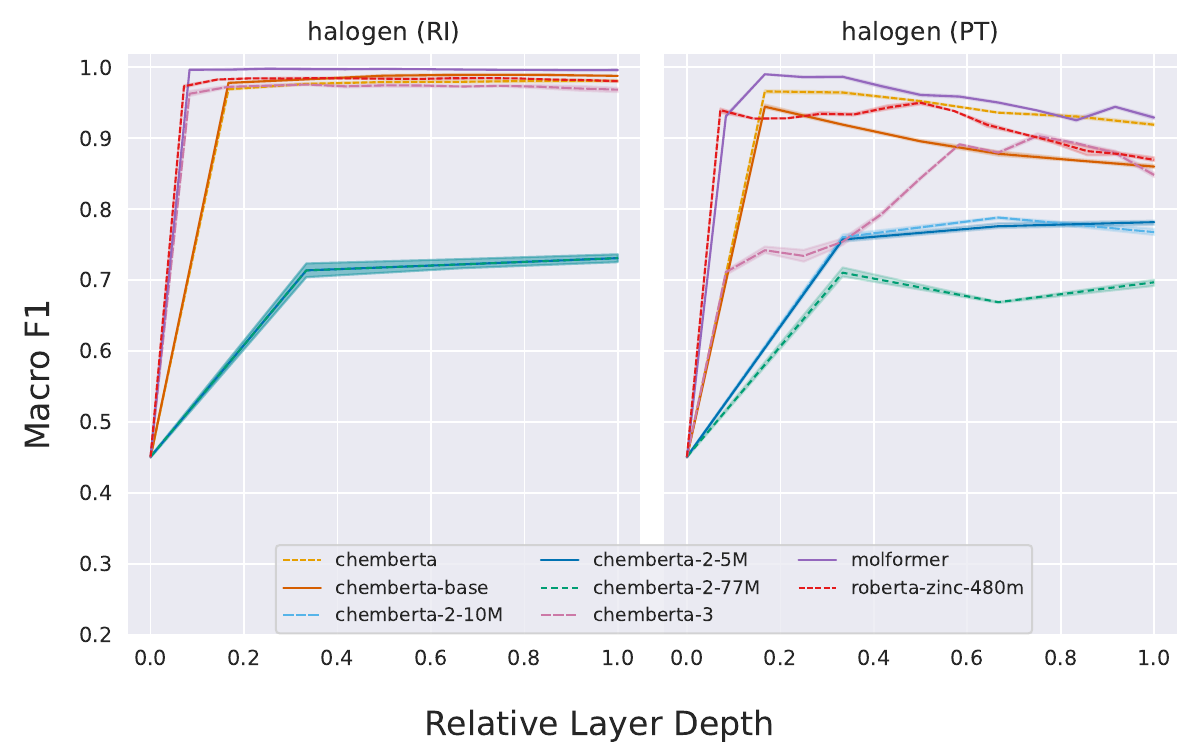}
    \caption{Layerwise probing performance (macro F1 score) on \texttt{halogens} for randomly initialized (RI) and pre-trained (PT) models. We observe that all models except for \texttt{chemberta-2-5M} and \texttt{chemberta-2-10M} unlearn halogens in the middle and upper layers.}
    \label{fig:halogens_base_performance_pt_vs_ri}
\end{figure*}

First, we observe slight performance improvements for \texttt{chemberta-2-5/10M} and substantial performance gains for \texttt{chemberta-2-77M} when probing for \texttt{halogens}. 
Interestingly, we find that the probing performance on halogens converges towards a joint upper bound which is also shared in the randomly initialized \texttt{chemberta-2} model.
This is in stark contrast to all other models shown in \Cref{fig:halogens_base_performance_pt_vs_ri} which have a substantially higher performance in the upper layers. 
One reason for this might be the tokenization issues of the \texttt{chemberta-2} models which result in turning halogens such as Cl and Br into C and B.
\footnote{see \href{https://github.com/seyonechithrananda/bert-loves-chemistry/issues/60}{Issue 1} and \href{https://github.com/deepchem/deepchem/issues/4253}{Issue 2} and \href{https://discuss.huggingface.co/t/tokenizer-not-recognising-words-in-vocabulary/20140}{Issue 3}}

Second, improved probing performance on halogens after further pre-training makes their encoding more robust, resulting in smaller changes after fine-tuning, detailed in \Cref{tab:app_delta_pt_cb2_halogens}. 
We conjecture that these changes might be more prominent without the tokenization issues.

Finally, similar to the effect of further pre-training on \texttt{furan}, \texttt{thiazole} and \texttt{thiophene} on downstream performance, the results in \Cref{tab:ft_cb_halogens_ft_performance} demonstrate that pre-training on more \texttt{halogen} data has a mixed effect on downstream performance. 
Whereas \texttt{chemberta-2-5M} improves on the downstream task and \texttt{chemberta-2-10M}'s performance is substantially negatively affected, \texttt{chemberta-2-77M} remains largely unchanged. 
One reason for this disparity despite all models sharing the same architecture might be the different hyperparameters which were not tuned individually for further pre-trained models. 
\Cref{tab:app_final_hyperparams,tab:app_final_hyperparams_esol} show different optimal hyperparameters for different \texttt{chemberta-2} models across both tasks.
Hence, extensively tuning the hyperparameters for the further pre-trained models might mitigate drops in the performance.

\begin{table*}[htb]
\centering
\begin{tabular}{@{}ll@{}}
\toprule
\textbf{Model}         & \textbf{RMSE}  \\ \midrule
\texttt{chemberta-2-5M}     &  0.646 \small(\textcolor{ForestGreen}{-0.018})   \\
\texttt{chemberta-2-10M} &  0.699 \small(\textcolor{red}{+0.108})   \\ 
\texttt{chemberta-2-77M} & 0.635 \small(\textcolor{Gray}{+0.003})   \\ \bottomrule
\end{tabular}
\caption{Fine-tuning performance (RMSE, lower is better) of \texttt{chemberta-2-5M}, \texttt{chemberta-2-10M}  and \texttt{chemberta-2-77M}  further pre-trained on data containing \textbf{\texttt{halogens}}. \textcolor{ForestGreen}{green} indicates improvement in RMSE (lower is better) while \textcolor{red}{red} denotes increased RMSE. \textcolor{Gray}{Gray} indicates negligible changes (<0.01) in RMSE.}
\label{tab:ft_cb_halogens_ft_performance}
\end{table*}

\begin{figure*}[hb]
\centering
        \includegraphics[width=0.7\linewidth]
{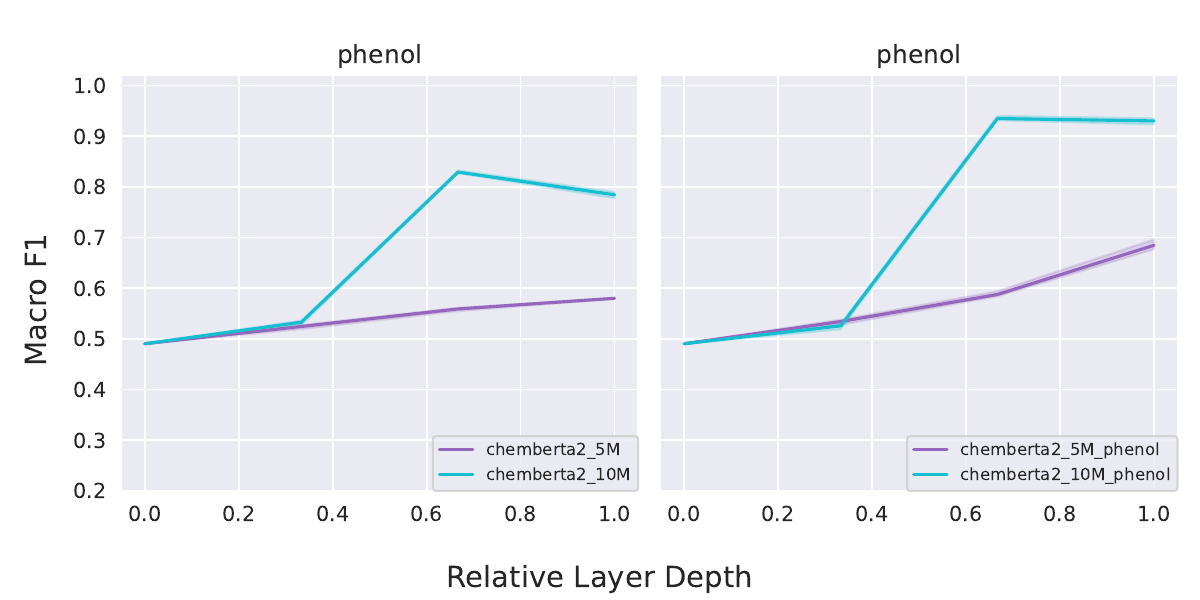}

    \caption{Probing performance of pre-trained models \textcolor{Plum}{\texttt{chemberta-2-5M}} and \textcolor{SkyBlue}{\texttt{chemberta-2-10M}} (left) and the same models further pre-trained on data containing \texttt{phenols} (right) on \texttt{phenols}. We observe that while further pre-training on data containing phenol leads to better probing performance on the corresponding substructure, the performance gap between the two models does not close. We hypothesize that longer pre-training might narrow this gap.}
    \label{fig:app_further_pt_bp_performance_PHENOL}
\end{figure*}

\paragraph{Same architecture, different performance}
\Cref{fig:app_further_pt_bp_performance_PHENOL} shows the effect of pre-training \texttt{chemberta-2-5M} and \texttt{chemberta-2-10M} on more data containing phenol substructures. 

Although further pre-training on data containing phenol improved the respective probing performance, the performance gap between the two models persisted. 
We conjecture that the difference in the amount of original pre-training data (5 million vs 10 million) might be responsible for this and that longer pre-training might further reduce this gap. 
Again, \Cref{tab:ft_cb_phenol_ft_performance} shows that further pre-training can both negatively and positively affect downstream performance. 

\begin{table*}[htb]
\centering
\begin{tabular}{@{}ll@{}}
\toprule
\textbf{Model}          & \textbf{RMSE}   \\ \midrule

\texttt{chemberta-2-5M}     & 0.622 \small(\textcolor{ForestGreen}{-0.042})   \\
\texttt{chemberta-2-10M} & 0.618 \small(\textcolor{red}{+0.027})   \\ \bottomrule
\end{tabular}
\caption{Fine-tuning performance (RMSE, lower is better) of \texttt{chemberta-2-5M} and \texttt{chemberta-2-10M} further pre-trained on data containing \textbf{\texttt{phenol}}. \textcolor{ForestGreen}{green} indicates improvement in RMSE (lower is better) while \textcolor{red}{red} denotes increased RMSE.}
\label{tab:ft_cb_phenol_ft_performance}
\end{table*}

\subsection{Discussion}

Our results suggest that substructure probing may serve as a diagnostic tool to identify whether a model was undertrained on specific substructures. 
Our additional pre-training experiments provide an exploratory demonstration that targeted further pre-training on infrequent substructures can improve the representations of these substructures and increase their representational robustness during fine-tuning.
Based on our findings, systematically studying the pre-training dynamics of \clms via probing could be a promising endeavor for future work. 

\section{The Peculiar Case of Chemberta-3}\label{sec:appendix-chemberta-3}

\texttt{Chemberta-3} exhibits a peculiar "dome" in the average probing performance at lower layers (\Cref{fig:bp}).  
We hypothesize that this might stem from the irregularities during the pre-training process. 
To investigate this as well as the impact of different training data, we further pre-train \texttt{chemberta-3} on subsets of the Guacamol~\citep{guacamol} (CC BY-SA 3.0) and ZINC-100M~\citep{chemberta3} datasets.

\subsection{Datasets}


To obtain the training data, we subsample the Guacamol \citep{guacamol} and ZINC-100M \citep{chemberta3} datasets. 
Guacamol comes with pre-defined train, validation and test splits with 1,273,104, 79,568 and 238,706 molecules, respectively. \citet{chemberta3} curated ZINC-100M by sampling 100M molecules from the ZINC20 database~\citep{zinc20}.

\subsection{Data Preprocessing and Sampling}

\paragraph{Guacamol}
We first canonicalize all SMILES strings. We then discard any molecules which appear in the downstream lipophilicity and solubility datasets. In the case of Guacamol, we discard 3,217 molecules. We then randomly sample 100,000 and 10,000 molecules from the training and validation sets, respectively.

\paragraph{ZINC-100M}
Due to the large size of ZINC-100M ($\sim$7.7GB), we first performed random sampling on the byte positions, obtaining 100,000 and 10,000 molecules for the training and validation sets, respectively. We then canonicalize the SMILES strings and check for SMILES strings which overlap with the those in the fine-tuning datasets.

\subsection{Pre-Training Setup}
We follow the same pre-training setup described in \Cref{ssec:app_further_pt_setup}. However, we train the model for 32 epochs (50k steps).

\subsection{Effect of Further Pre-Training}
\begin{figure*}[!htb]
    \centering
    \includegraphics[width=0.7\linewidth]{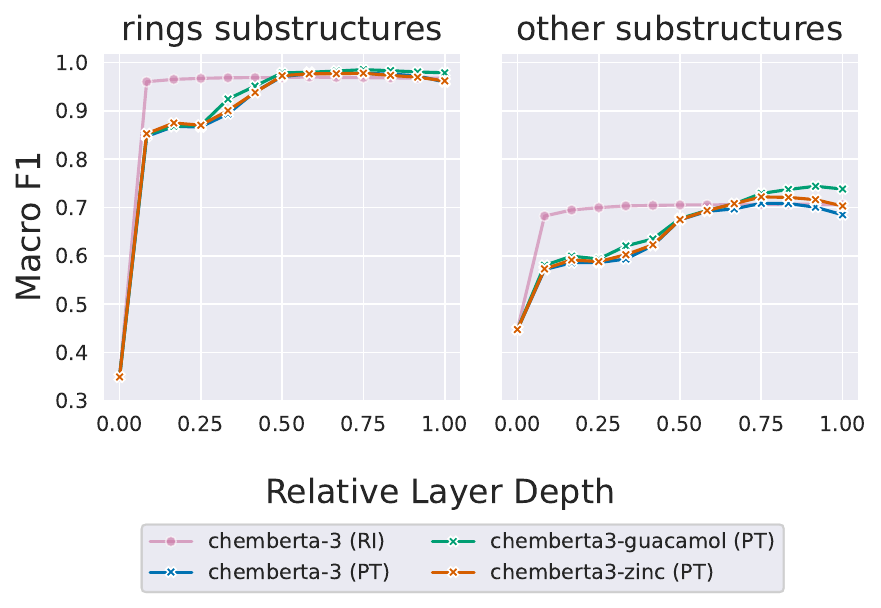}

    \caption{Effect of further pre-training \texttt{chemberta-3} on different data. \textcolor{blue}{\texttt{chemberta-3}~(PT)} denotes the pre-trained model,  \textcolor{RedOrange}{\texttt{chemberta-3-zinc~(PT)}} is the model further pre-trained on a 100k subset of Guacamol, while \textcolor{ForestGreen}{\texttt{chemberta-3-guacamol~(PT)}} has been further pre-trained on a subset of ZINC-100M. \textcolor{Thistle}{\texttt{chemberta-3~(RI)}} is the randomly initialized \texttt{chemberta-3}.}
    \label{fig:app_cb3_further_pt_plots}
\end{figure*}

\begin{figure*}[htb]
  \begin{subfigure}[t]{1.0\textwidth}
        \centering
         \includegraphics[width=0.7\linewidth]
         {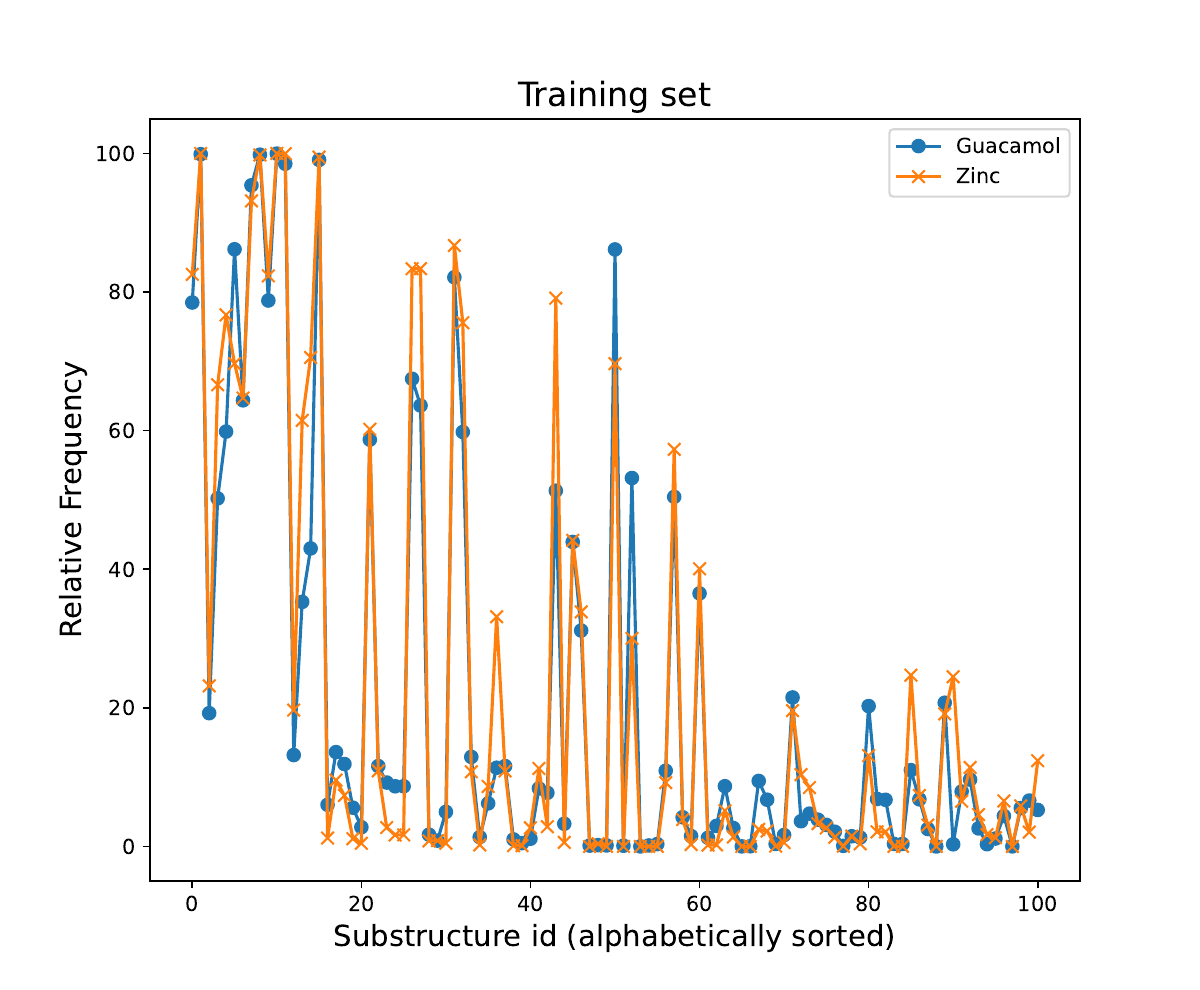}

    \end{subfigure}

    \caption{Comparison of relative frequency distributions of molecular substructures in the training splits of the subsampled ZINC and Guacamol datasets used for further pre-training \texttt{chemberta-3}.}
    \label{fig:freq_distr_plots_zinc_guacamol}
\end{figure*}

\Cref{fig:app_cb3_further_pt_plots} shows the effect of further pre-training \texttt{chemberta-3} on subsets of Guacamol and ZINC-100M. 
Interestingly, the model further pre-trained on Guacamol exhibits substantially better probing performance compared to the model trained on ZINC across all molecular substructures.
Further pre-training on ZINC benefits other (non-ring) substructures more than rings, with improved performance mostly in upper layers. 

Importantly, the lower layers which are characterized by substantially lower performance remain mostly unaffected by further pre-training on ZINC while showing a slight improvement for Guacamol.
To better understand these effects, we analyze the frequency distributions of molecular substructures in \Cref{fig:freq_distr_plots_zinc_guacamol} across both datasets, finding that they are mostly similar.
Surprisingly, the actual overlap in terms of molecules (i.e., SMILES string) is almost zero with only four overlapping strings in the training split.
This suggest that small differences in the training frequency of molecular substructures may already substantially affect the encodings during pre-training.
Furthermore, there might be other effects beyond molecular substructures that need to be studied in future work. 

Nonetheless, we conclude that further pre-training a model on a small, carefully curated dataset can already mitigate some of the negative probing performances. 

\section{Lipophilicity and Solubility Prediction with Other Models}\label{sec:appendix-baselines}

We provide additional experiments on the downstream tasks using models other than \clms. 
More specifically, we investigate models using traditional chemical features, referred to as ``molecular fingerprints'' and more recent, decoder-only foundation models.


\subsection{Models Trained on Fingerprints}\label{sec:appendix-fingerprints}
Fingerprints are a common way of representing molecules in chemistry to be used as features in machine learning models. 
They are often generated using hand-crafted algorithms that output a binarized feature vector representing the presence of specific molecular substructures in a molecule.
One of the most famous methods is Morgan's algorithm~\citep{morgan1965fingerprints}, with various other methods that have been developed over time. 
In this work, we evaluate four different types of fingerprints:

\begin{description}[noitemsep,topsep=3pt,itemsep=3pt]
    \item[\textbf{MFP}] Morgan fingerprints as introduced by \citet{morgan1965fingerprints}.
    \item[\textbf{RDFP}] The fingerprinting method provided in RDKit~\citep{greg_landrum_2020_3732262}.
    \item[\textbf{ATFP}] Atom pair fingerprints introduced by \citep{atompairfingerprits}. In the RDKit implementation, an atom is represented by a tuple of the atomic number, number of pi electrons, and the degree of the atom, with the option of adding chirality information. 
    \item[\textbf{TTFP}] Topoligical torsion fingerprints are similar to atom pair fingerprints but use 4-atom sequences to capture more local graph structure. \citep{topologicaltorsionfingerprints}. 
\end{description}

We extract fingerprints using a radius of 2, which is equivalent to the same value used in MoleculeNet~\citep{wu2018moleculenetbenchmarkmolecularmachine}---the benchmark from which we use the lipophilicity and solubility datasets---and evaluate four different vector sizes $dim$$\in$\{300, 512, 1024, 2048\} 
For each of the extracted fingerprints, we then train a logistic regression model (LR), a support vector machine (SVM), and a gradient boosted tree (XGB).
For the SVM, we additionally tune different values for $c$=\{0.00001, 0.0001, 0.001, 0.01, 0.1, 1, 4, 16, 64, 256, 1024\} using the validation set. 

\begin{table*}
  \centering
  \begin{tabular}{lrllllc}
    \toprule
    Model & dim & MFP & RDFP & ATFP & TTFP & Avg \\
    \midrule
    LR & 300 & 0.84 & 1.02 & 0.88 & 0.90 & 0.91 \\
     & 512 & 0.84 & 1.00 & 0.91 & 0.87 & 0.91 \\
     & 1024 & 0.85 & 0.98 & 0.95 & 0.95 & 0.93 \\
     & 2048 & 1.09 & 1.20 & 1.27 & 1.05 & 1.15 \\
     & Avg & 0.91 & 1.05 & 1.00 & 0.94 & - \\
     \midrule
    SVM & 300 & 0.72 & 0.94 & 0.75$^*$ & 0.77 & 0.80 \\
     & 512 & 0.71 & 0.89$^*$ & 0.73$^*$ & 0.73 & 0.77 \\
     & 1024 & 0.68 & 0.75 & 0.68 & 0.70 & 0.70 \\
     & 2048 & 0.67 & 0.72 & \textbf{0.64}$^*$ & 0.69 & 0.68\\
     & Avg & 0.70 & 0.83 & 0.70 & 0.72 & - \\
     \midrule
    XGB & 300 & 0.76 & 0.97 & 0.80 & 0.80 & 0.83 \\
     & 512 & 0.73 & 0.91 & 0.76 & 0.80 & 0.80 \\
     & 1024 & 0.72 & 0.83 & 0.74 & 0.74 & 0.76 \\
     & 2048 & 0.72 & 0.79 & 0.69 & 0.73 & 0.73 \\
     & Avg & 0.73 & 0.88 & 0.75 & 0.77 & - \\
     \midrule
   Avg & - & 0.78 & 0.92 & 0.82 & 0.81 & - \\

\bottomrule
  \end{tabular}
  \caption{Test set performance [RMSE ($\downarrow$)] on \textbf{lipophilicity}. We evaluate four different methods for generating fingerprints (MFP, RDFP, ATFP, TTFP) and evaluate dimension sizes of 300, 512, 1024, and 2048. For the SVM, c=4 worked best except for the ones marked with $^*$ (c=16). Interestingly, we find that both XGB and SVM perform better for larger dimensions, while LR performs better for lower dimensions.}
  \label{tab:ft-fingerprints-lipo}
\end{table*}

\begin{table*}
  \centering
  \begin{tabular}{lrllllc}
    \toprule
    Model & dim & MFP & RDFP & ATFP & TTFP & Avg \\
    \midrule
    LR & 300 & 1.81 & 1.63 & 1.44 & 1.39 & 1.57 \\
     & 512 & 2.06 & 1.93 & 2.11 & 1.87 & 1.99 \\
     & 1024 & 4.74 & 6.87 & 367.63 & 11.18 & 97.61 \\
     & 2048 & 2.53 & 2.61 & 5.84 & 4.03 & 3.75 \\
     & Avg & 2.79 & 3.26 & 94.26 & 4.62 & - \\
     \midrule
    SVM & 300 & 1.19$_{c=256}$ & 1.02$_{c=64}$ & 0.93$_{c=16}$ & 0.99$_{c=16}$ & 1.03 \\
    & 512 & 1.06$_{c=256}$ & 0.91$_{c=256}$ & 1.00$_{c=4}$ & 1.14$_{c=4}$ & 1.03 \\ 
    & 1024 & 1.06$_{c=64}$ & 0.90$_{c=1024}$ & 0.92$_{c=4}$ & 1.20$_{c=16}$ & 1.02 \\ 
    & 2048 & 1.00$_{c=256}$ & 0.88$_{c=1024}$ & \textbf{0.83$_{c=16}$} & 1.11$_{c=16}$ & 0.96 \\
     & Avg & 1.08 & 0.93 & 0.92 & 1.11 & - \\
     \midrule
    XGB & 300 & 1.23 & 1.16 & 1.01 & 1.08 & 1.12 \\
    & 512 & 1.12 & 1.06 & 1.01 & 1.14 & 1.08 \\
    & 1024 & 1.13 & 1.01 & 1.04 & 1.16 & 1.09 \\
    & 2048 & 1.17 & 0.86 & 0.93 & 1.17 & 1.03 \\
     & Avg & 1.16 & 1.02 & 1.00 & 1.13 & - \\
     \midrule
   Avg & - & 1.66 & 1.74 & 32.06 & 2.29 & - \\

\bottomrule
  \end{tabular}
  \caption{Test set performance [RMSE ($\downarrow$)] on \textbf{ESOL} (solubility prediction). We evaluate four different methods for generating fingerprints (MFP, RDFP, ATFP, TTFP) and evaluate dimension sizes of 300, 512, 1024, and 2048. For the SVM, we report the best-performing c for each configuration. We find that both XGB and SVM maintain a robust performance around 1.00 across all dimensions, while LR performance varies substantially, especially for $\text{dim}=1,024$.}
  \label{tab:ft-fingerprints-esol}
\end{table*}

\paragraph{Results.} 
\Cref{tab:ft-fingerprints-lipo} and \Cref{tab:ft-fingerprints-esol} show the performance (RMSE) of different fingerprinting algorithms across different models for lipophilicity and solubility, respectively. 
Most notably, the SVM consistently performs best across all dimensions, fingerprints, and tasks, showing the best performance for atom pairs fingerprinting (ATFP) with a dimension of 2048 and a $c$ of 16. 
Interestingly, we find that Morgan fingerprints (MFP) perform most stable across different dimensions and best on average. 
Finally, we find that for lipophilicity prediction the SVM and XGB both benefit from higher dimensions, while this is the opposite for the LR model. 
Similarly, the SVM and XGB both perform robustly across all dimensions and fingerprints, while LR performance varies a lot; especially for higher dimensions. 

\subsection{Foundation Models}\label{sec:appendix-LLM}

Recently, an increasing number of works have investigated the potential of foundation models for chemistry tasks~\citep{guo2023largelanguagemodelschemistry,Guo2024MolPuzzle}.
While they find that some do perform decently on MPP tasks, they primarily focus on classification tasks.
In this work, we provide complementary results on two regression tasks (i.e., lipophilicity and solubility prediction).

\paragraph{Experimental setup.}
We conduct experiments for two large language models, namely \texttt{Llama-3.2-3B-Instruct}~\citep{dubey2024llama} and \texttt{gpt-oss-20B}~\citep{openai2025gptoss120bgptoss20bmodel}. 
We prompt each model in a zero-shot setting, asking it to predict the logD (or logS) value of the given molecule (basic). 
We further evaluate three additional setups to accommodate for the model's lack of chemical knowledge. 
First, we provide explanations on which functional groups increase and decrease the  logD (or logS) value (expl).
Second, we provide a list of extracted molecular substructures present in the molecule (hint).
Finally, we pass both piece of information to the model (both).
For generation, we use nucleus sampling~\citep{Holtzman2020The} with $p=0.95$, a temperature of 0.8 and a maximum token budget of 4,096 tokens.
We evaluate all three reasoning levels for the \texttt{gpt-oss-20B} model, i.e., low, medium, and high.
For the high reasoning level, we find that the model tends to generate very lengthy responses that exceed the token budget. 
We thus further evaluate token budgets of 8,192 and 16,384. 
All experiments were conducted on a high performance computing cluster with 4 NVIDIA A100 (40GB). 
Each GPU ran for $\approx$112 hours ($\sim$4.7 days in total).

\begin{table*}
  \centering
  \begin{tabular}{lcrrrr}
    \toprule
    Model & reasoning & basic & expl & hint & both  \\
    \midrule
    SVM$_\text{(c=16)}$ + ATFP & \multicolumn{5}{c}{0.64} \\
    \texttt{molformer}  & \multicolumn{5}{c}{0.57} \\ 
    \midrule
    \texttt{Llama-3.2-3B-Instruct} & - & 188.24 & 61.13 & 42.06 & 25.93 \\
    \texttt{gpt-oss-20b}$_{4,096}$ & low & 2.89 & 2.66 & 2.82 & 2.84 \\
    \texttt{gpt-oss-20b}$_{4,096}$ & medium & 3.11 & 2.80 & \textbf{2.53} & 2.94 \\
    \texttt{gpt-oss-20b}$_{4,096}$ & high & 237,981.69 & 246.42 & 19.95 & 98.97 \\
    \texttt{gpt-oss-20b}$_{8,192}$ & high & 139.36 & 43.04 & 27.67 & 24.60 \\
    \texttt{gpt-oss-20b}$_{16,384}$ & high & 61.84 & 376.55 & 337.18 & 15.45 \\
\bottomrule
  \end{tabular}
  \caption{Test set performance [RMSE ($\downarrow$)] on lipophilicity. We find that the \texttt{gpt-oss-20b}$_{4,096}$ model using the medium reasoning level together with hints performs best out of all LLMs and that providing explanations as well as hints can improve the performance. Nonetheless, all LLMs are substantially outperformed by the SVM using fingerprints and the molformer model.}
  \label{tab:llm-results-lipo}
\end{table*}

\begin{table*}
  \centering
  \begin{tabular}{lcrrrr}
    \toprule
    Model & reasoning & basic & expl & hint & both  \\
    \midrule
    SVM$_\text{(c=16)}$ + ATFP & \multicolumn{5}{c}{0.83} \\
    \texttt{molformer}  & \multicolumn{5}{c}{0.59} \\ 
    \midrule
    \texttt{Llama-3.2-3B-Instruct} & - & 233.11 & 213.79 & 477.07 & 940,720,868.84 \\
    \texttt{gpt-oss-20b}$_{4,096}$ & low & 9.27 & 31.97 & 9.51 & \textbf{8.96} \\
    \texttt{gpt-oss-20b}$_{4,096}$ & medium & 48.95 & 51.68 & 14.11 & 23.31 \\
    \texttt{gpt-oss-20b}$_{4,096}$ & high & 193.50 & 41.91 & 95.61 & 36.32 \\
    \texttt{gpt-oss-20b}$_{8,192}$ & high & 23.53 & 44.42 & 42.34 & 193.03 \\
    \texttt{gpt-oss-20b}$_{16,384}$ & high & 32.94 & 83.57 & 19.39 & 60.72 \\
\bottomrule
  \end{tabular}
  \caption{Test set performance [RMSE ($\downarrow$)] on esol (solubility prediction). We find that the \texttt{gpt-oss-20b}$_{4,096}$ model using the low reasoning level and both (hints and explanations) performs best. Interestingly, we find that especially for larger contexts, providing explanations or hints can deteriorate the performance. Again, all LLMs are substantially outperformed by the SVM using fingerprints and the molformer model.}
  \label{tab:llm-results-esol}
\end{table*}

\paragraph{Results.}
\Cref{tab:llm-results-lipo} and \Cref{tab:llm-results-esol} provide the results of our experiments for lipophilicity prediction and solubility prediction, respectively. 
Overall, we can see that all LLMs perform worse compared to the models that use \clms or fingerprints. 
We further find that a high reasoning level does not \texttt{gpt-oss-20b} automatically lead to an improve performance but instead, can produce responses that exceed the token budget (as it consistently happens for the high reasoning level). 
Interestingly, providing the models with either explanations or information about the present functional groups does improve their performance, however providing both leads to a worse performance for lipophilicity prediction.
This follows the findings by \citet{ganeeva-etal-2024-lost, ganeeva-etal-2025-two} who find that LLMs might still be lacking in terms of compositionality (as they seem incapable of putting together the provided explanation and the functional groups).
In contrast, we do not consistently observe this behavior for solubility prediction, as providing both even leads to the best result (for \texttt{gpt-oss-20b}$_\text{4,096}$, low). 
However, we also see that the performance of smaller models or lower and medium reasoning levels is substantially worse for solubility prediction compared to lipophilicity prediction.
\Cref{fig:gpt-oss-response} showcases an example response for the best performing LLM (\texttt{gpt-oss-20b}$_4,096$, with hints and medium reasoning level). 
As can be seen, the model seemingly ``reasons'' about the task, but assigns the wrong sign to the predicted logD value, indicating that it does not have actual knowledge about the task.

\paragraph{Prompt templates.} 
In the following, we provide all prompt templates that we used in our experiments. 
Note, that the basic prompt (we provide individual templates for lipophilicity and solubility prediction) is always present, and that the respective sub-prompts are appended accordingly. 
We always provide the molecule last with the prefix (``This is the molecule'') shifted accordingly.
The setting both combines all three templates (one of the basic templates, chemical explanations, and preprocessed functional groups).

\begin{tcolorbox}[
    colframe=Gray!40!black,
    colback=Gray!5,
    coltitle=white,
    fonttitle=\bfseries,
    title=Basic Prompt (lipophilicity),
    colbacktitle=Gray!40!black
]
\textbf{System}: You are now working as an excellent expert in chemistry and drug discovery.
\newline
\textbf{User}: Predict the lipophilicity the following molecule. The lipohilicity of a molecule is defined by the logD value, the logarithmic form of the distribution coefficient D. The molecule is represented using the canonical form of its SMILES string.
\newline 

This is the molecule: 

\{canonicalized SMILES string\}
\end{tcolorbox}

\begin{tcolorbox}[
    colframe=Gray!40!black,
    colback=Gray!5,
    coltitle=white,
    fonttitle=\bfseries,
    title=Basic Prompt (solubility),
    colbacktitle=Gray!40!black
]
\textbf{System}: You are now working as an excellent expert in chemistry and drug discovery.
\newline
\textbf{User}: Predict the solubility of the following molecule. The solubility of a molecule is measured by the log solubility in mols per liter. The molecule is represented using the canonical form of its SMILES string.
\newline 

This is the molecule: 

\{canonicalized SMILES string\}
\end{tcolorbox}

\begin{tcolorbox}[
    colframe=Gray!40!black,
    colback=Gray!5,
    coltitle=white,
    fonttitle=\bfseries,
    title=+ Chemical Explanations:,
    colbacktitle=Gray!40!black
]
For prediction it is also important to consider the following functional groups that affect \{lipophilicity, solubility\}.
These are functional groups that increase \{lipophilicity, solubility\}:

\{List of groups that increase logD or logS value\}

These are functional groups that decrease \{lipophilicity, solubility\}:

\{List of groups that decrease logD or logS value\}
\newline

This is the molecule: 

\{canonicalized SMILES string\}
\end{tcolorbox}

\begin{tcolorbox}[
    colframe=Gray!40!black,
    colback=Gray!5,
    coltitle=white,
    fonttitle=\bfseries,
    title=+ Preprocessed Functional Groups:,
    colbacktitle=Gray!40!black
]
As a hint, you are also given the functional groups that are present in the molecule. Following functional groups are present in the molecule:

\{all present groups\}
\newline

This is the molecule:

\{canonicalized SMILES string\}

\end{tcolorbox}

\clearpage

\begin{figure*}[!htb]
  \begingroup
  \scriptsize
  \arrayrulecolor{black}
  \newcolumntype{L}[1]{>{\small\raggedright\arraybackslash}m{#1}}
  \begin{tabular}{|*{1}{@{ }L{.98\textwidth}@{ }|}}
    \hline

We need to predict logD value (probably logD at physiological pH, e.g., logD7.4). This is a computational estimate. We are given a SMILES: C\#Cc1cccc(Nc2nc3cc(C(=O)O)ccc3c3cncnc23)c1
\\ \\

Let's parse: It's a heteroaromatic system fused? Actually: C\#C-? Let's analyze. The base skeleton: phenyl ring (c1cccc...) with a -C\#C substituent at position 1? The SMILES: C\#C c1 c c c c (N c2 n c3 c c (C(=O)O) c c c3 c3 c n c n c2 3) c1. So there is a phenyl ring with a -C\#C group (propyne). At the para position (maybe 4) there is an anilide linking to a bicyclic heteroaromatic (imidazo[...])? Actually c2nc3cc(C(=O)O)ccc3c3cncnc23. That seems to be a fused bicyclic heteroaromatic containing two rings: a benzodiazine? It appears to be a quinazoline ring (a benzene fused with a diazine). Quinazoline: fused benzene and pyrimidine. So we have a 4-? It attaches via an aniline NH to the quinazoline ring.
\\ \\

Also there is a carboxylic acid group (C(=O)O) attached to the quinazoline ring.
\\ \\

The entire molecule contains one carboxylic acid (pKa ~4.5). At physiological pH, it will be deprotonated (anionic). This will significantly reduce logD. Also there is one NH (aniline). There are hetero nitrogens: 4? In quinazoline ring: two ring nitrogens (positions 1 and 3). Additional ring may have two more? Wait: c3cncnc23: That is a fused pyrimidine? Actually c3 c n c n c 2 3. So ring 3 has two nitrogens. So total hetero atoms: 4 ring nitrogens, 1 amine N, 1 carboxylate O's (2). So 5 nitrogens, 3 oxygens. So high hetero content.
\\ \\

Also there's a nitrile (C\#C) group: actually it's an alkyne (C\#C) group attached to phenyl; no hetero.
\\ \\

So logP (neutral) might be moderate-high due to aromatic rings ($\sim$3?). But logD7.4 will be lower due to deprotonated carboxylate (negative). Many hetero nitrogens may also be basic? In quinazoline, the ring nitrogens are not protonated at physiological? They are aromatic; they can act as weak bases (pKa ~2-3). So mostly neutral. The aniline NH is weakly basic; likely neutral.
\\ \\

So overall, the logD7.4 may be around 1.5? Actually carboxylate reduces lipophilicity drastically. For example, carboxylate-containing molecules like ibuprofen logD7.4 $\sim$3.5? Wait ibuprofen is acid (pKa ~4.5). Ibuprofen has logP 3.5, logD7.4 $\sim$3.5? Actually at neutral, logD decreases a bit; but still high due to hydrophobic. But our molecule has 3 aromatic rings (two fused hetero aromatic + phenyl). That gives high aromatic surface. Carboxylate reduces logP by $\sim$2 units maybe. For a 3-ring acid, logD7.4 maybe $\sim$2.5-3.0. Example: diclofenac (two phenyl rings + acid) has logP 4.5, logD7.4 4.0? Actually diclofenac logP 4.5, logD7.4 ~4.0? Not sure. Another acid with hetero: warfarin logP 3.1, logD7.4 $\sim$2.5. So our compound may be similar to warfarin.
\\ \\

Thus I'd guess logD7.4 $\sim$2.0-2.5. Let's pick 2.3.
\\ \\

But maybe due to hetero nitrogens, more hydrophilic, so logD lower $\sim$1.8.
\\ \\

Let's compute approximate cLogP using fragments: Each phenyl ring $\sim$2.3; hetero ring reduces; carboxylate reduces $\sim$1.5. Hard.
\\ \\

I'll answer: logD $\approx$ 1.9.
\\ \\

Given the question: \"Predict the lipophilicity the following molecule.\" So I should provide a single numeric value. I'll answer 2.0 ($\pm$0.5). Provide explanation.
\\ \\

Let's produce final.assistantfinal**Predicted logD (pH~$\approx$~7.4)**  
$\approx$ **2.0**
\\ \\

The molecule contains three aromatic rings (one phenyl and one fused benzodiazine) that give it a relatively high baseline lipophilicity, but the presence of a carboxylic acid (which will be largely anionic at physiological pH) and several hetero nitrogen atoms dramatically reduces its distribution into the octanol phase.  A logD value around 2.0 is consistent with similar tricyclic heteroaromatic acids (e.g., warfarin, diclofenac-like scaffolds) when measured at neutral to slightly basic conditions.
\\
    \hline
  \end{tabular}
  \endgroup
  \caption{Response of \texttt{gpt-oss-20b}$_{4,096}$ with medium reasoning level and preprocessed functional groups as hints. While the model seemingly ``reasons'' about possible implications of different molecular subgroups, it predicts a logD value of 2.0 while the true value lies at -1.067.}
  \label{fig:gpt-oss-response}
\end{figure*}

\end{document}